# A Generative Deep Learning Approach for Crash Severity Modeling with Imbalanced Data

Junlan Chen[1,2], Ziyuan Pu[1, 3*], Nan Zheng[2], Xiao Wen[4], Hongliang Ding[5], Xiucheng Guo[1]

[1]School of Transportation, Southeast University, No.2 Southeast University Road, Nanjing, China, 211189. E-mail: chenjunlan@seu.edu.cn; ziyuan.pu@monash.edu; seuguo@163.com

[2]Department of Civil Engineering, Monash University, Melbourne, Australia. E-mail: nan.zheng@monash.edu

[3]SKey Laboratory of Transport Industry of Comprehensive Transportation Theory (Nanjing Modern Multimodal Transportation Laboratory), Ministry of Transport, PRC.

[4]Department of Civil and Environmental Engineering, The Hong Kong University of Science and Technology, Clear Water Bay, Kowloon, Hong Kong SAR. E-mail: xwenan@connect.ust.hk

[5]Institute of Smart City and Intelligent Transportation, Southwest Jiaotong University. E-mail: hongliang.ding@connect.polyu.hk

Corresponding author: Ziyuan Pu, ziyuan.pu@monash.edu

**ABSTRACT**

Crash data is often greatly imbalanced, with the majority of crashes being non-fatal crashes, and only a small number being fatal crashes due to their rarity. Such data imbalance issue poses a challenge for crash severity modeling since it struggles to fit and interpret fatal crash outcomes with very limited samples. Usually, such data imbalance issues are addressed by data resampling methods, such as under-sampling and over-sampling techniques. However, most traditional and deep learning-based data resampling methods, such as synthetic minority oversampling technique (SMOTE) and generative Adversarial Networks (GAN) are designed dedicated to processing continuous variables. Though some resampling methods have improved to handle both continuous and discrete variables, they may have difficulties in dealing with the collapse issue associated with sparse discrete risk factors. Moreover, there is a lack of comprehensive studies that compare the performance of various resampling methods in crash severity modeling. To address the aforementioned issues, the current study proposes a crash data generation method based on the Conditional Tabular GAN. After data balancing, a crash severity model is employed to estimate the performance of classification and interpretation. A comparative study is conducted to assess classification accuracy and distribution consistency of the proposed generation method using a 4-year imbalanced crash dataset collected in Washington State, U.S. Additionally, Monte Carlo simulation is employed to estimate the performance of parameter and probability estimation in both two- and three-class imbalance scenarios. The results indicate that using synthetic data generated by CTGAN-RU for crash severity modeling outperforms using original data or synthetic data generated by other resampling methods. This study can provide valuable insights for traffic safety researchers and engineers into crash severity modeling, especially when handling imbalanced crash data with various data types.

# 1 INTRODUCTION

During recent decades, considerable efforts have been put into exploring the correlations between crash severity and risk factors (Pu *et al.* 2021b, Wen *et al.* 2021a, Li *et al.* 2024). In general, a set of crash data is essential to feed the multi-dimensional features of different crash types (fatal injury (FI) and non-fatal injury (nFI) crashes) to crash severity models (Zeng *et al.* 2019, Kardar and Davoodi 2020, Pu *et al.* 2021a). However, the rare nature of fatal crashes results in inherent imbalance issues of the crash dataset which usually contains excessive non-fatal crashes and very limited fatal crashes. The issue of data imbalance significantly impacts the performance of crash severity modeling, particularly when it comes to classifying and interpreting crash severity. (Abd Rahman and Yap 2016, Yahaya *et al.* 2020a, Liu *et al.* 2021).

In previous studies, such data imbalance issue is usually addressed through data resampling methods. The imbalanced crash dataset is re-balanced by under-sampling the samples in major categories (Yang et al. 2018, Wang et al. 2019) or over-sampling the samples in minor categories (Abou Elassad *et al.* 2020, Peng *et al.* 2020). However, according to previous studies (Johnson and Khoshgoftaar 2019), under-sampling techniques potentially make the entire dataset lose valuable information about non-fatal crashes. Although over-sampling techniques avoid such information loss, they may cause overfitting issues and fail to capture distribution of risk factors during the data resampling process (Tantithamthavorn et al. 2020). For example, Cai *et al.* (2020) conducted a comparison of synthetic crash data distribution using resampling techniques like SMOTE and under-sampling. They observed that SMOTE led to overfitting, while under-sampling struggled to adequately represent the non-crash data. There are some other studies proposed to combine both under- and over-sampling (also known as mixed-sampling) to generate synthetic data. Results show that mixed-sampling can achieve better performance compared to solely relying on under- or over-sampling (Puri and Gupta 2021).

To address the shortcomings of traditional resampling methods that cannot capture the distribution of data well, researchers have proposed using deep learning-based generative models, such as Generative Adversarial Networks (GANs), to generate synthetic samples in minority classes.(Cai *et al.* 2020, Lin *et al.* 2020). GANs optimize the generator and discriminator in a min-max game to minimize the distribution discrepancy between synthetic and original data (Goodfellow *et al.* 2014, Lin *et al.* 2020). Previous literature has documented that GAN-based models outperform traditional resampling methods in terms of maintaining distribution consistency and achieving higher prediction accuracy when modeling with synthetic data generated by GANs. (Cai *et al.* 2020, Lin *et al.* 2020). Nevertheless, these models are designed for continuous variables (Katal *et al.* 2013, Cai *et al.* 2020). While discrete risk factors such as road environments, weather conditions and driver characteristics, play an important role in crash severity modeling, and often have sparse representations in available data (Pu *et al.* 2020, Li *et al.* 2021a, Wen *et al.* 2021b). To address this issue, a data resampling method was proposed on Conditional Tabular Generative Adversarial Networks (CTGAN), which utilize a mode-specific normalization mechanism with variational Gaussian mixture mode (VGM) to handle continuous and discrete variables simultaneously. A notable advantage of CTGAN is its effectiveness in handling sparse variables, which makes it a suitable method for crash severity modeling. This is achieved through the use of conditional vectors in the generator that contain information about the discrete variables, thus enabling the generation of samples with a distribution similar to that of the original data (Xu *et al.* 2019).

Thus, in this study, we introduce a crash data generation method based on CTGAN. Following data balancing, we utilize a crash severity model to evaluate classification and interpretation performance. Furthermore, we conduct a comparative study to assess the classification accuracy and distribution consistency of the proposed generation method using real crash data. Additionally, Monte Carlo simulation is used to estimate the performance of parameter and probability estimation in both two- and three-class imbalance scenarios. The major contributions of the study are summarized as follows:

(1) A crash data generation method is developed based on CTGAN which is capable of handling continuous and discrete risk factors simultaneously. To effectively address the sparsity of discrete risk factors, the proposed data generation method used a conditional generator to estimate the distribution of real data for sample generation.

(2) The experiment using real data and Monte Carlo Simulation are conducted to assess the interpretative consistency of synthesized samples, focusing on distribution consistency and parameter recovery. These experiments cover three types of scenarios: two-class imbalance, three-class imbalance, and various resampling.

(3) A comprehensive study is conducted to compare the performance of three types (over-, under-, mixed-sampling) of resampling methods in crash severity modeling (e.g., SMOTE-NC, TVAE, random under-sampling (RU)) to the proposed data generation method. According to the results, the proposed data generation method outperforms all other baseline models.

The following paper is organized in the following manner. Section 2 reviews the literature on both traditional and deep learning based resampling methods in traffic safety research. Section 3 describes the crash severity modeling framework, including CTGAN, baseline resampling methods, crash severity model, and evaluation metrics. Section 4 presents the data preparation process. Model performance, sensitivity analysis, and model performance evaluation are included in Section 5. Finally, Section 6 provides conclusions and recommendations for future research.

## 2 LITERATURE REVIEW

### *2.1 Traditional Resampling Methods*

The development of crash severity classifiers is a significant challenge due to imbalanced crash data, as noted by (Chen *et al.* 2016, Yahaya *et al.* 2020b, Wen *et al.* 2021a)). In crash severity modeling, Typical methods to address this issue consist of under-sampling (Ha and Lee 2016), over-sampling (Zhang *et al.* 2023), and mixed-sampling (Katrakazas *et al.* 2019). To balance crash severity data, under-sampling drops non-fatal crash samples while over-sampling generates fatal crash samples. For example, Ariannezhad *et al.* (2021) conducted RU with random forest for real-time crash prediction, and Jeong *et al.* (2018) classified motor vehicle crash injuries based on a dataset using under-sampling in combination with bagging. While under-sampling simplifies implementation, it may result in the loss of potentially valuable information. When modeling with resampled data through under-sampling, it can have a detrimental effect on classification accuracy, particularly in the presence of a high-class imbalance (i.e., a majority-to-minority class ratio ranging from 100:1 to 10,000:1) (Leevy *et al.* 2018, Johnson and Khoshgoftaar 2019). Recent studies indicated that over-sampling techniques such as random oversampling and Synthetic Minority Oversampling Technique (SMOTE) outperform under-sampling methods, as they can retain all information from the original data and result in better performance in terms of prediction accuracy and goodness-of-fit (Peng *et al.* 2020, Yahaya *et al.* 2020b). Kovacs (2019) implemented and evaluated 85 oversampling techniques, such as polynom-fit-SMOTE, ProWSyn, and SMOTE-IPF. It is identified that among the algorithms with the top three best performances, there are two SMOTE-based models. However, SMOTE-based models are subject to generalization, overfitting, and variance issues (Islam *et al.* 2021, Li *et al.* 2021b). Additionally, most SMOTE-based models are limited to handling only continuous variables and are unable to effectively capture the interactions among variables during data generation (Katal *et al.* 2013, Cai *et al.* 2020).

*2.2 Deep Learning-Based Resampling Methods*

In recent years, deep learning generative models have been in the spotlight in traffic safety analysis, particularly when dealing with complex data structures (e.g., nonlinear dependencies, high-dimensional interactions) in crash severity modeling. For example, (Islam *et al.* 2021) adopted a variational autoencoder (VAE) to generate crash data for minority classes and found that VAE could overcome the issue of overfitting due to its flexible decision boundaries, allowing for the generation of diverse and realistic samples from different parts of the data distribution. In addition to VAE, generative adversarial network (GAN) has been also leveraged to deal with the real-time crash prediction problem with imbalanced crash datasets (Cai *et al.* 2020, Lin *et al.* 2020). For example, Cai *et al.* (2020) adopted a deep convolutional generative adversarial network (DCGAN) to synthesize minority crash data which could take into consideration the correlations between risk factors. The results demonstrated that the crash prediction model using DCGAN-generated balanced data achieved better performance in terms of both prediction accuracy (e.g., sensitivity and specificity), and distribution consistency compared to those using SMOTE and RU techniques. However, it should be noted that VAE and GANs are primarily designed for continuous variables. While they can handle discrete data to some extent by converting them into continuous variables through one-hot encoding, their assumptions and objective functions are still based on the continuous domain, which may lead to suboptimal performance for discrete data (Xu *et al.* 2019, Ma *et al.* 2020).

To this end, some researchers have developed advanced generative models which are applicable to both continuous and discrete variables. For instance, Table GAN (Park et al. 2018) combines a deep convolutional GAN with a classifier neural network to ensure the values in generated records are consistent. Medical Wasserstein GAN (MedWGAN) leverages gradient penalty and boundary seeking GAN to generate more realistic synthetic patient records (Baowaly et al. 2019). Variational auto-encoder for heterogeneous mixed-type data (VAEM) uses a two-stage training procedure to generate mixed-type data with heterogeneous marginals and missing data. In VAEM, latent variables can be conditioned on a categorical variable to generate new samples with specific categorical values (Ma *et al.* 2020). However, deep generative models may struggle with handling discrete variables, particularly when they are sparse. This is because the input of the generator in a GAN is a vector sampled from a standard multivariate normal distribution (MVN), and the issue of class imbalance is not considered. In light of this drawback, we propose a data resampling method based on Conditional Tabular GAN (CTGAN) to handle sparse discrete variables in tabular data. This is achieved through a conditional generator and a training-by-sampling strategy, which enables the generation of synthetic data by learning the conditional distribution of the data based on its specific value at each variable. The generator and discriminator in CTGAN are both built upon deep neural networks, enabling them to capture all possible associations between variables (Xu *et al.* 2019).

## 3 METHODOLOGY

### *3.1 Crash Severity Modeling Framework*

In this study, we aim to improve and evaluate the performance of crash severity modeling by addressing data imbalance challenges using the proposed deep learning generative method. The design framework is depicted in Fig. 1. The input data consists of traffic crash risk factors with multiple data types, which have been divided into training and testing sets for training and evaluating our model, respectively. We develop our data resampling approach based on CTGAN, with a comprehensive explanation provided in Section 3.2. Then, to demonstrate the performance of the proposed data generation method, a comparative study is conducted by comparing the proposed data generation method with the selected baseline over-sampling models. Furthermore, we conduct a comparative assessment of the proposed data generation method against various data resampling techniques, encompassing over-sampling, under-sampling, and mixed-sampling. The classification accuracy in crash severity modeling serves as a pivotal metric for evaluating the influence of the resampled data, ultimately guiding the selection of the optimal model based on accuracy. To guarantee the interpretation performance of the resampled dataset generated by our approach, we employ experiments using real datasets, and Monte Carlo simulations using simulated datasets in three scenarios (i.e., two-class imbalance, three-class imbalance, and various resampling ratios). These datasets are assessed in terms of classification accuracy, distribution consistency, and parameter recovery. The output of this study yields more accurate and reliable insights into the significant risk factors and their effects on crash severity.

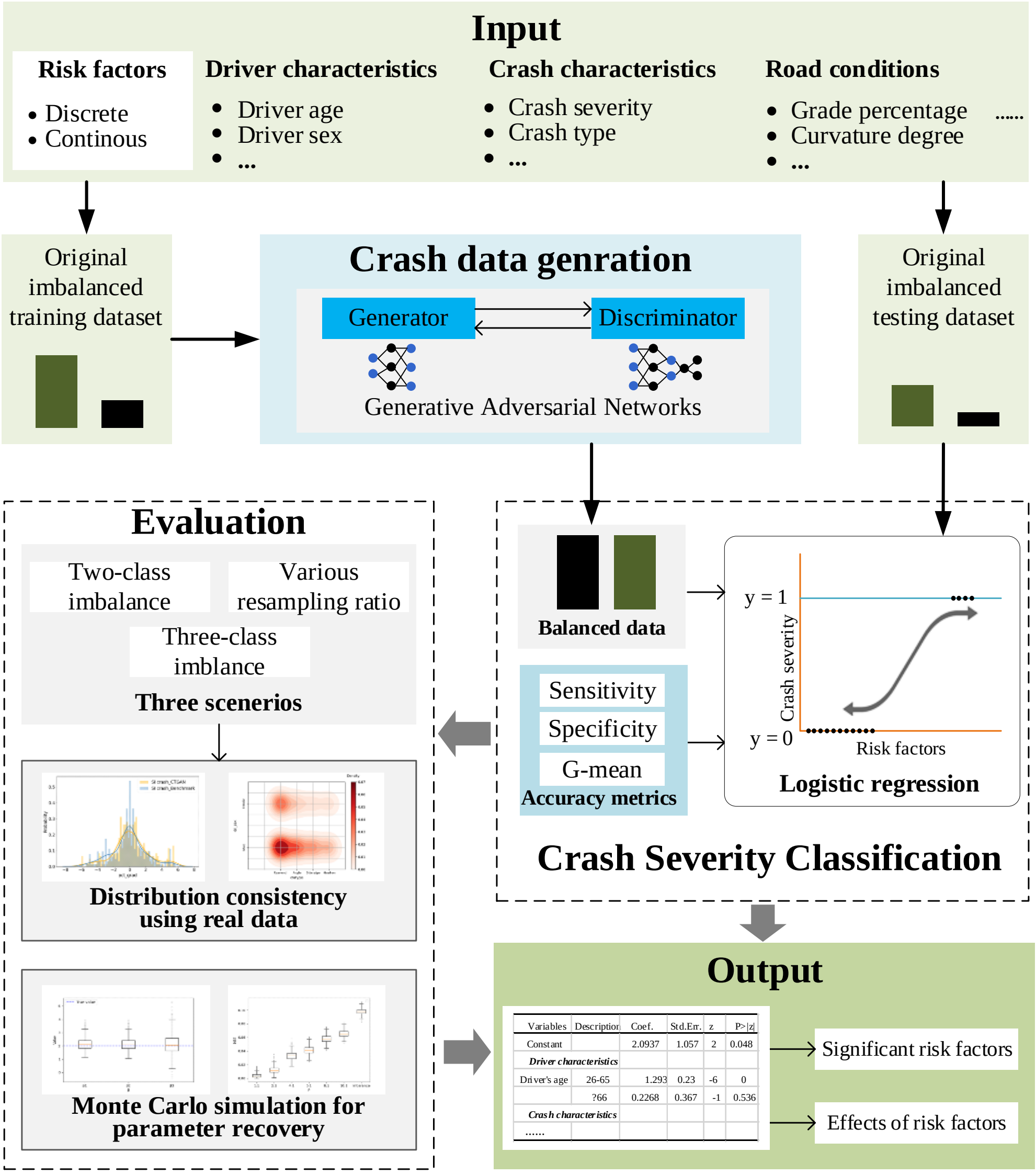

Fig. 1 Crash severity modeling framework

*3.2 Proposed Generation Method*

The Generative Adversarial Network (GAN) consists of two competing networks: generator network G and discriminator network D to generate samples through the adversarial training process (Goodfellow *et al.* 2014). Generator G generates data that might “fool” the discriminator by capturing the distribution characteristics of data while discriminator D is to distinguish whether the input sample comes from the original data set or from generator G. However, the ordinary GAN encounters some challenges when dealing with crash severity data where there are numerous discrete variables, such as crash type, driver information, and weather conditions. To address these issues, CTGAN introduces the mode-specific normalization mechanism where the discrete variables are represented by one-hot vectors. Conditional tabular generative adversarial networks (CTGAN) is a GAN-based approach for modelling tabular data distributions and sample rows in distributions which can learn the attributes of the fatal crashes well and thus ensure the reliability of synthetic crash samples (Xu *et al.* 2019).

In more detail, the distribution of each continuous risk factor is fitted independently using the variational Gaussian mixture (VGM) model. VGM is aimed at finding the best $k$ Gaussian models to represent the crash data through expectation maximization (EM), and the optimal number $k$ can be determined by weight advance. Through the calculation of the probability density of crash data, CTGAN samples the mode and then uses the sampled mode to normalize the values. During this process, the parameter $\beta$ is used to indicate which distribution the crash sample belongs to, and the parameter $\alpha$ is used to measure the value of the sample in this mode. Thus, the continuous risk factors are simply expressed by giving the values of the parameter $\alpha$ and $\beta$. The representation of a row in the crash data can be the concatenation of continuous and discrete risk factors as shown in Eq. (1).

$$r_j = \alpha_{1,j} \oplus \beta_{1,j} \oplus \ldots \oplus \alpha_{N_c,j} \oplus \beta_{N_c,j} \oplus d_{1,j} \oplus \ldots \oplus d_{N_d,j} \quad (1)$$

where $j$ is the index of a row in the crash severity data, $r_j$ is the $j$-th $N_c$ and $N_d$ denote the number of the continuous and discrete variables, respectively. $d_{i,j}$ is the one-hot representation of a discrete value. $\oplus$ is used to concatenate vectors.

CTGAN is also capable of addressing data imbalance for discrete risk factors, by minimizing the distribution discrepancy of discrete risk factors in generated and raw samples. When training a GAN generator, the input noise comes from a prior distribution (usually a standard multivariate normal distribution). The GANs' model does not account for imbalances in discrete risk factors, which in turn leads to the under-representation of crash samples for secondary categories of risk factors. The solution proposed by CTGAN is able to take information from discrete risk factors as input and learn to map the input to the desired output accordingly. This solution consists of three key elements: conditional vectors, generator loss, and sample training. Specifically, the condition vector is a one-hot encoding that includes all discrete risk factors, except that one category in the selected discrete risk factor has a value of 1, and the values of the other discrete risk factors are all 0. Conditional vectors contain information about discrete risk factors that force the generator to generate crash samples with a distribution of discrete variables similar to the training data. Sampling training crash samples conditional vectors and training crash data according to the log frequency of each class so that CTGAN can uniformly explore all possible discrete risk factors. The generator loss is used to enforce the generator to generate crash samples under the selected condition by adding the cross-entropy between the condition vector and the generated samples to the loss term. Fig. 2 displays the structure of the CTGAN model. The conditional generator can generate synthetic crash samples that are conditional on one of the discrete risk factors. In this paper, CTGAN is proposed as an over-sampling method to generate fatal crash samples.

In order to generate high-quality synthetic data, it is important to consider the interactions between different risk factors. The generator and discriminator in CTGAN each contain four fully connected layers, which enable the model to capture all possible associations between the risk factors and generate more realistic data. In the generator, the batch-normalization method and $ReLU$ activation function are used. Synthetic crash data is generated after two hidden layers via hybrid activation functions. The scalar value $\alpha_i$ is generated by $tanh$ while the mode indicator $\beta_i$ and the discrete value $d_i$ are generated by Gumbel SoftMax. The conditional generator $G(z, cond)$ can finally be expressed as in Eq. (2).

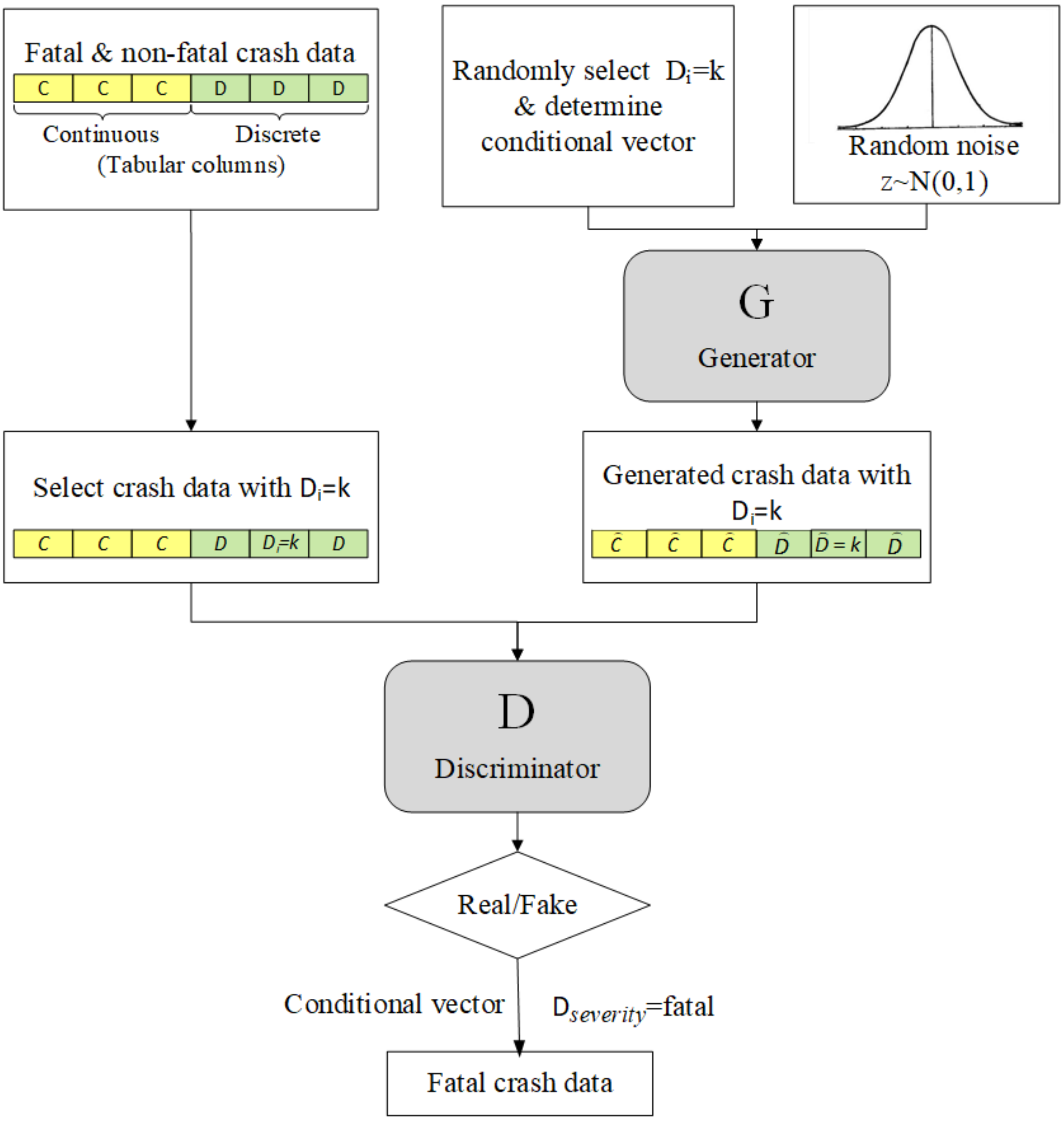

Fig. 2 Structure of CTGAN model

$$
\begin{cases}
h_0 = z \oplus cond \\
h_1 = h_0 \oplus \mathrm{Re}LU(BN(FC_{|cond|+|z| \to 256}(h_0))) \\
h_2 = h_1 \oplus \mathrm{Re}LU(BN(FC_{|cond|+|z|+256 \to 256}(h_1))) \\
h_3 = leaky_{0.2}(BN(FC_{|cond|+|z|+512 \to 256}(h_2))) \\
h_4 = leaky_{0.2}((FC_{256 \to 128}(h_3))) \\
\hat{\alpha}_i = \tanh(FC_{128 \to 1}(h_4)) \quad 1 \le i \le N_c \\
\hat{\beta}_i = gumbel_{0.2}(FC_{128 \to m_i}(h_4)) \quad 1 \le i \le N_c \\
\hat{d}_i = gumbel_{0.2}(FC_{128 \to |D_i|}(h_4)) \quad 1 \le i \le N_d
\end{cases}
\tag{2}
$$

where $z$ is noise from a standard normal distribution, $cond$ is the conditional vector, $\oplus$ is used to concatenate vectors. *ReLU* (Rectified Linear Unit), SoftMax, *tanh* (Hyperbolic tangent), *BN* (Batch Normalization) are used for batch normalization (Ioffe and Szegedy 2015). $leaky_\varepsilon(x)$ (Srivastava *et al.* 2014) refers to the application of a leaky *ReLU* activation function the input $x$ with a specified leaky ratio ε. $FC_{n \to m}(x)$ refers to the application of a linear transformation on a $n$-dimensional input to a $m$-dimensional output. $gumbel_\gamma(x)$ refers to the application of Gumbel SoftMax (Jang *et al.* 2016) with parameter $\gamma$ on a vector $x$.

In the discriminator, the leaky *ReLU* function and dropout technique (Srivastava *et al.* 2014) are applied to each hidden layer. The PacGAN (Jordon et al. 2019) framework is used to prevent mode collapse, where there are 10 crash samples in each Pac (i.e., mini-batch). The Pac determines the number of real samples included in each mini-batch during the discriminator training process, whose value is usually set to a small number, such as 10, to encourage diversity in the generated samples and prevent mode collapse. Specifically, the discriminator model which examines multiple crash samples in combination rather than in isolation helps to prevent mode collapse by promoting the diversity of the generated crash samples. By randomly selecting a small number of real crash samples to include in each mini-batch, CTGAN encourages the generator to produce a variety of different outputs rather than simply memorizing and reproducing the input crash data. (Salimans et al. 2016). The architecture of the discriminator (with Pac size 10) $D(r_1,\ldots,r_{10},cond_1,\ldots,cond_{10})$ is calculated in Eq. (3):

$$\begin{cases} h_0 = r_1 \oplus \ldots \oplus r_{10} \oplus cond_1 \oplus \ldots \oplus cond_{10} \\ h_1 = drop(leaky_{0.2}(FC_{10|r|+10|cond| \to 256}(h_0))) \\ h_2 = drop(leaky_{0.2}(FC_{256 \to 256}(h_1))) \\ h_3 = drop(leaky_{0.2}(FC_{256 \to 128}(h_2))) \\ h_4 = drop(leaky_{0.2}(FC_{128 \to 64}(h_3))) \\ D(\cdot) = FC_{64 \to 1}(h_4) \end{cases} \tag{3}$$

where $drop$ is used for dropout (Srivastava *et al.* 2014). The other symbols are defined the same way as in Eq. (1) and Eq. (2).

Table 1 provides the detailed architecture of the CTGAN model used in this study, which can be seen in Eqs. (2) and (3). The generator network and discriminator network of the model have four and five layers, respectively.

Table 1 CTGAN model architecture

| Generator architecture | | | Discriminator architecture | | |
|---|---|---|---|---|---|
| #Layer | Layer type | Layer description | #Layer | Layer type | Layer description |
| 1 | Fully connected + BN +ReLU | #Neuron = \|cond\|+\|z\| Output size = \|cond\|+\|z\|+256 × 1 | 1 | Fully connected +LeakyReLU(0.2) +Dropout | #Neuron = 256 Output size = 256 × 1 |
| 2 | Fully connected + BN +ReLU | #Neuron = \|cond\|+\|z\|+256 × 1 Output size = \|cond\|+\|z\|+512× 1 | 2 | Fully connected +LeakyReLU(0.2) +Dropout | #Neuron = 256 Output size = 256 × 1 |
| 3 | Fully connected + BN +LeakyReLU(0.2) | #Neuron = \|cond\|+\|z\|+512× 1 Output size =256 × 1 | 3 | Fully connected +LeakyReLU(0.2) +Dropout | #Neuron = 256 Output size =128 × 1 |
| 4 | Fully connected + BN +LeakyReLU(0.2) | #Neuron = 256 Output size = 128 × 1 | 4 | Fully connected +LeakyReLU(0.2) +Dropout | #Neuron = 128 Output size = 64 × 1 |
| | | | 5 | Fully connected + Sigmoid | #Neuron = 64 Output size = 1 |

Various numbers epochs, namely 20, 50, 100, 200, 400, and 600, are evaluated and 200 is ultimately chosen. To achieve favorable training performance, Hensel *et al.* (2017) provided mathematical proof of convergence to the Nash equilibrium and demonstrated that using different learning rates for the generator and discriminator is effective. For this study, learning rates of 0.0005 and 0.0001 are used for the discriminator and generator, respectively.

### *3.3 Baseline Resampling Methods*

The methods for balancing crash severity data can be broadly classified into three groups: over-sampling, under-sampling, and mixed-sampling. In this study, we employed state-of-the-art data resampling techniques from each category, namely synthetic minority over-sampling technique - nominal continuous (SMOTE-NC), tabular variational autoencoder (TVAE), random under-sampling (RU), and CTGAN-RU, which are capable of handling multi-formatted data containing both discrete and continuous variables.

**SMOTE-NC over-sampling**: Most SMOTE methods (such as SMOTE, Borderline-SMOTE, and polynom-fit-SMOTE) are developed in consideration of only continuous factors while they may fail to deal with discrete factors. SMOTE-NC, as a variant of the SMOTE technique, handles both continuous and discrete risk factors differently and retains the original tag for resampling the data center classification characteristics (Chawla *et al.* 2002b). Note that the SMOTE-NC method only works when there is at least one continuous attribute in the data set and may generate duplicated samples.

**TVAE over-sampling**: Variational autoencoder (VAE) is a neural network generative model consisting of two networks that model two probability density distributions (Diederik P Kingma 2013, Kingma and Welling 2019, Xu *et al.* 2019). VAE can capture the underlying distribution of the input data and generate new samples from that distribution by inferential network and generative network. TVAE is a variant of the VAE to adapt tabular data (Xu *et al.* 2019).

**Random under-sampling:** The under-sampling technique removes some samples in the majority class to achieve a balanced distribution of samples. Based on the under-sampling technique, the random under-sampling (RU) technique is to reduce the majority of samples randomly and uniformly. In this case, the majority of non-fatal crash data were randomly removed from total samples to achieve a balance between fatal crash data and non-fatal crash data. The main drawback of the RU is that it may discard potentially useful samples for the modelling process (Mujalli *et al.* 2016).

**CTGAN-RU mixed-sampling:** RU may eliminate useful information or omit noises leading to biased decision boundaries (Ha and Lee 2016). Although SMOTE reduces the possibility of overfitting, it also raises the overfitting problem due to the generation of incorrectly generated examples (Zhu et al. 2017). Some researchers have found that a combination of the two has a better performance compared to using oversampling or under-sampling alone (Chawla *et al.* 2002a). In this study, we proposed a mixed-sampling technique combining the CTGAN and under-sampling techniques, called CTGAN-RU. First, samples in the majority class, i.e., non-fatal crashes, are randomly selected with even possible by the RU. Then, the selected non-fatal crashes and the real fatal crashes are used to generate the synthetic FI crashes by CTGAN.

### *3.4 Crash Severity Classification*

Crash severity classification modeling is conducted to verify the performance (i.e., classification accuracy and interpretative performance) of synthetic data using the proposed data generation method. Since crash severity in this study is dichotomous, the binary logit regression is used for analyzing the relationship between outcome variables of binary and risk variables (Young and Liesman 2007, Tamakloe et al. 2022). Considering two types of crash severity with the probability $p$ for fatal crashes ($y = 1$) and $1 - p$ for non-fatal crashes, the utility function can be specified as shown in Eq. 4.

$$\begin{gathered} Y \sim \text{Bernoulli}(p) \\ \text{Logit}(p) = \log\left(\frac{p}{1-p}\right) = \beta_0 + X^{'}\beta \end{gathered} \tag{4}$$

where $\beta_0$ is the intercept, $X$ is the vector of the risk factors, $\beta$ is the vector corresponding to regression coefficients to be estimated.

*3.5 Evaluation Metrics*

In order to evaluate the accuracy of each crash severity class, three performance indicators, sensitivity, specificity, and G-mean, are adopted. Sensitivity is defined as the proportion of correctly classified positive events, which is in line with the main goal of this study - correctly predicting minority fatal crashes in the crash classification, so it is one of the important indicators of the classifier. Specificity is the ratio of correctly classified negative events over the total observed negative events, that is, the ratio of correctly predicted non-fatal crashes over the total observed non-fatal crashes. The equations that define these two indicators are:

$$Sensitivity = \frac{TP}{TP + FN} \tag{5}$$

$$Specificity = \frac{TN}{TN + FP} \tag{6}$$

where the True Positive (TP) means the number of fatal crashes that are correctly classified, and the False Positive (FP) means the number of non-fatal crashes that are incorrectly classified as fatal crash events. False Negative (FN) indicates the number of incorrectly classified non-fatal crash events, and True Negative (TN) indicates the number of correctly classified non-fatal collision events.

The G-mean is considered a measure of stability that independently looks at the correct classification of positive and negative classes. It attempts to maximize the accuracy of both classifications through good balance and is particularly suitable for treating unbalanced data sets. When the sensitivity and specificity are high and the difference between them is small, the value of the G-mean is large, indicating that the classifier has better performance. G-mean is described as follows:

$$G-mean=\sqrt{Sensitivity\cdot Specificity} \tag{7}$$

## 4 DATA PREPARATION

### *4.1 Data Description*

This study focuses on the crashes that occurred on interstate and highway segments in Washington State (e.g., I-5, I-90, U.S. 2, U.S. 12, SR 20) from January 2014 to December 2018 (Nujjetty 2014). The dataset includes variables related to crash characteristics, roadway characteristics (e.g., basic road, curve, grade), driver and vehicle characteristics, etc. Since the crash data was collected statewide by all police departments on a report form, the raw dataset contains missing data and very detailed crash information. A set of rules is established for the variable selection process: (1) variables should have no missing values; (2) variables should be independent of each other; (3) variables that have few correlations with the crash severity outcomes should be excluded.

Finally, 14 variables are retained of which the summary statistics have been shown in Table 2. The multi-collinearity test is conducted to evaluate the correlations between the independent variables. Results indicate that the variance inflation factor (VIF = 1.011) is less than five for all independent variable pairs. This finding confirms the suitability of all candidate variables for the following logistic regression modeling.

As can be seen, the majority of variables are discrete (e.g., driver's age, type of collision), while some variables related to the characteristics of the road segments are continuous (e.g., curvature degree, grade percentage). In this study, fatal crashes include dead at the scene, dead on arrival to the hospital, and died at the hospital, and non-fatal crashes contain other types of crash severity documented in the crash report (e.g., no injury, possible injury, non-severe injury, and severe injury). One can observe that the dataset is highly imbalanced where fatal crashes only made up 0.05% of total crashes while non-fatal crashes accounted for 99.95%.

Table 2 Summary of variables statistical description

| Variables (discrete) | Description | Count | Mean | Non-fatal crash | Fatal crash |
|---|---|---|---|---|---|
| ***Driver characteristics*** | | | | | |
| Driver's_age (dri_age) | | | | | |
| ≤25 | ≤ 25 = 1; Others = 0 | 31557 | 0.201 | 31533 | 24 |
| 26-65 | 26–65 = 1; Others = 0 | 115760 | 0.737 | 115712 | 48 |
| ≥66 | ≥66 = 1; Others = 0 | 9763 | 0.062 | 9754 | 9 |
| Driver's_sex (dri_sex) | | | | | |
| Male | Male = 1; Female= 0 | 98811 | 0.629 | 98761 | 50 |
| Drunk_driving (drunk_dri) | | | | | |
| Drunk | Drunk driving = 1; Not = 0 | 1523 | 0.01 | 1506 | 17 |
| Drug driving (drug_dri) | | | | | |
| Drug | Drug driving = 1; Not = 0 | 604 | 0.004 | 594 | 10 |
| ***Crash characteristics*** | | | | | |
| Type_of_collision (cratype) | | | | | |
| Rear-end | Rear-end collision = 1; Others = 0 | 115851 | 0.738 | 115820 | 31 |
| Angle | Angle collision = 1; Others = 0 | 8085 | 0.051 | 8067 | 18 |
| Sidewipe | Sidewipe collision = 1; Others = 0 | 32731 | 0.208 | 32718 | 13 |
| Head-on | Head-on collision = 1; Others = 0 | 413 | 0.003 | 394 | 19 |
| Airbag_status (airbag) | | | | | |
| Active | Airbag active = 1; Others = 0 | 10251 | | 10210 | 41 |
| ***Temporal characteristics*** | | | | | |
| Time_of_day (time) | | | | | |
| Peak hour | Peak hours = 1; Others = 0 | 82455 | 0.525 | 82429 | 26 |
| Day_of_week (weekday) | | | | | |
| Weekend | Weekend = 1; Others = 0 | 28702 | 0.183 | 28685 | 17 |
| ***Road characteristics*** | | | | | |
| Location (rur-urb) | | | | | |
| Urban | Rural = 1; Urban = 0 | 145414 | 0.926 | 145368 | 46 |

| Variables (continuous) | Non-fatal crash | | | | Fatal crash | | | |
|---|---|---|---|---|---|---|---|---|
| | mean | std | min | max | mean | std | min | max |
| ***Road characteristics*** | | | | | | | | |
| Posted_speed_limit (mph) (spd_limit) | 58.646 | 5.842 | 30 | 70 | 60.988 | 6.728 | 35 | 70 |
| Curvature_degree (deg_curv) | 0.685 | 1.379 | 0 | 19.420 | 0.811 | 1.277 | 0 | 5 |
| Grade_percentage (%) (pct_grad) | -0.179 | 2.201 | -6.920 | 6.720 | 0.069 | 2.154 | -5.960 | 5.4 |
| ***Crash characteristics*** | | | | | | | | |
| Number_of_vehicles involved (num_vehs) | 2.368 | 0.668 | 1 | 14 | 2.494 | 0.793 | 2 | 5 |

Note: Peak hours are from 6:00 to 9:00, and 16:00 to 19:00
"+" represents an upgrade, a "-" represents a downgrade
"Std.dev" represents the standard deviation

### *4.2 Data Resampling*

In this study, three types of resampling techniques, namely, over-sampling (i.e., CTGAN, TVAE, SMOTE-NC), under-sampling (RU), and mixed-sampling (CTGAN-RU) are adopted to balance the original training dataset respectively. For the resampled training dataset, the fatal to non-fatal crashes ratio is 1:1. It should be noted that the original test dataset remained unchanged when evaluating the trained models to get the best possible estimate of developed models when used on entirely new data.

The details of data organization for algorithm training and testing are summarized in Table 3, containing the original and the balanced datasets. 70% of the crash data (109,956) is randomly chosen for training and the remaining 30% is used for testing (47,124). In the over-sampling scenario, 109,842 synthetic fatal crashes are generated using CTGAN, SMOTE-NC, or TVAE. Thus, the total over-sampling training dataset consists of the original dataset and the synthetic fatal crashes. In the under-sampling scenario, the RU techniques both selected 57 non-fatal crashes, which is equal to the number of fatal crashes in the original training dataset. In the mixed-sampling scenario, fatal crashes are generated and non-fatal crashes are dropped simultaneously. In more detail, 57 fatal samples are synthetic and the number of non-fatal crashes decreased to 114 using CTGAN-RU.

Table 3 Experimental design

| Data type | Original total dataset | Original test dataset | Original training dataset | Over-sampling training | Under-sampling training | Mixed-sampling |
|---|---|---|---|---|---|---|
| Non-fatal | 156,999 | 47,100 | 109,899 | 109,899 | 57 | 114 |
| Fatal | 81 | 24 | 57 | 109,899 | 57 | 114 |
| Total | 157,080 | 47,124 | 109,956 | 219,798 | 114 | 228 |

# 5 NUMERICAL RESULTS

## *5.1 Crash Severity Classification Accuracy*

Table 4 illustrates the evaluation results of five types of resampling methods in terms of sensitivity, specificity, and G-mean. The binary logit model trained with the synthetic dataset exhibits higher sensitivity and G-mean, but lower specificity than those trained with the original training data. The evaluation results demonstrate better classification performance for non-fatal crashes than for fatal crashes in both the original and resampling. Specifically, in the original training dataset, there is a significant bias for non-fatal, and no fatal is predicted correctly due to the imbalance of the two types of crashes. It is revealed that the data resampling techniques indeed improve the model performance in the fatal crashes (the minority class) and slightly degrade that in the non-fatal crashes (the majority class) since fatal crashes have more representation. One can see that the binary logit model based on the synthetic data balanced by CTGAN-RU has the highest G-mean values (0.851) suggesting that it has the best performance overall for two types of crash severity. It should be noted that although this model's specificity (0.827) is not the highest one, and more importantly, G-mean is a more suitable metric to reflect the general performance of the sensitivity and specificity.

Among the various data resampling techniques, the CTGAN method produced substantially higher sensitivity rates than both the SMOTE-NC and TVAE methods, albeit lower than the RU method. Meanwhile, the RU method exhibited notably higher specificity rates compared to over-sampling methods, but conversely, also demonstrated the lowest specificity among the resampling methods. By combining the strengths of CTGAN and RU, the CTGAN-RU approach implemented using the binary logit model effectively enhances the performance of both minority and majority cases. Thus, based on the predictive capability of the test datasets, the CTGAN-RU approach is preferred for data balancing in the context of crash data classification and interpretation.

Table 4 Classification accuracy of FI and nFI crash severity models

| Measurements | Original training dataset | CTGAN | CTGAN-RU | SMOTE-NC | TVAE | RU |
|---|---|---|---|---|---|---|
| Sensitivity | 0.000 | 0.667 | **0.875** | 0.583 | 0.458 | 0.75 |
| Specificity | 1.000 | 0.840 | **0.827** | 0.912 | 0.890 | 0.678 |
| G-mean | 0.000 | 0.748 | **0.851** | 0.729 | 0.639 | 0.713 |

*5.2 Sensitivity Analysis of CTGAN-RU in Various Resampling Ratios*

In this section, we investigate the impact of the ratio of the CTGAN-RU on the performance of the model. Specifically, a range of experiments are conducted by varying the ratios of minority crashes (fatal ones) to majority crashes (non-fatal) during the under and over-sampling process using the logit model. The experiment designs during under- and over-sampling are detailed in Fig. 3, and the horizontal axis represents the ratio of non-fatal to fatal crashes in the original training data selected by RU. Fig. 3(1) illustrates the model performance by varying the ratios of fatal to non-fatal (1:1, 2:1, 5:1, 10:1, 20:1, 40:1, 50:1, 60:1, 80:1, 100:1), where the vertical axis represents the values of sensitivity, specificity, and G-mean distinguished by various colors. On the other hand, in Fig. 3(2) - (4), various colors represent the ratio of non-fatal to fatal crashes (1:1,2:1,3:1,4:1,5:1) generated by CTGAN based on RU. For example, in Fig. 3(2), the blue points enclosed by red circles represent a ratio of RU of 2:1 (non-fatal:114, fatal:57) and a ratio of CTGAN-RU of 1:1 (non-fatal:114, fatal:114).

In terms of sensitivity and G-mean, the performance of the crash severity model degrades with an increase in the number of non-fatal crashes selected by RU, even when the number of fatal crashes remains constant. This trend is illustrated in Fig. 3(1). Results indicate that the tendency of specificity, sensitivity, and G-mean of the logit model trained upon CTGAN-RU is similar to that of the logit model trained upon RU (Fig. 3(2)-(4)). The performance of the model is better for fatal crashes when more fatal cases are generated by CTGAN, while the opposite is observed for non-fatal crashes, as presented in Fig. 3 (2) and (3). A change in the ratio of the RU or CTGAN-RU had a greater effect on sensitivity than on specificity. In general, for the same RU ratio, a smaller CTGAN-RU ratio corresponds to better sensitivity and G-mean performance. The best performance is observed when the ratio of RU is 2:1 (0:114, 1:57) and CTGAN-RU is 1:1 (0:114, 1:114), with respect to the sensitivity (value = 0.875) of fatal crashes data and the G-mean (value = 0.851) of the whole crashes.

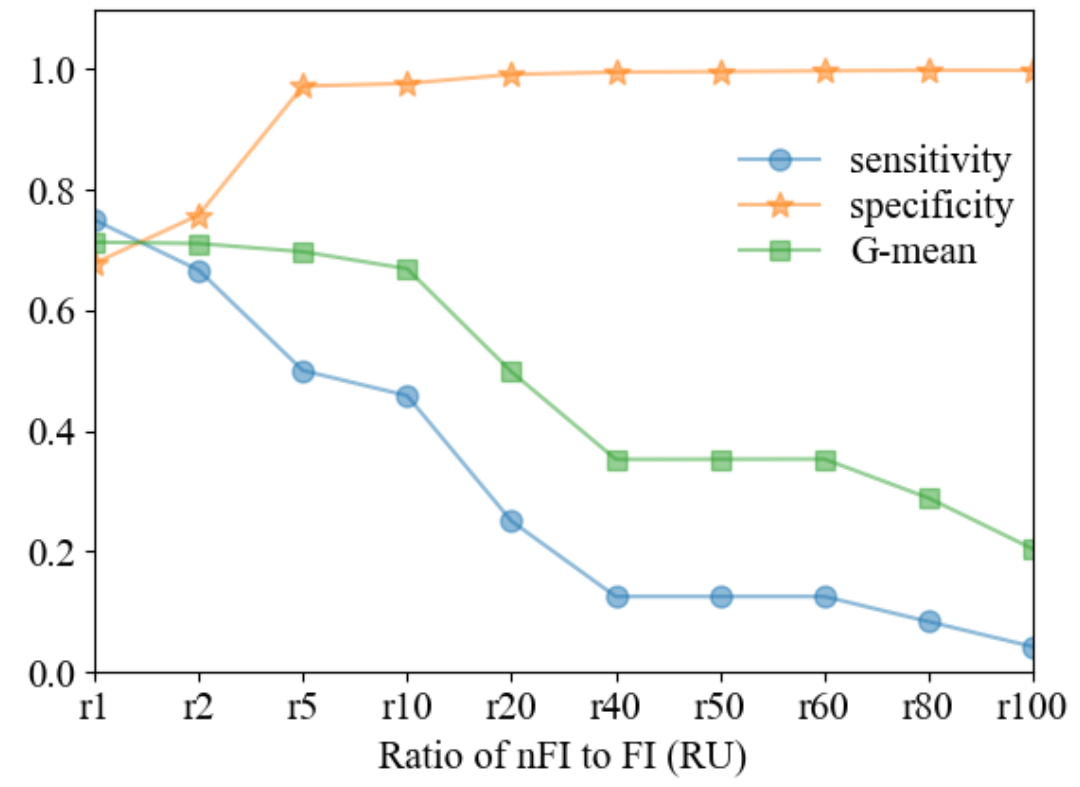

(1) RU (sensitivity, specificity, G-mean)

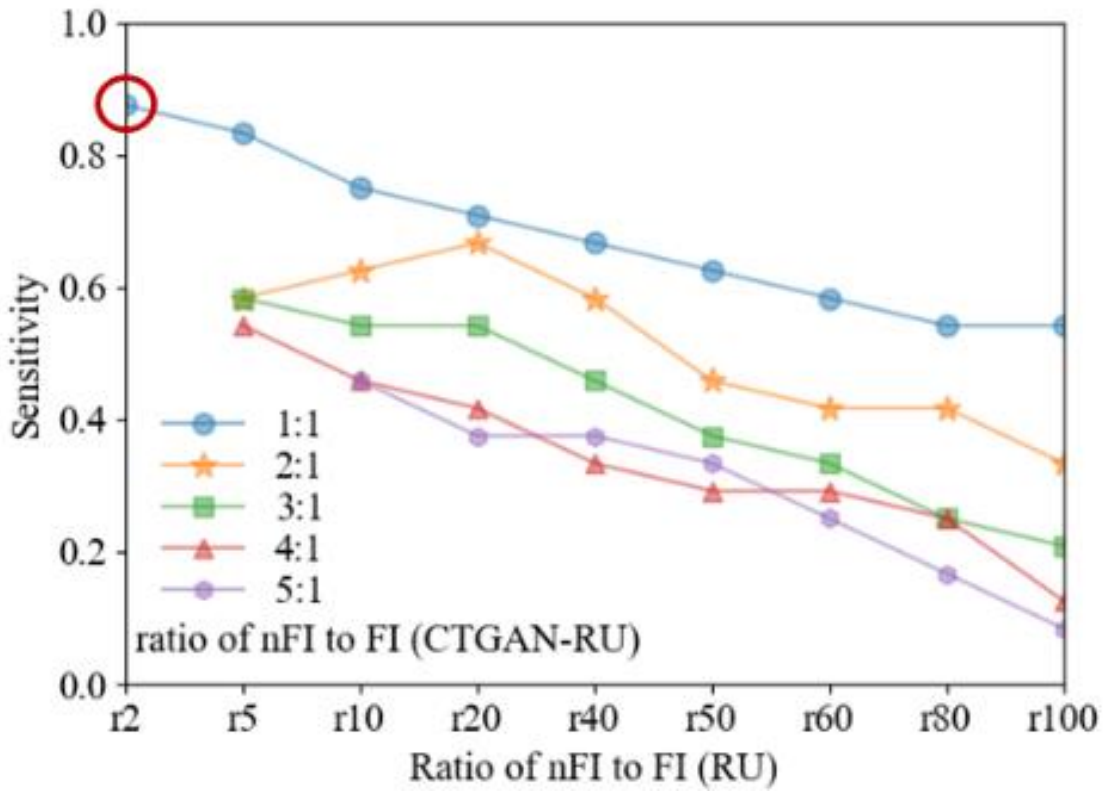

(2) CTGAN-RU (Sensitivity)

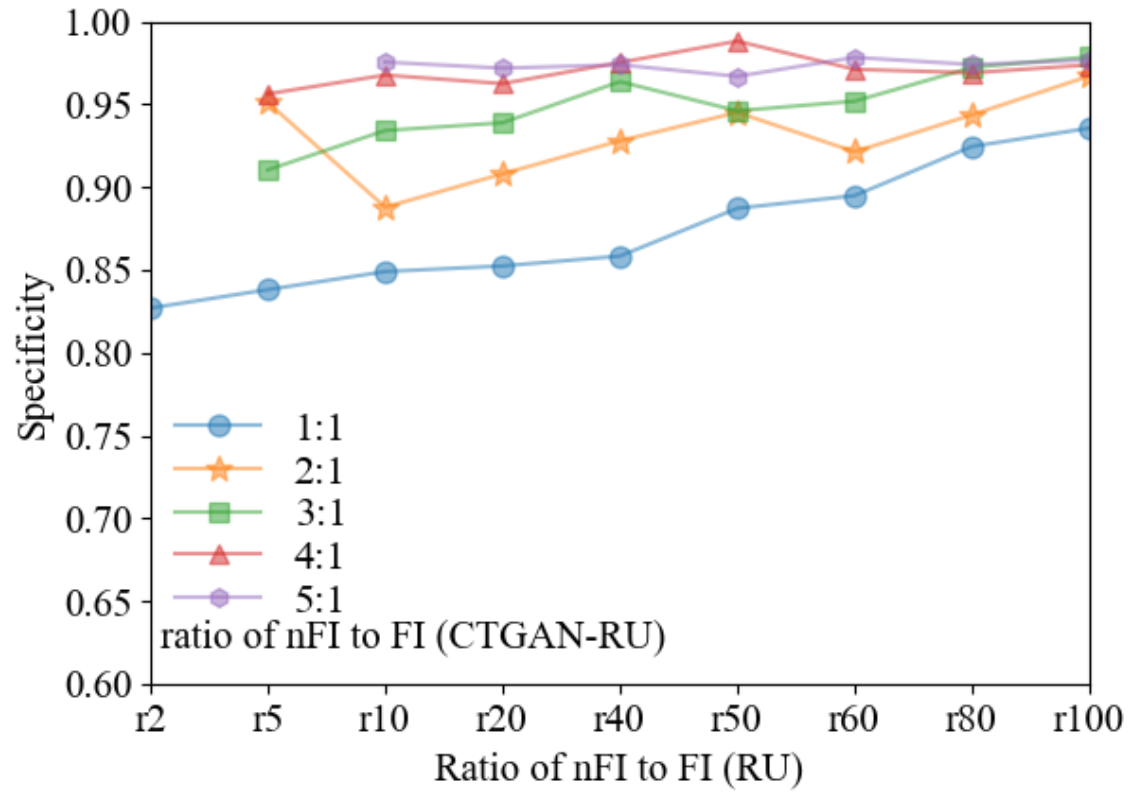

(3) CTGAN-RU (Specificity)

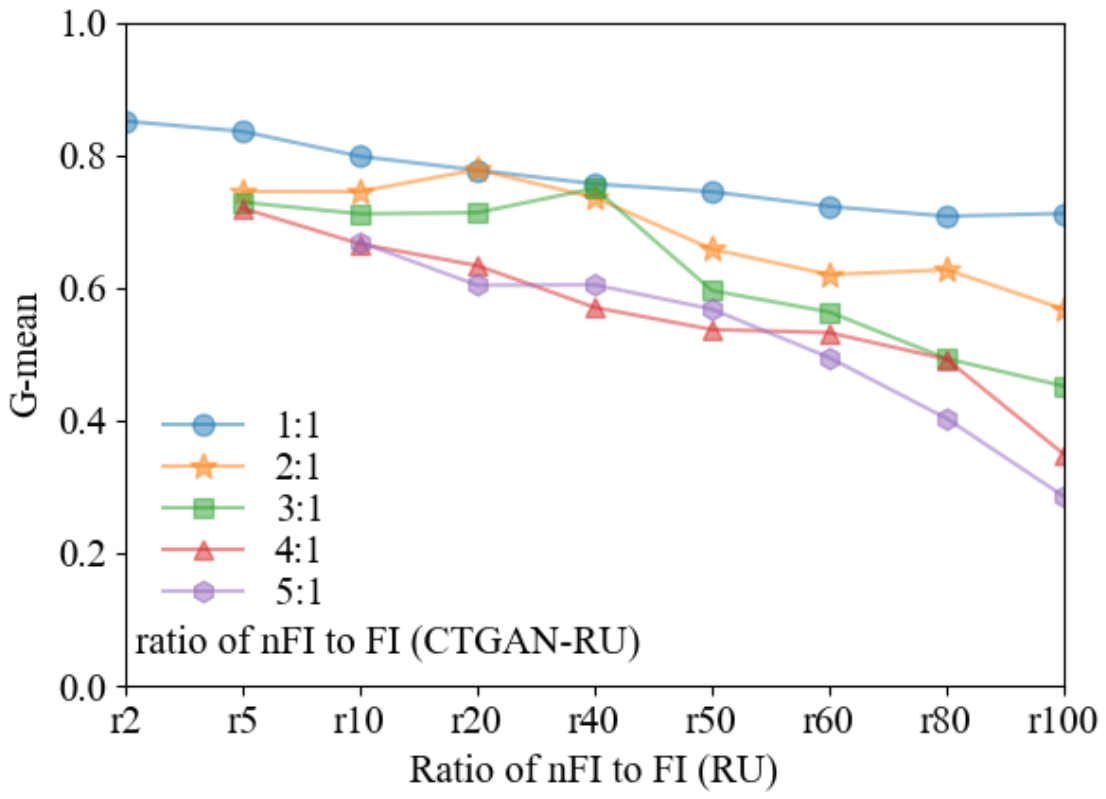

(4) CTGAN-RU (G-mean)

Fig. 3 Measurements of CTGAN-RU with various ratios

### *5.3 Distribution Consistency Evaluation*

In this subsection, we evaluate the distribution consistency of the dataset generated by CTGAN-RU in comparison to the balanced benchmark dataset across three distinct scenarios using real datasets. These scenarios include a two-class imbalance for binary logit modeling, various sampling ratios for binary logit modeling, and a three-class imbalance for ordered logit modeling.

#### *5.3.1 Two-class imbalance*

The details of two-class imbalance experiment designs for interpretative consistency evaluation are summarized in Fig.4, containing the balanced benchmark data and the resampled datasets. The balanced crash data is randomly selected from the real original total data containing 453 severe injury (SI) (i.e., severe injury and fatal crashes) and 453 non-severe injury (nSI) (property damage only, possible injury crashes, and non-severe injury crashes). 70% of the balanced crash data is randomly selected as the balanced benchmark data and the remaining 30% is reserved for testing. Subsequently, the imbalanced training and testing data is reduced to a ratio of 5:1 of the nSI and SI by randomly selecting SI from the balanced benchmark data. Resampled methods and correspondent resampled ratios are the same in Section 4.2. The number of observations for SI and nSI crashes is shown in Table 5.

Table 5 Number of observations for two-class imbalance (SI and nSI crashes)

| Data type | Balanced total dataset | Test dataset | Benchmark dataset | Imbalanced training dataset | Over-sampling training | Under-sampling training | Mixed-sampling training |
|---|---|---|---|---|---|---|---|
| nSI | 453 | 136 | 317 | 317 | 317 | 63 | 126 |
| SI | 453 | 27 | 317 | 63 | 317 (generated) | 63 | 126 (generated) |
| Total | 906 | 272 | 634 | 380 | 634 | 126 | 252 |

Table 6 summarizes the binary logit model's classification accuracy in terms of sensitivity, specificity, and G-mean. The results show that the performance of synthetic data using CTGAN-RU (0.914) is better than benchmark data (0.903) in terms of specificity and G-mean. However, the sensitivity, specificity, and G-mean of the model based on synthetic data using SMOTE-NC, and TVAE are remarkably lower than those using CTGAN or CTGAN-RU. This justifies the capability of the proposed data generation method.

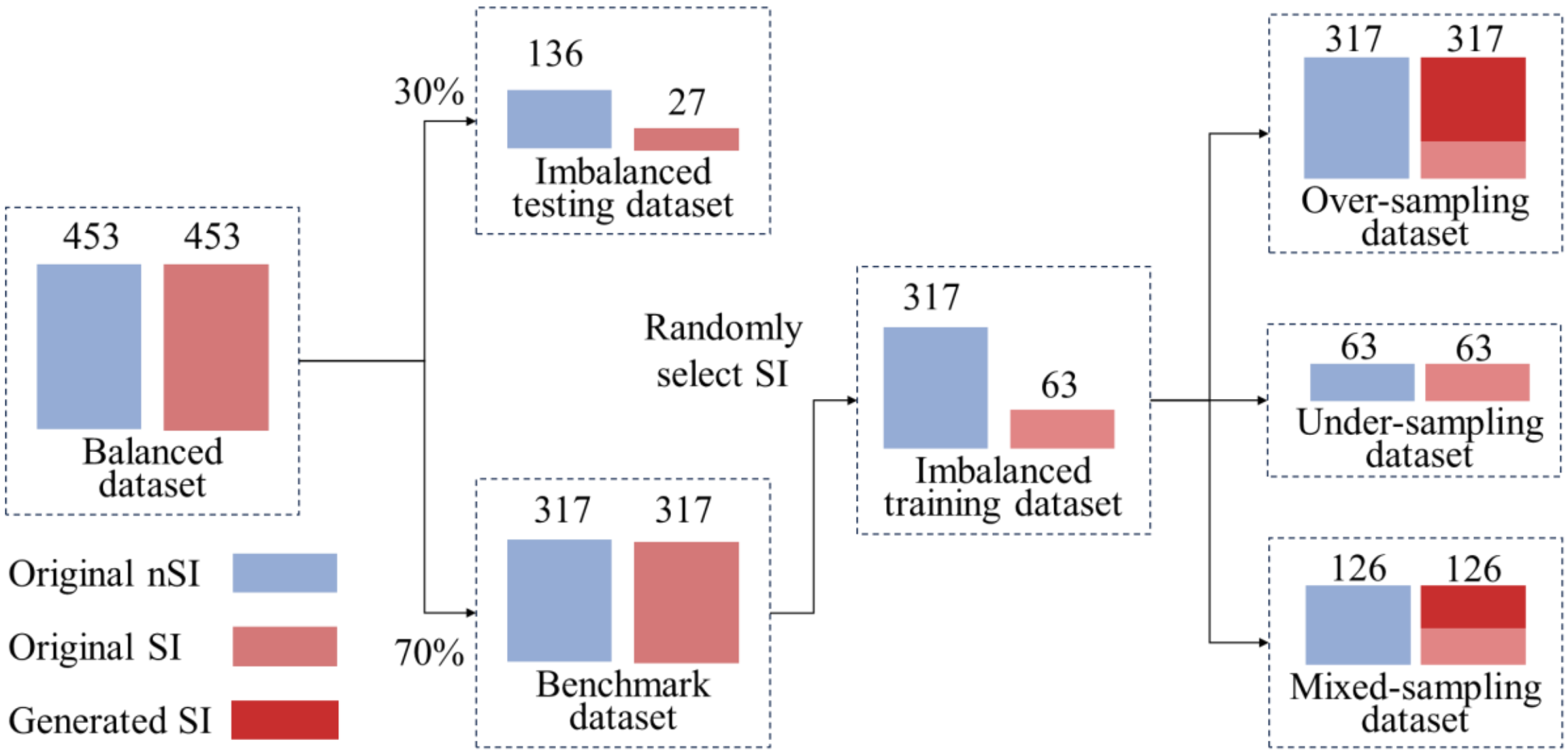

Fig.4 Experimental design for SI and nSI crashes

Table 6 Classification accuracy of binary logit modeling (SI and nSI crashes)

| Measurements | Benchmark dataset | CTGAN | CTGAN-RU | SMOTE-NC | TVAE | RU |
|---|---|---|---|---|---|---|
| Sensitivity | 1.000 | 1.000 | **0.963** | 0.889 | 0.593 | 0.815 |
| Specificity | 0.816 | 0.794 | **0.868** | 0.868 | 0.860 | 0.897 |
| G-mean | 0.903 | 0.891 | **0.914** | 0.878 | 0.714 | 0.855 |

Fig. 5(1)-(14) illustrates the distribution of balanced benchmark data and synthetic data by CTGAN, CTGAN-RU, SMOTE, TVAE, and RU. As shown in Fig. 5(1)-(14), deviations in the distribution of synthetic data based on the CTGAN and CTGAN-RU are much smaller than those based on SMOTE, TVAE, and RU for all variables. For instance, the distribution generated by SMOTE and TVAE has deficiencies, including abnormal points and distributed

concentration in some specific variables (e.g., continuous variables: Curvature degree and Grade percentage). The generated distributions of discrete variables, including sparse variables, are more accurate with CTGAN and CTGAN-RU compared to SMOTE and TVAE. Specifically, when considering the driver's age, where the age ranges of 18 to 25 and over 65 are relatively sparse compared to the age range of 25 to 65, the distributions generated by CTGAN and CTGAN-RU are comparable for all age ranges. These findings also suggest that CTGAN and CTGAN-RU can better capture the data characteristics containing sparse variables and provide a well-fitted distribution of risk factors for SI crashes.

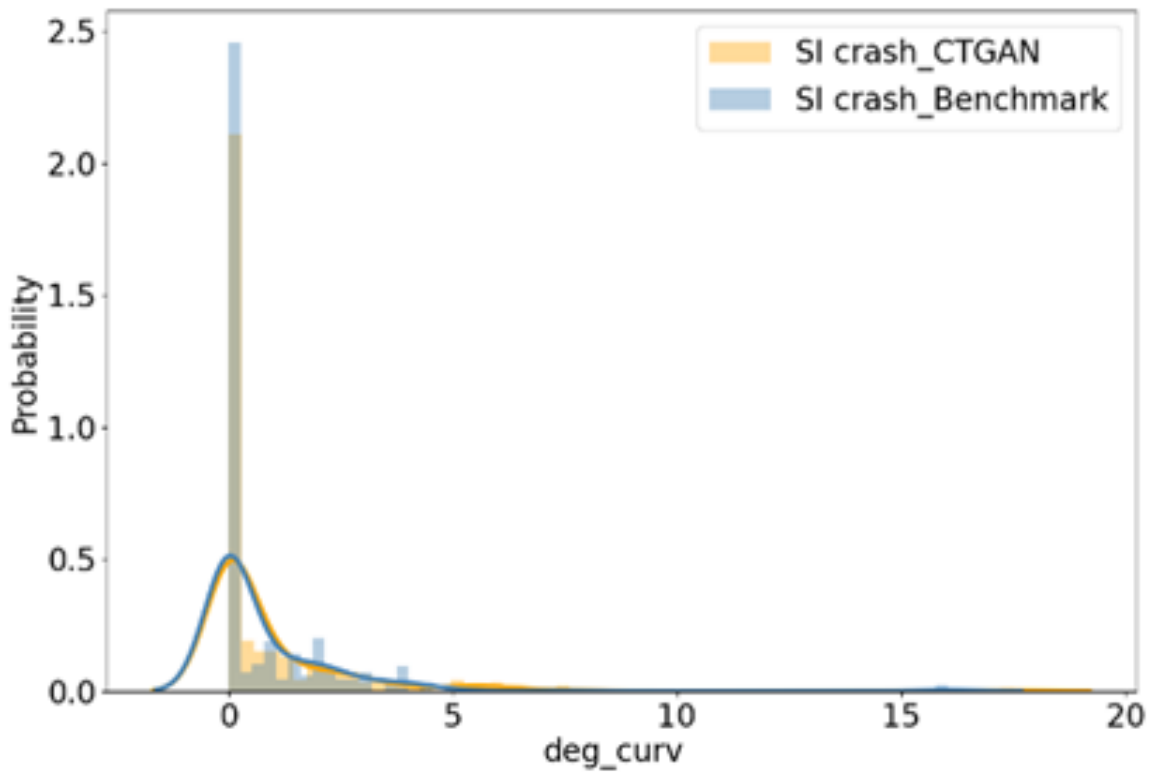

(1) Curvature degree (CTGAN)

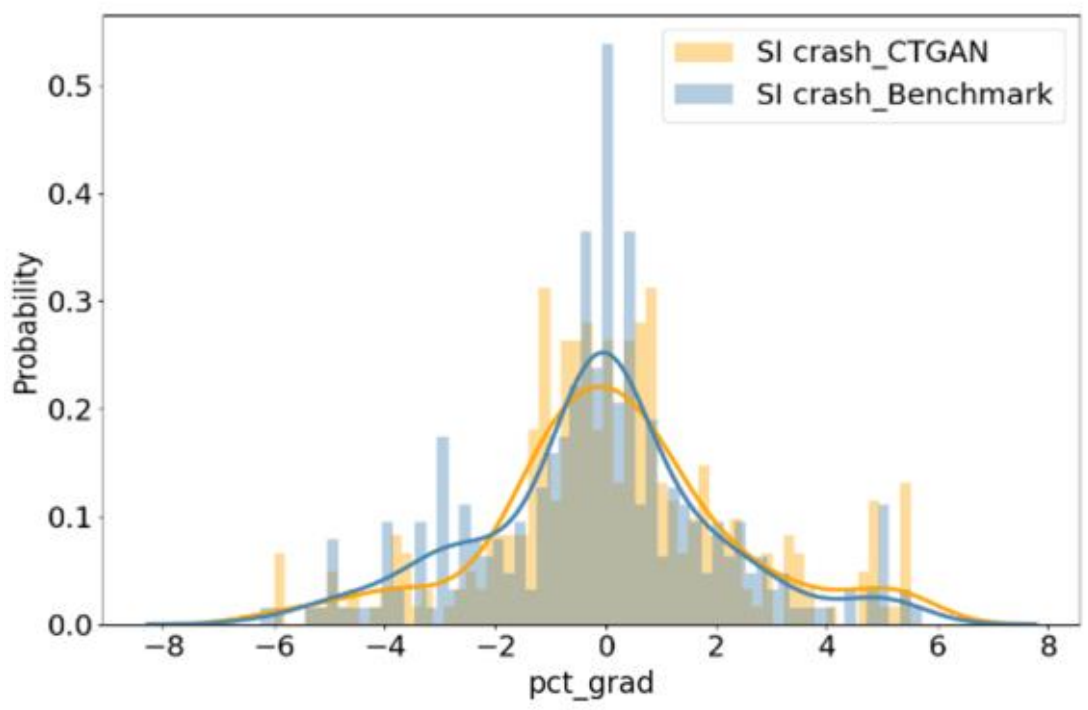

(2) Grade percentage (CTGAN)

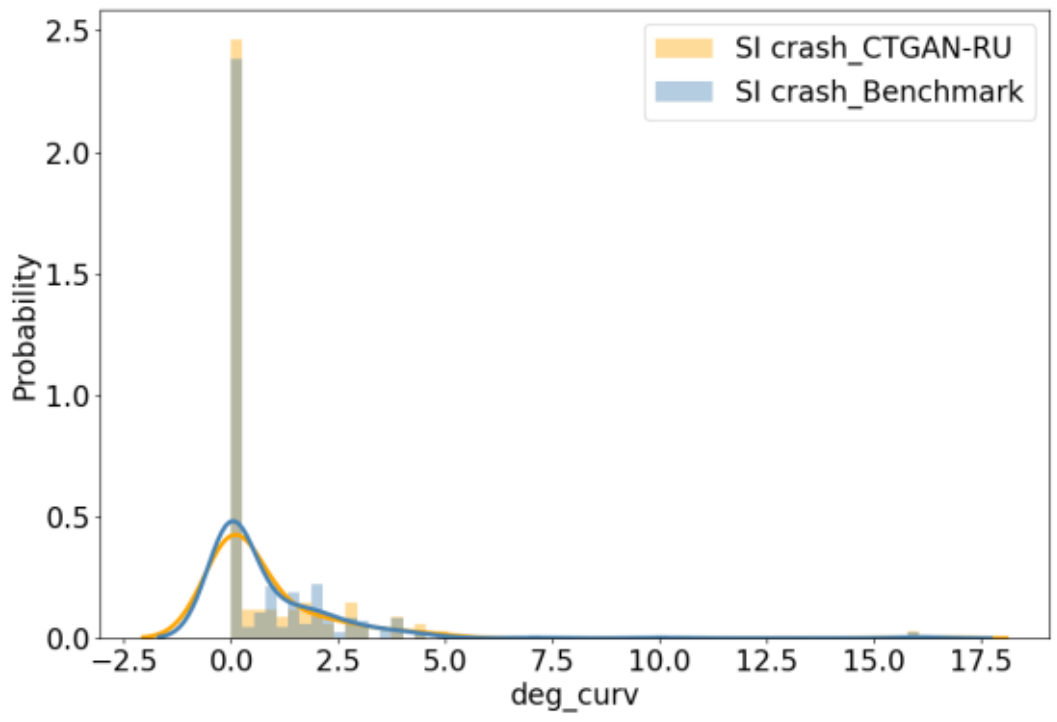

(3) Curvature degree (RU-CTGAN)

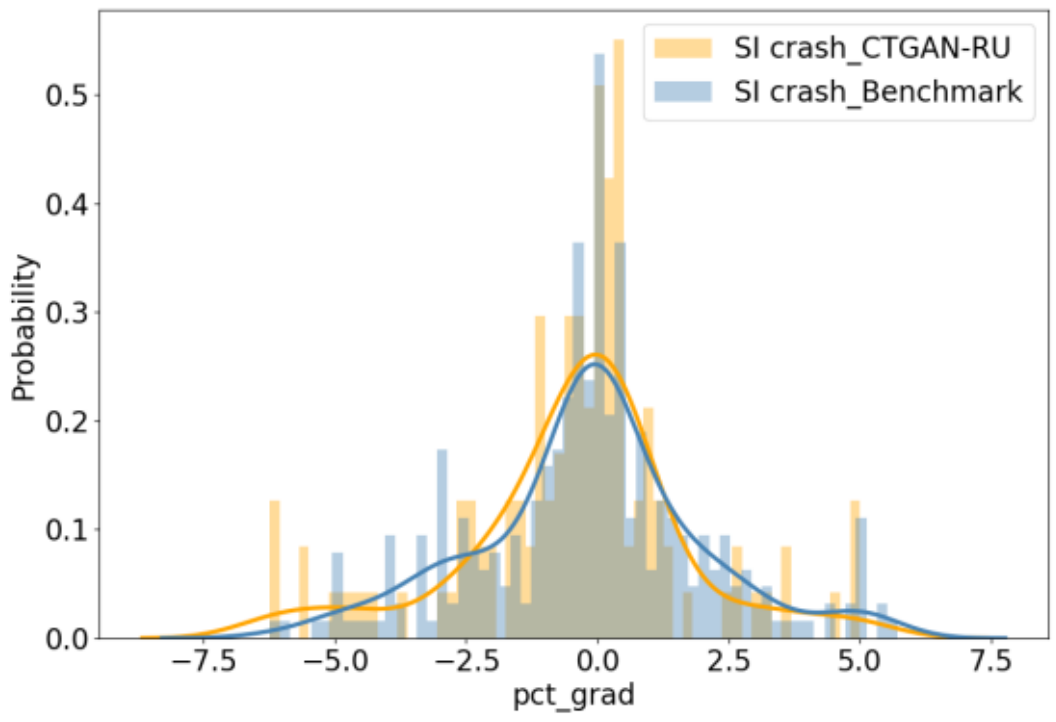

(4) Grade percentage (RU-CTGAN)

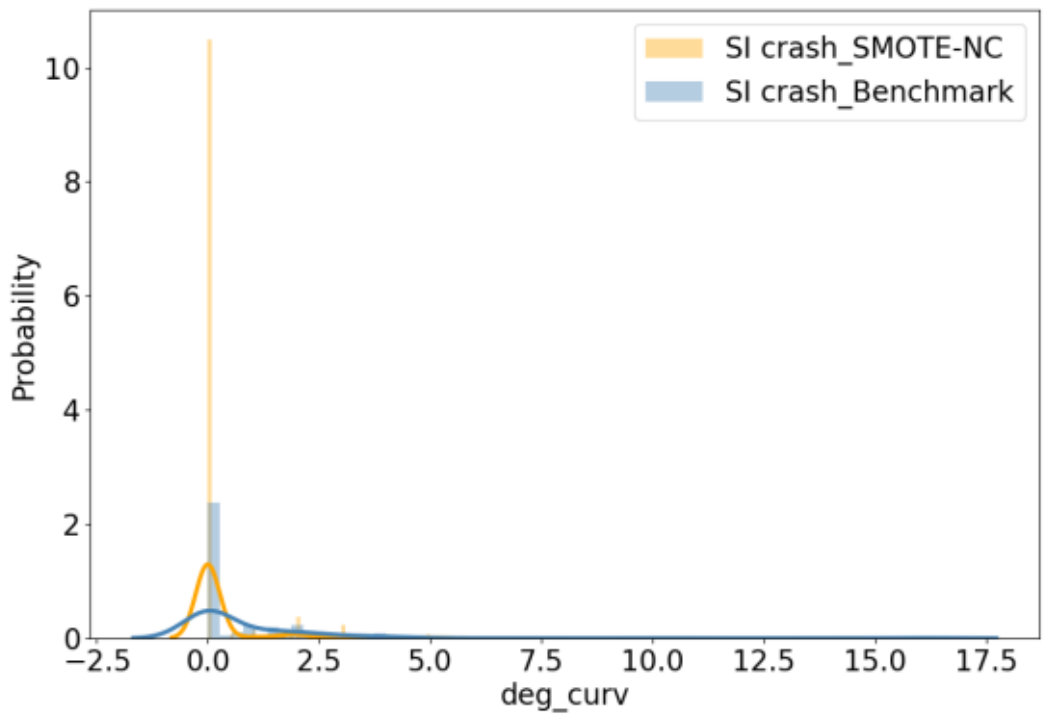

(5) Curvature degree (SMOTE-NC)

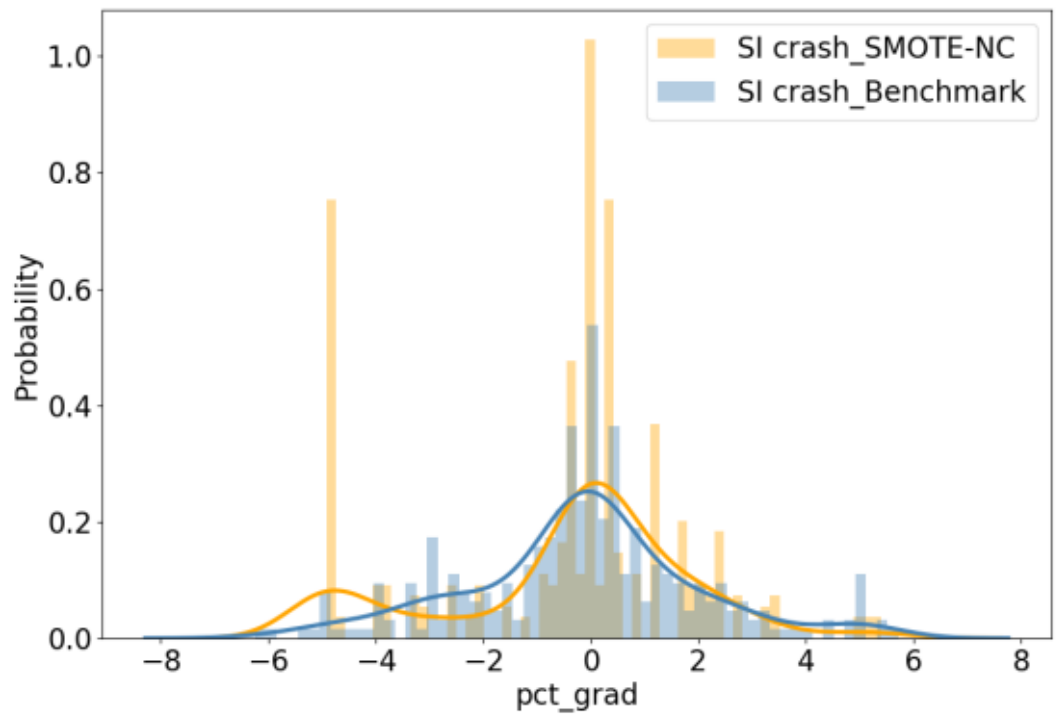

(6) Grade percentage (SMOTE-NC)

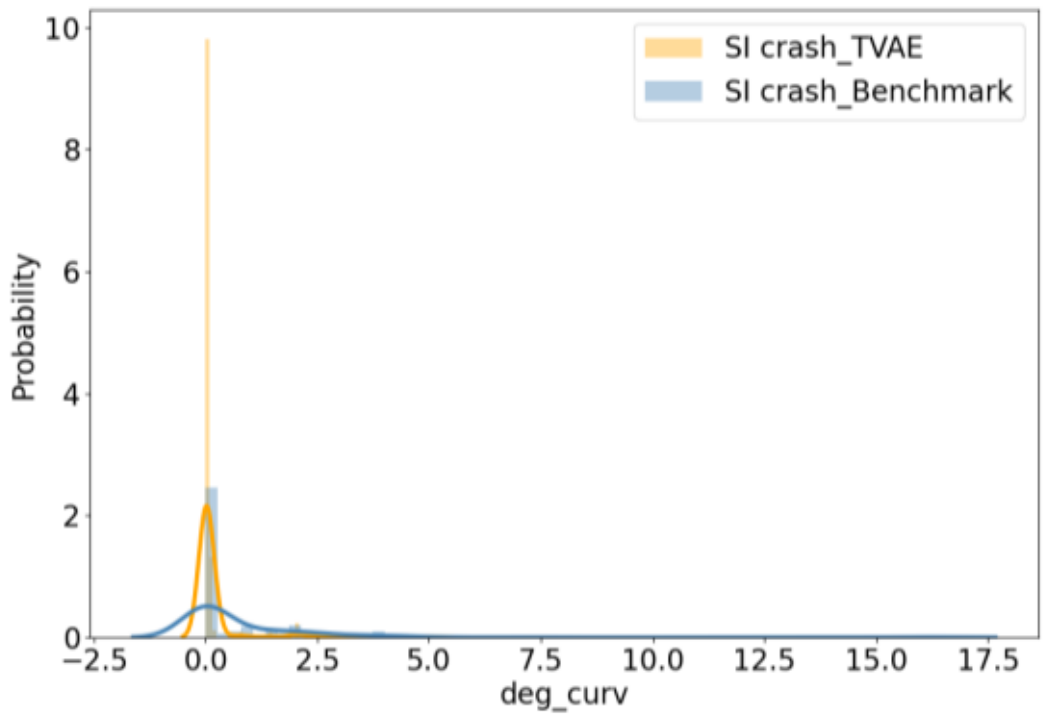

(7) Curvature degree (TVAE)

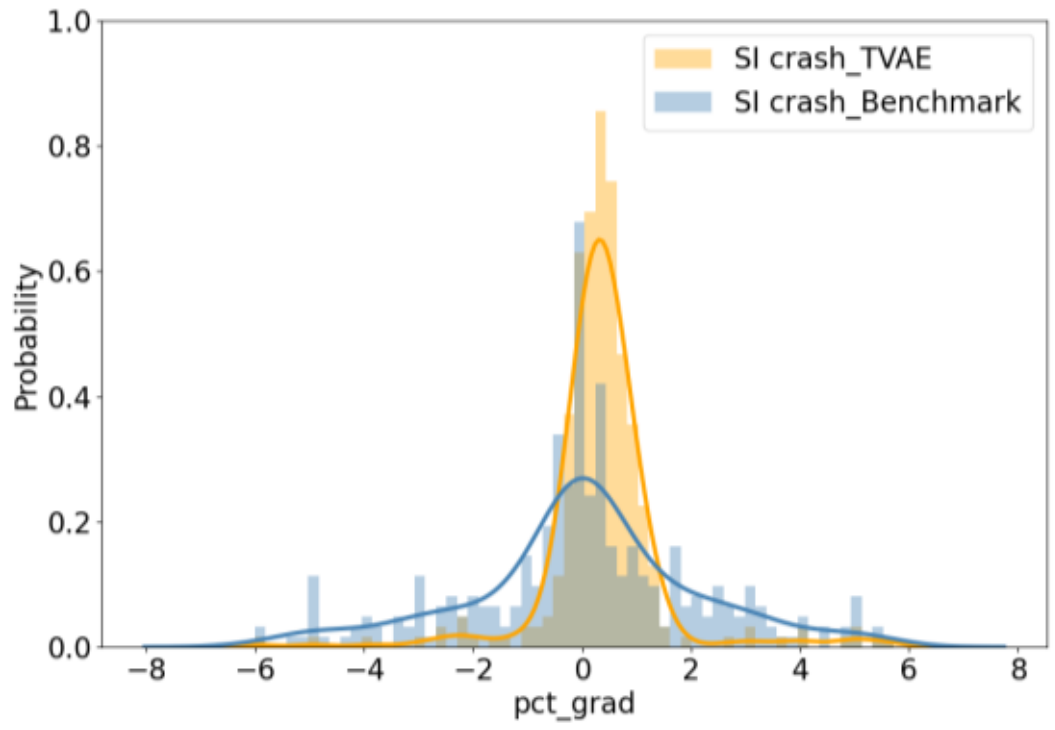

(8) Grade percentage (TVAE)

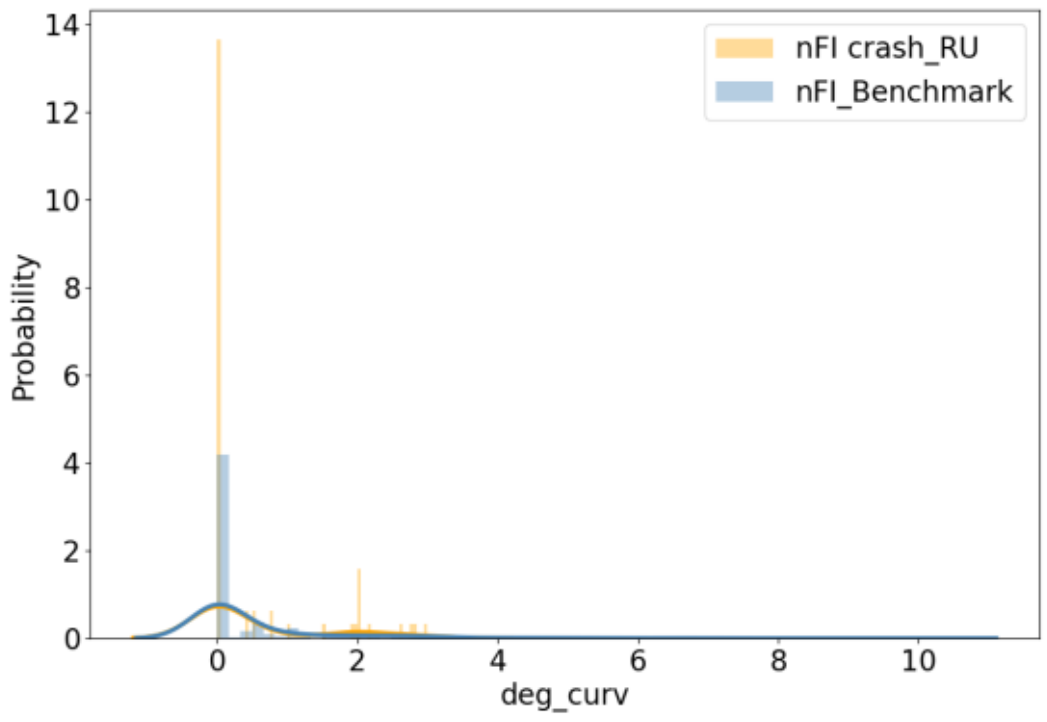

(9) Curvature degree (RU)

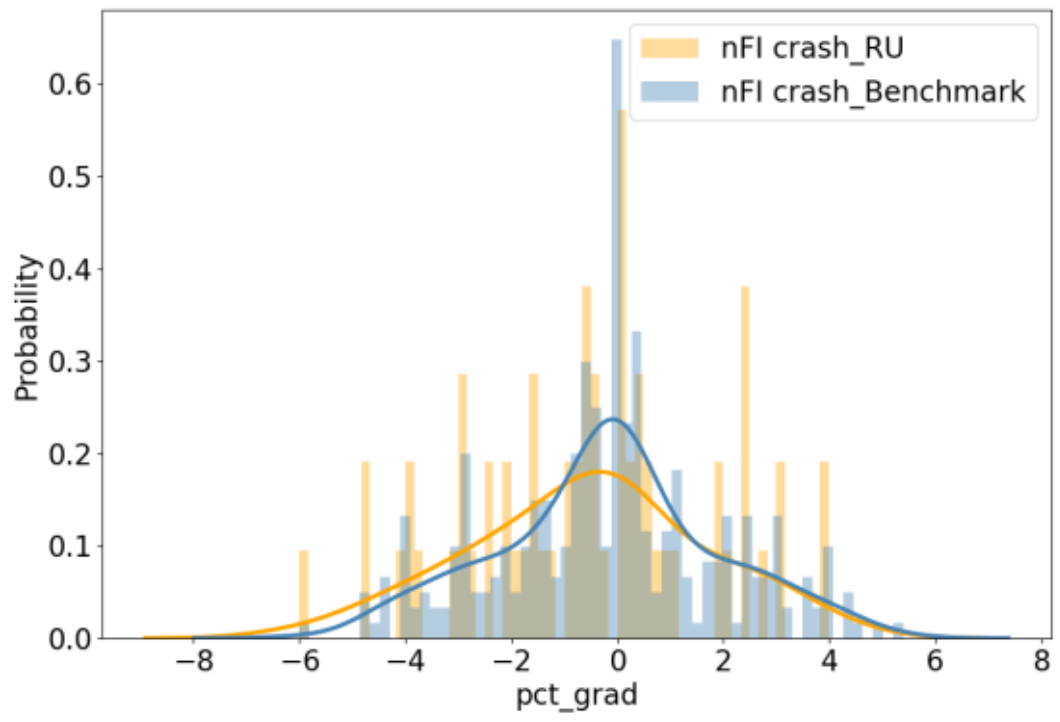

(10) Grade percentage (RU)

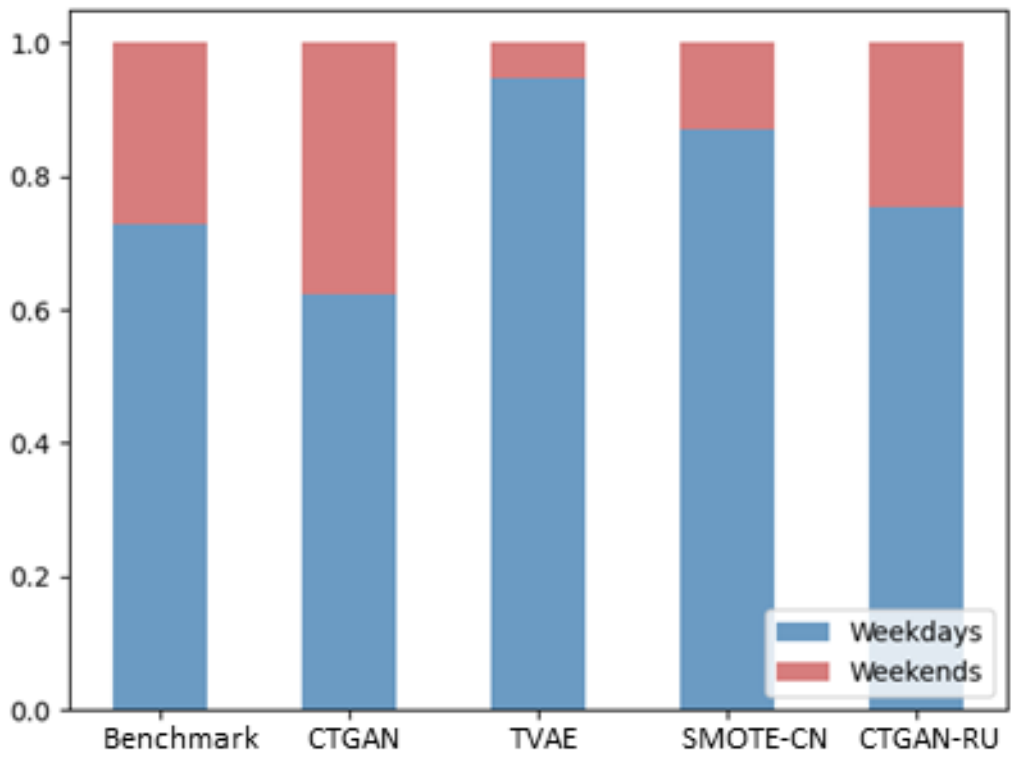

(11) Day of week

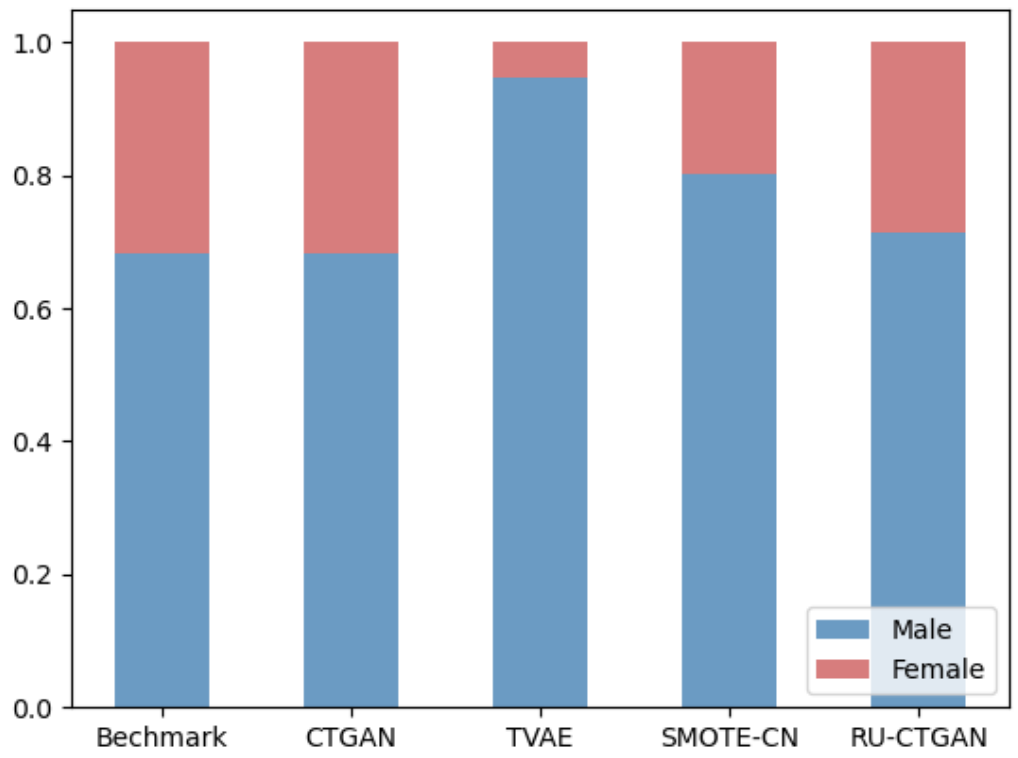

(12) Sex

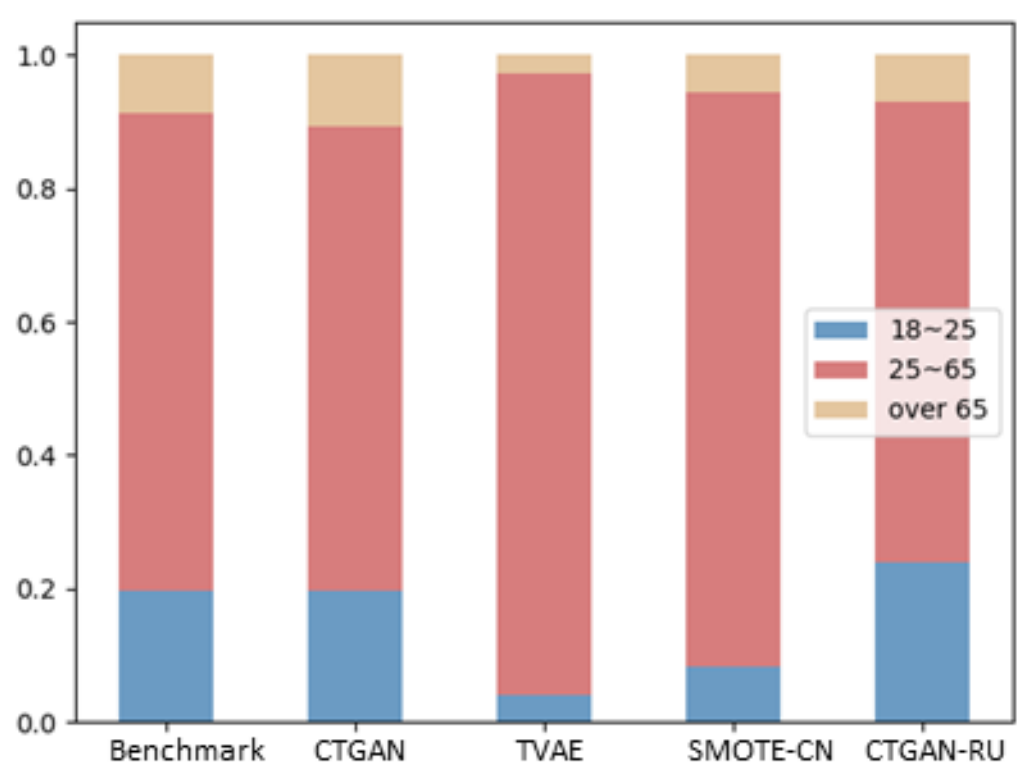

(13) Driver's age

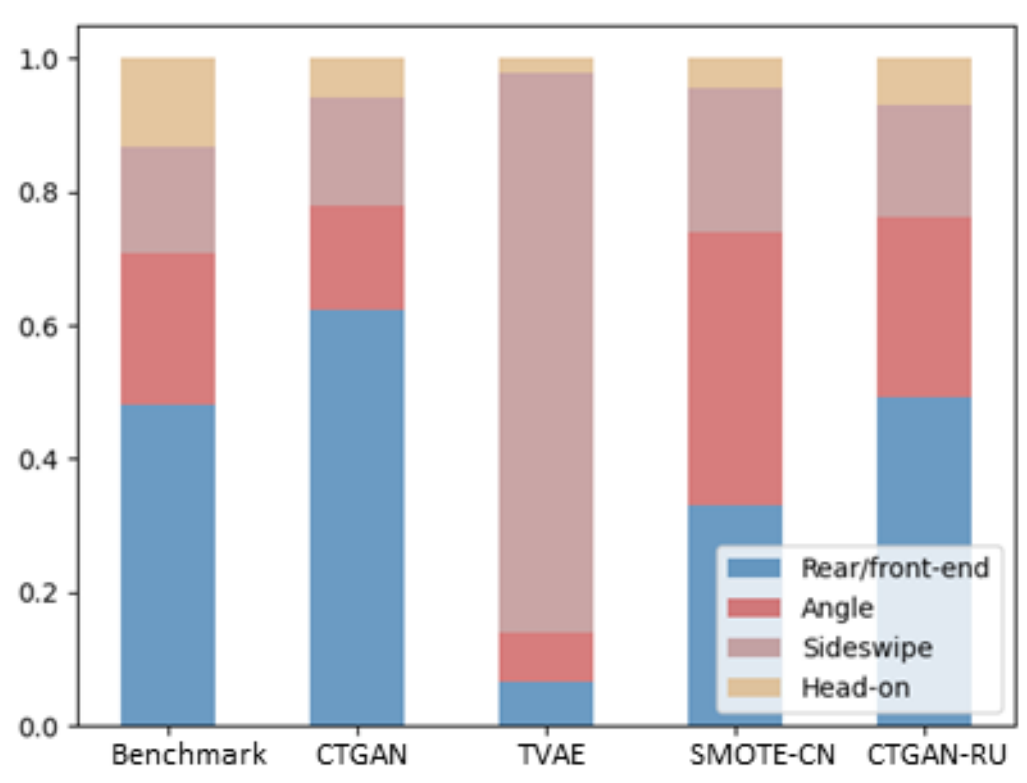

(14) Type of collision

Fig.5 Distribution of synthetic data

Also, density heatmaps of the joint distribution of variables including driver's age, type of collision, posted speed limit and curvature degree are calculated for the comparison among the balanced benchmark data and the synthetic data, where darker colors indicate higher density (see Fig. 6(1-10)). As shown in Fig. 6(1-10), deviations in the joint distribution of synthetic data based on the CTGAN and CTGAN-RU are much smaller than those based on SMOTE, TVAE, and RU for all variables. Both Fig. 6(3) and Fig. 6(5) demonstrate that the data synthesized by CTGAN and CTGAN-RU exhibit two highly dense areas, which are consistent with those of the benchmark dataset. This applies even to minor modes, such as angle collision, side-swipe collision, and head-on collision. The joint distribution generated by SMOTE and TVAE exhibits deficiencies, including varying numbers, shapes, and locations of densities (as shown in Fig.6(7)-(10)). These findings also suggest that the estimated joint probability distribution from CTGAN or CTGAN-RU can capture the variability of the benchmark data, including the minor modes.

In summary, the binary crash severity model developed using synthetic data generated by the proposed data generation method, CTGAN-RU performs similarly to that developed using the balanced benchmark data. For instance, the overall classification accuracy based on the CTGAN-RU method is the highest and closest to that based on the balanced benchmark data. Also, the estimated and joint probability distributions from CTGAN are similar to that of the balanced benchmark data, indicating that minor modes are properly generated and that the CTGAN addresses the mode collapse issue.

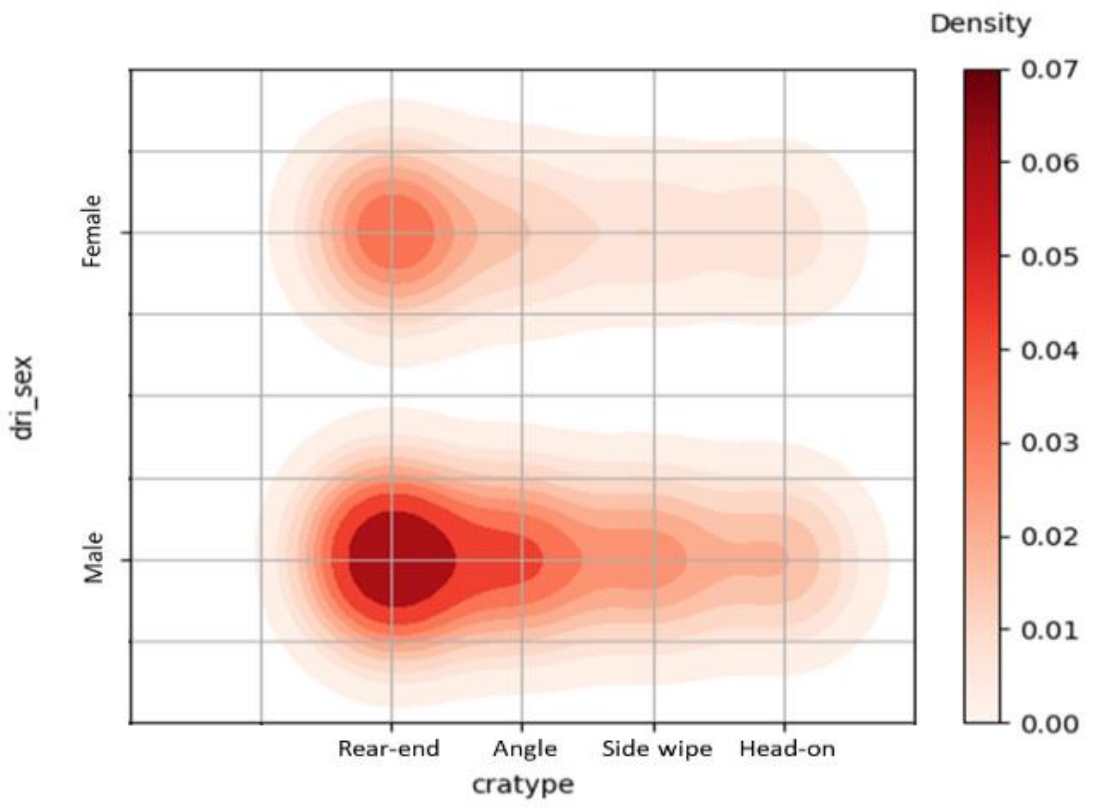

(1) Benchmark

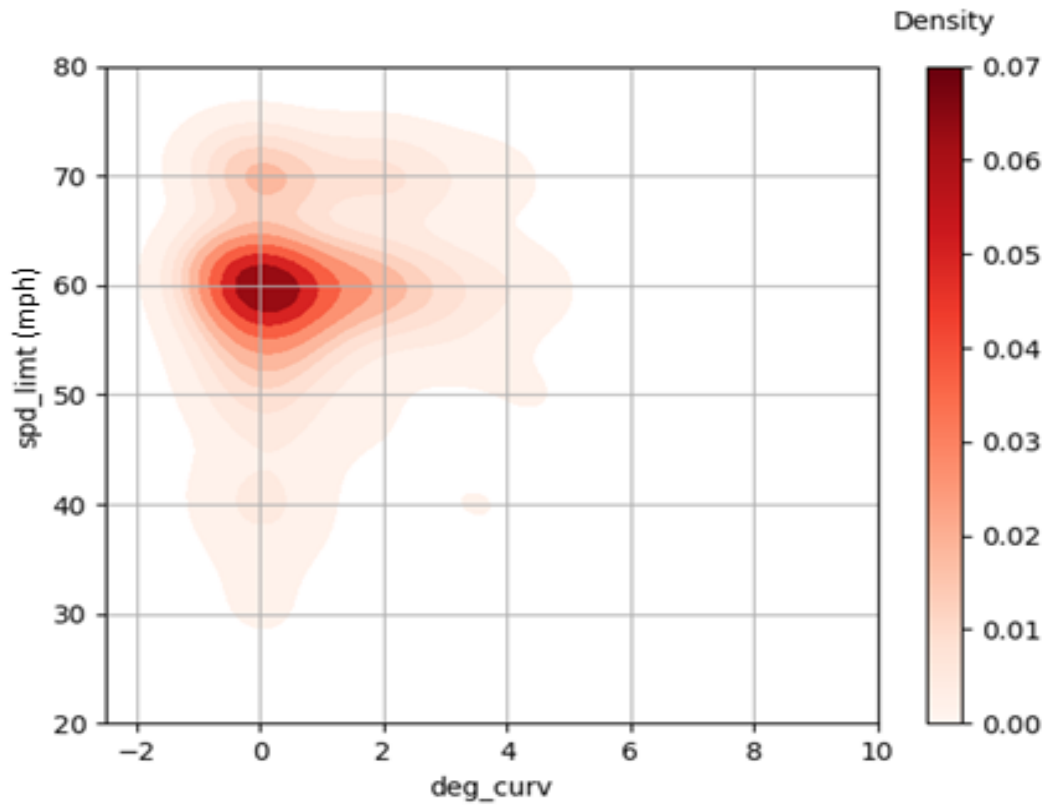

(2) Benchmark

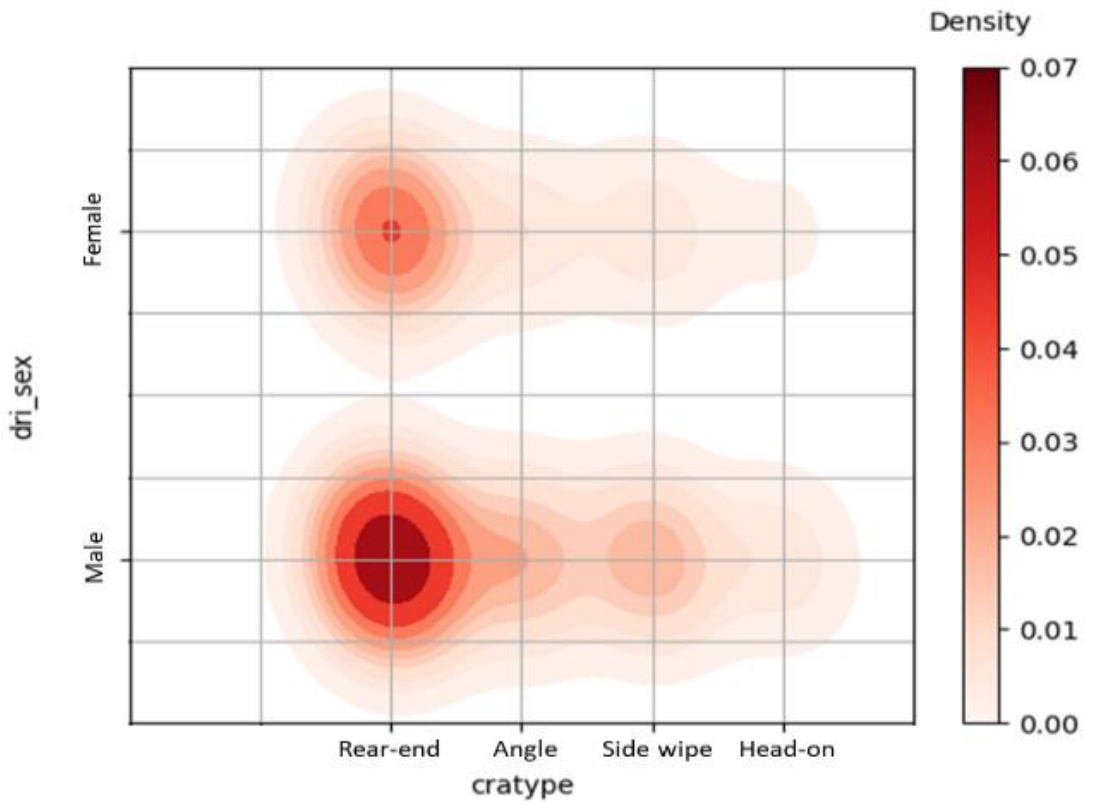

(3) CTGAN

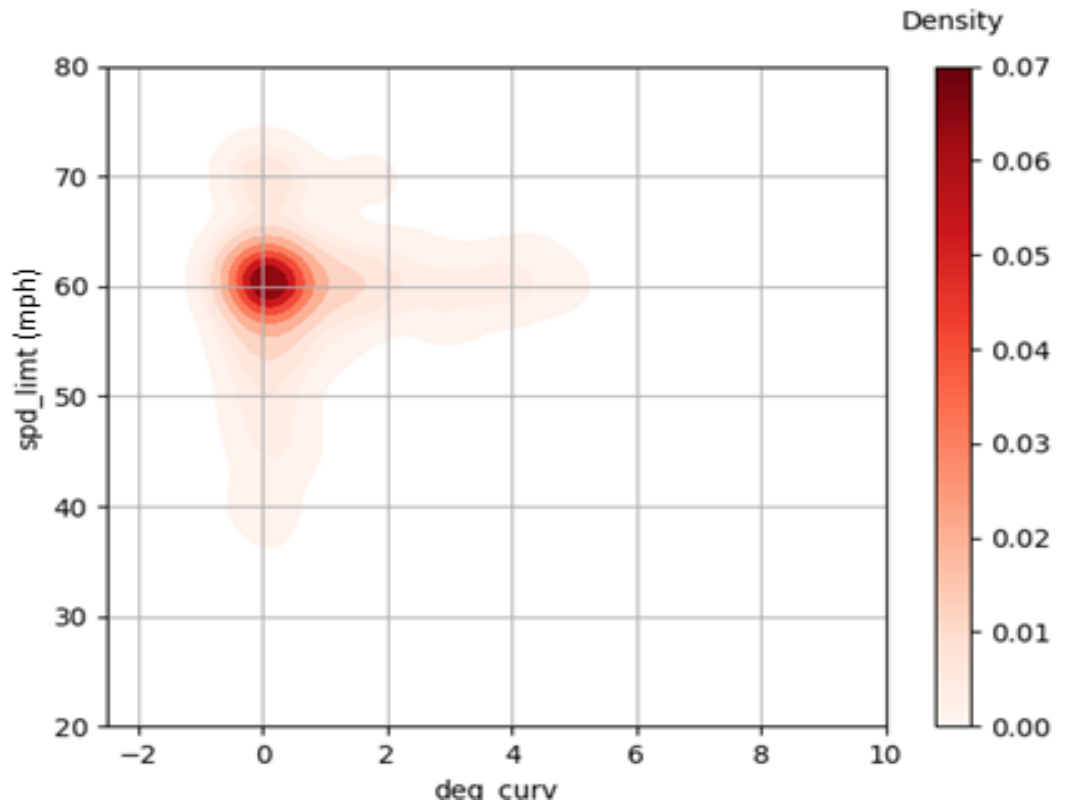

(4) CTGAN

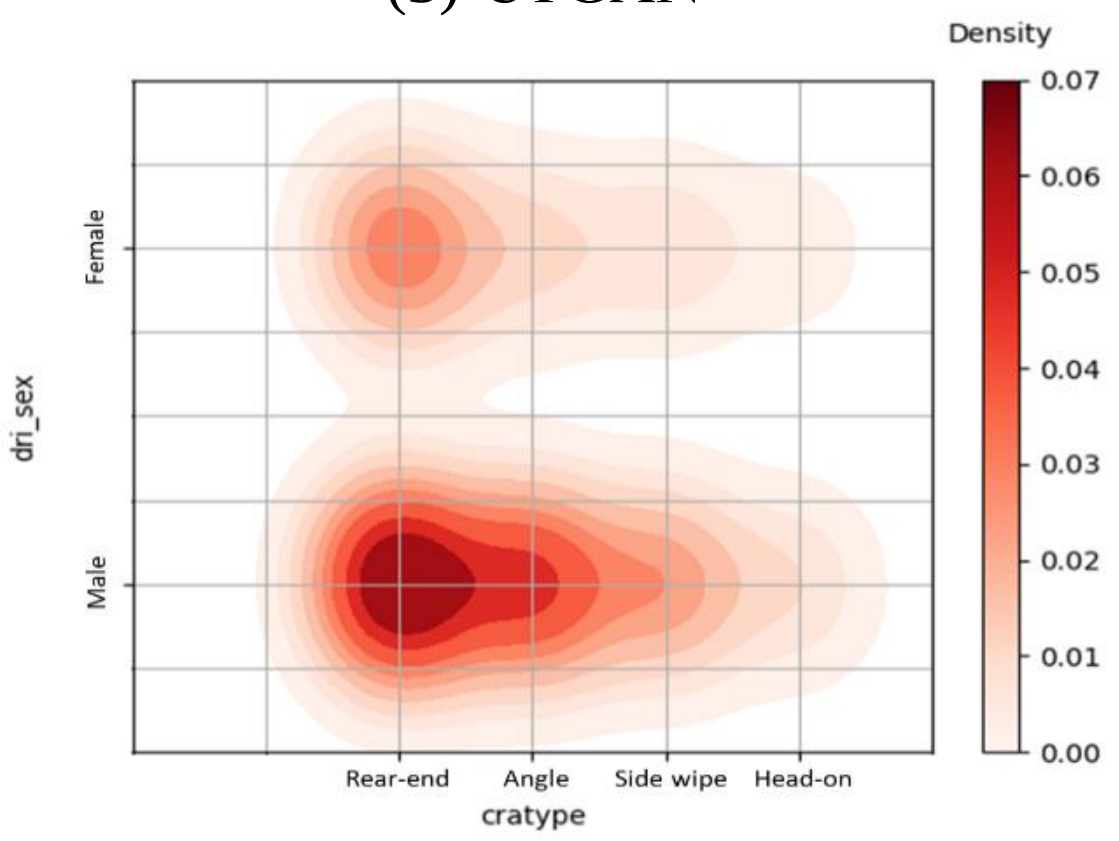

(5) RU-CTGAN

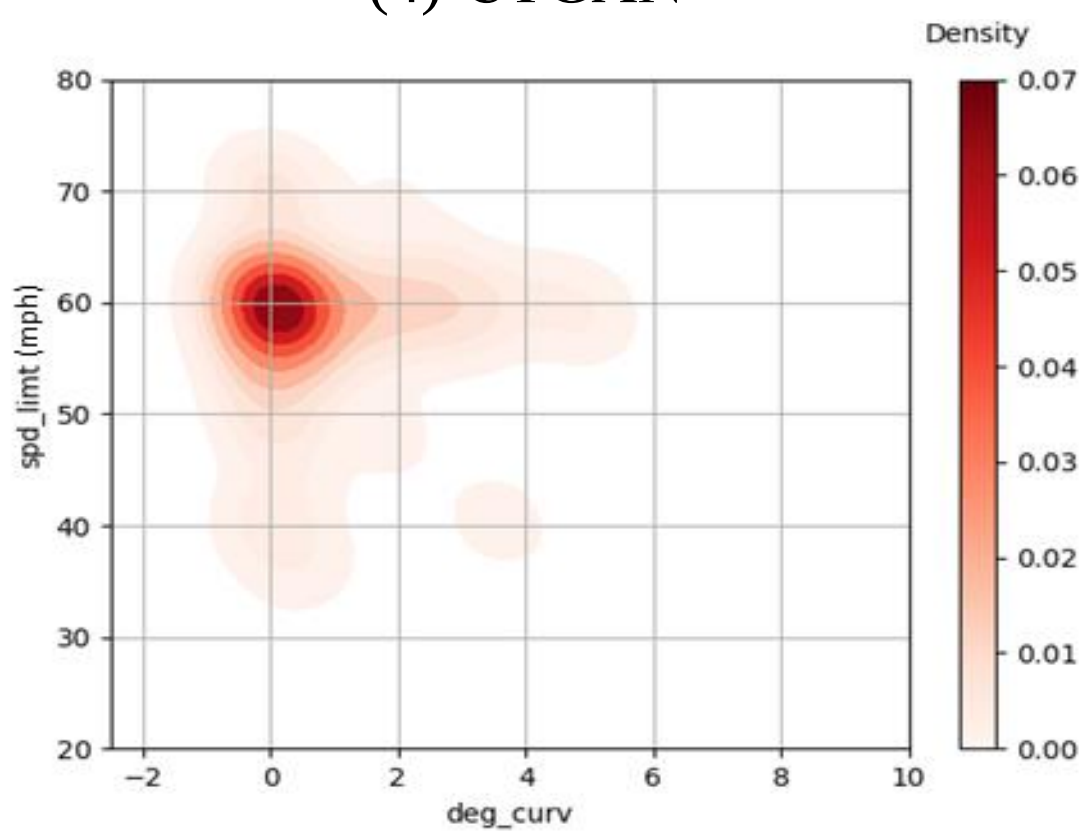

(6) RU-CTGAN

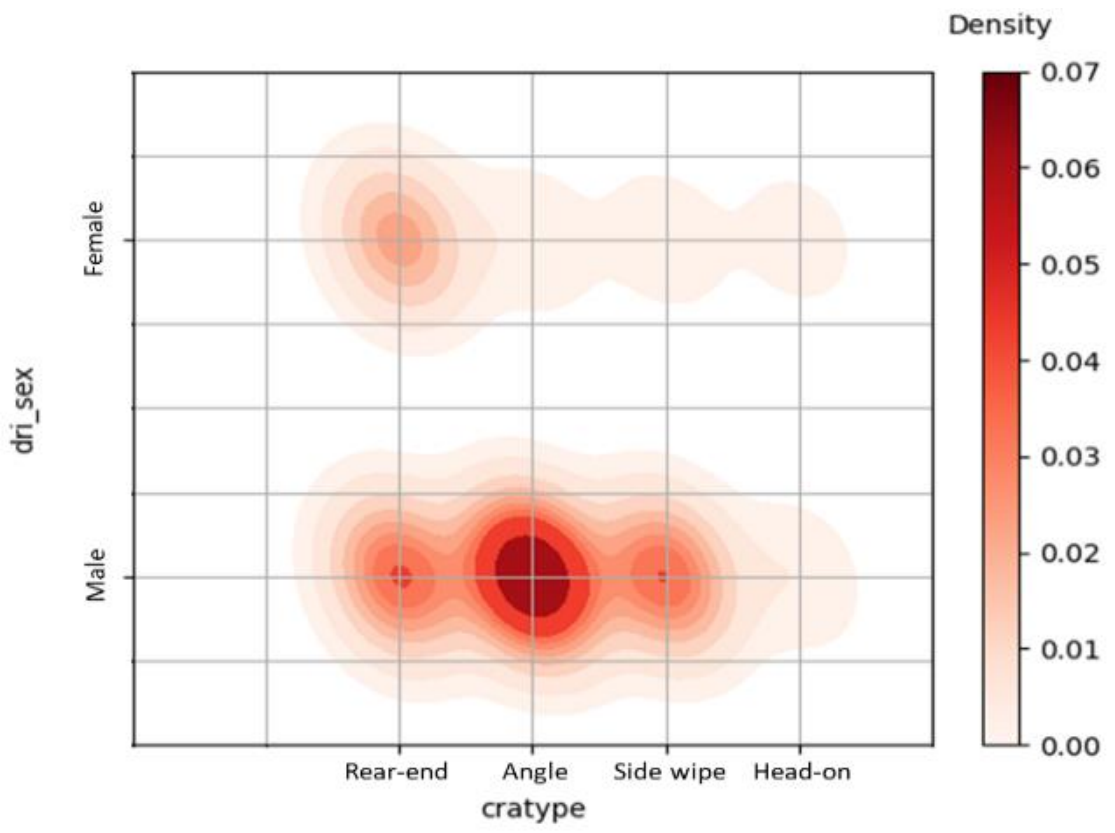

(7) SMOTE-NC

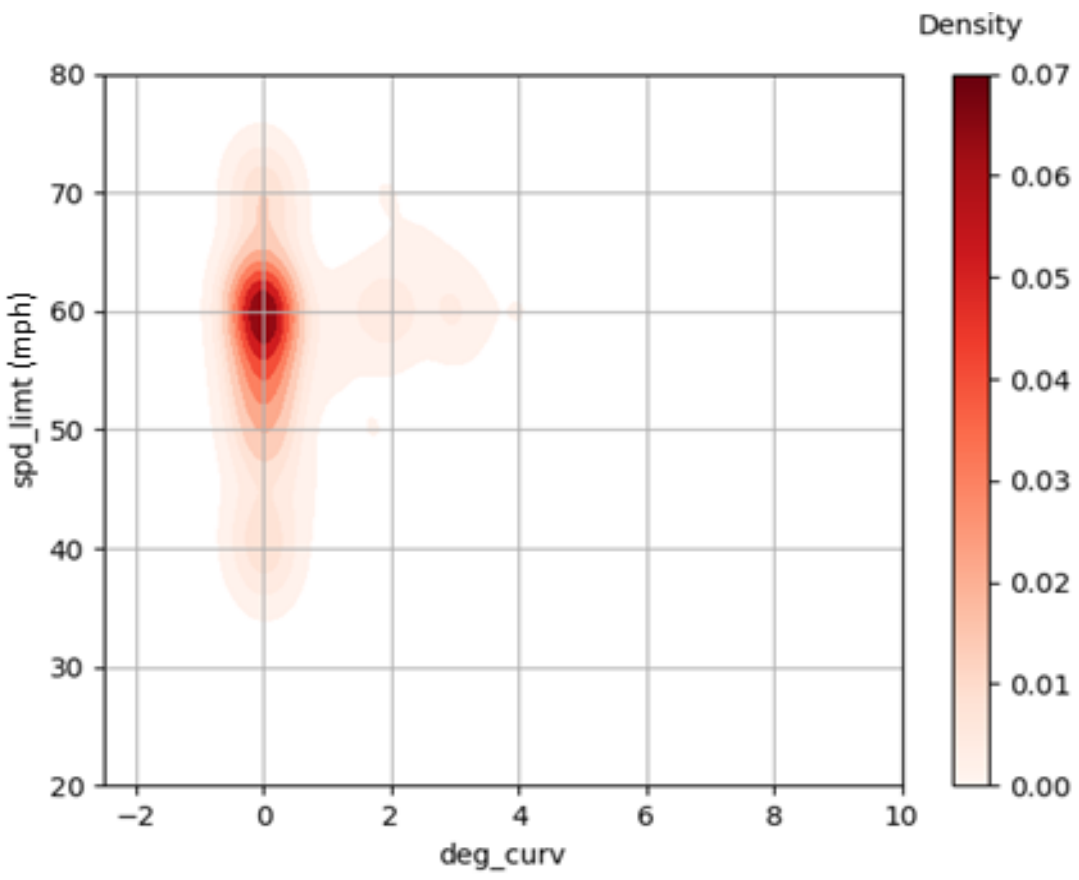

(8) SMOTE-CN

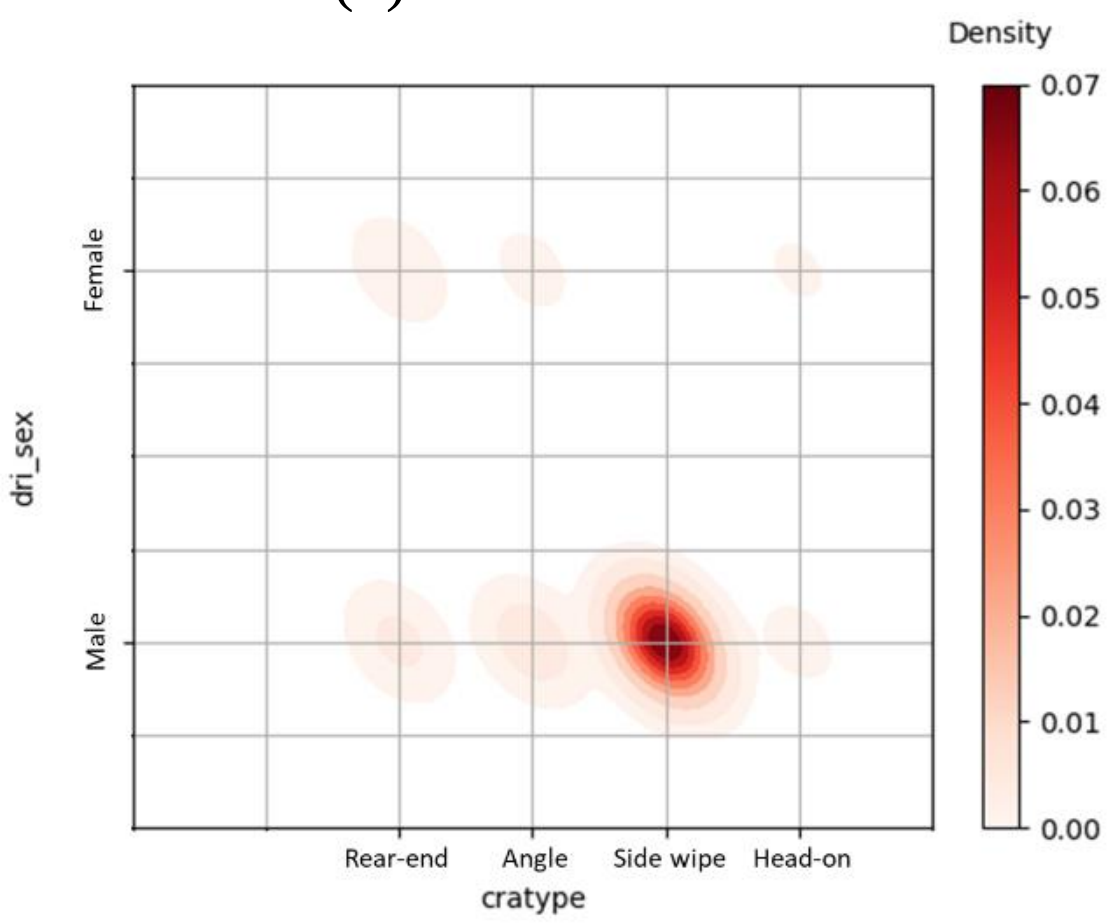

(9) TVAE

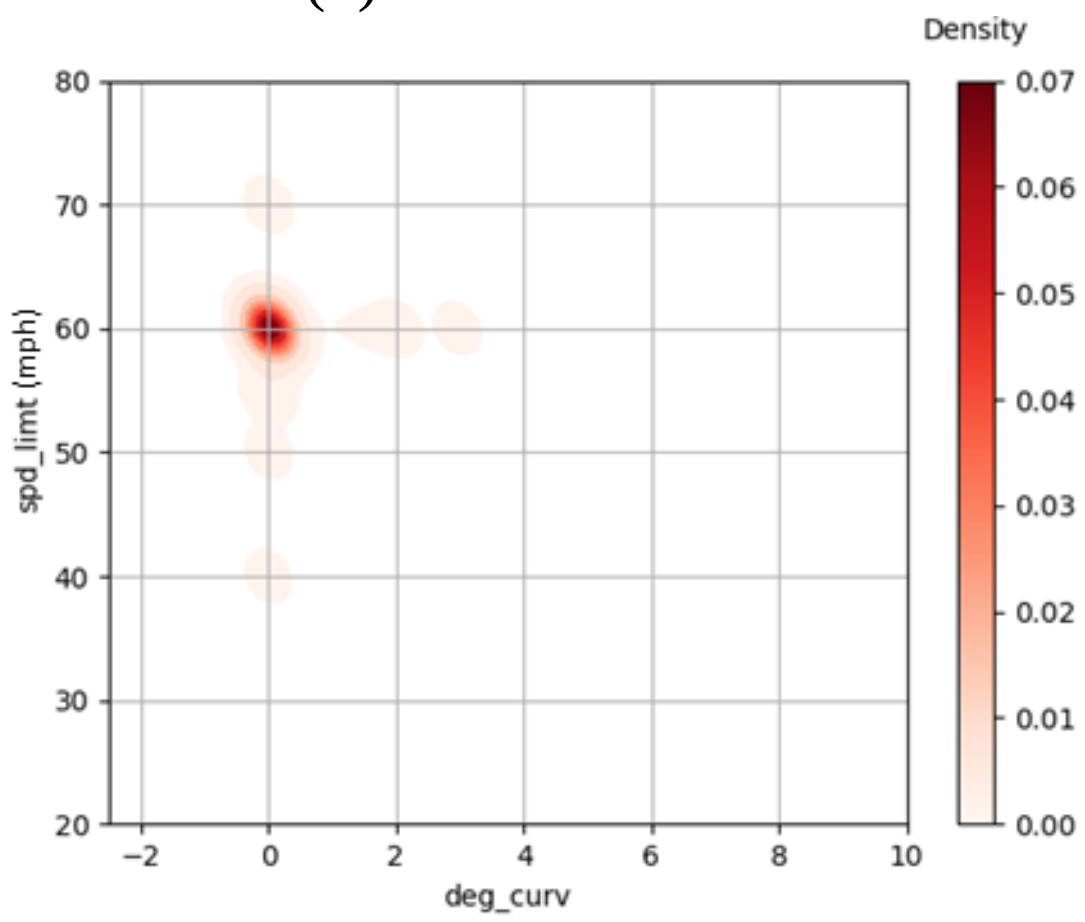

(10) TVAE

Fig.6 Joint probability distribution of synthetic data

*5.3.2 Various resampling ratios*

This part will examine how the resampling ratio of CTGAN-RU affects the consistency of interpretation regarding two-class imbalance issues using the real dataset. More precisely, a series of experiments are conducted wherein we vary the ratios of minority crashes (SI crashes) to majority crashes (nSI crashes) during the data generation processes. The experimental designs are detailed in Table 7. The results of the experiment for the ratio of 1:1 have been presented in Section 5.3.1. So here, we will explore situations with larger ratios (10:1, 8:1, 5:1, 4:1, 2:1). The benchmark dataset remains the same as that in Section 5.3.1.

Table 7 Experimental designs for various resampling ratios of CTGAN-RU

| Data type | Balanced total dataset | Imbalanced testing dataset | Benchmark dataset | Imbalanced training dataset | CTGAN-RU | | | | |
|---|---|---|---|---|---|---|---|---|---|
| | | | | | 10:1 | 8:1 | 5:1 | 4:1 | 2:1 |
| nSI | 453 | 136 | 317 | 317 | 300 | 240 | 150 | 120 | 60 |
| SI | 453 | 9 | 317 | 15 | 30 | 30 | 30 | 30 | 30 |
| Tatal | 906 | 145 | 634 | 332 | 330 | 270 | 180 | 150 | 90 |

Table 8 displays the evaluation of the model for different sampling ratios of CTGAN-RU. One can observe that reducing the nSI to SI ratio results in an improvement in the performance of the crash severity model in terms of sensitivity and G-mean. Particularly, when the ratio is 1:1, the model's performance aligns most closely with the benchmark dataset. Another interesting finding is that the more SI crashes there are, the better the model performs. Modifying the ratio of nSI to SI has a greater impact on sensitivity than on specificity. In general, the optimal performance is achieved when the ratio of nSI to SI is 1:1.

Table 8 Classification accuracy for various resampling ratios of CTGAN-RU

| Measurements | Benchmark dataset | CTGAN-RU (10:1) | CTGAN-RU (8:1) | CTGAN-RU (5:1) | CTGAN-RU (4:1) | CTGAN-RU (2:1) | CTGAN-RU (1:1) |
|---|---|---|---|---|---|---|---|
| Sensitivity | 0.773 | 0.409 | 0.507 | 0.500 | 0.545 | 0.636 | 0.778 |
| Specificity | 0.789 | 0.953 | 0.831 | 0.953 | 0.953 | 0.921 | 0.772 |
| G-mean | 0.781 | 0.624 | 0.649 | 0.690 | 0.721 | 0.766 | 0.775 |

Fig. 7(1)-(14) presents the conditional and joint probability distributions of both the balanced benchmark data and synthetic data generated by CTGAN-RU. Fig. 7(1)-(8) demonstrate that the discrepancies in the synthetic data distribution generated by CTGAN-RU decrease as the nSI to SI ratio decreases. For example, the dataset with a 10:1 ratio (nSI: SI) displays significant irregularities, including abnormal data points and skewed concentration, particularly within the low-frequency distribution segment. Compared to the ratio of 10:1, there are fewer abnormal points and more similar distributions to the benchmark dataset. Considering the discrete variables (see Fig.7(6)-(8), i.e., *Driver's sex*, *Location*, *Weekday*), the smallest ratio yields the most stable performance compared to the benchmark dataset across various

resampling ratios. When it comes to the joint distribution illustrated in Figs. 7(9)-(14), it is apparent that the discrepancies in the joint distribution of the synthetic data are significantly smaller at a 2:1 ratio compared to larger ratios. Based on these findings, it can be inferred that a lower ratio of the majority class (nSI) to the minority class (SI) is more effective in capturing data hidden patterns, especially for sparse variables, and results in a well-fitted distribution of risk factors for SI crashes.

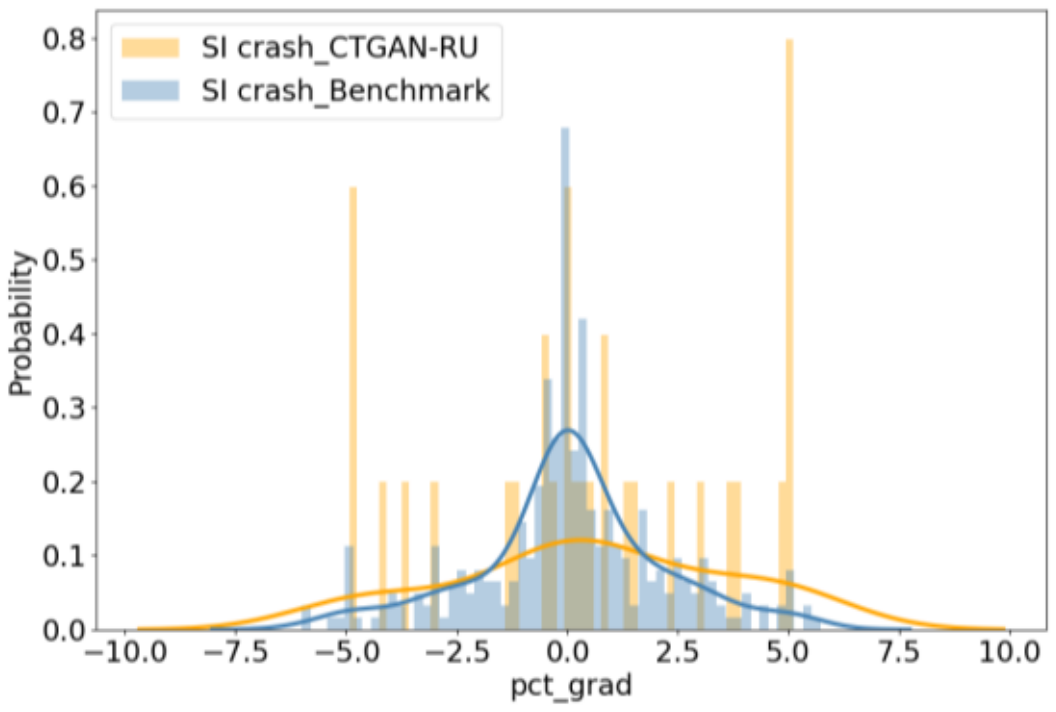

(1) Ratio of SI: nSI = 10:1

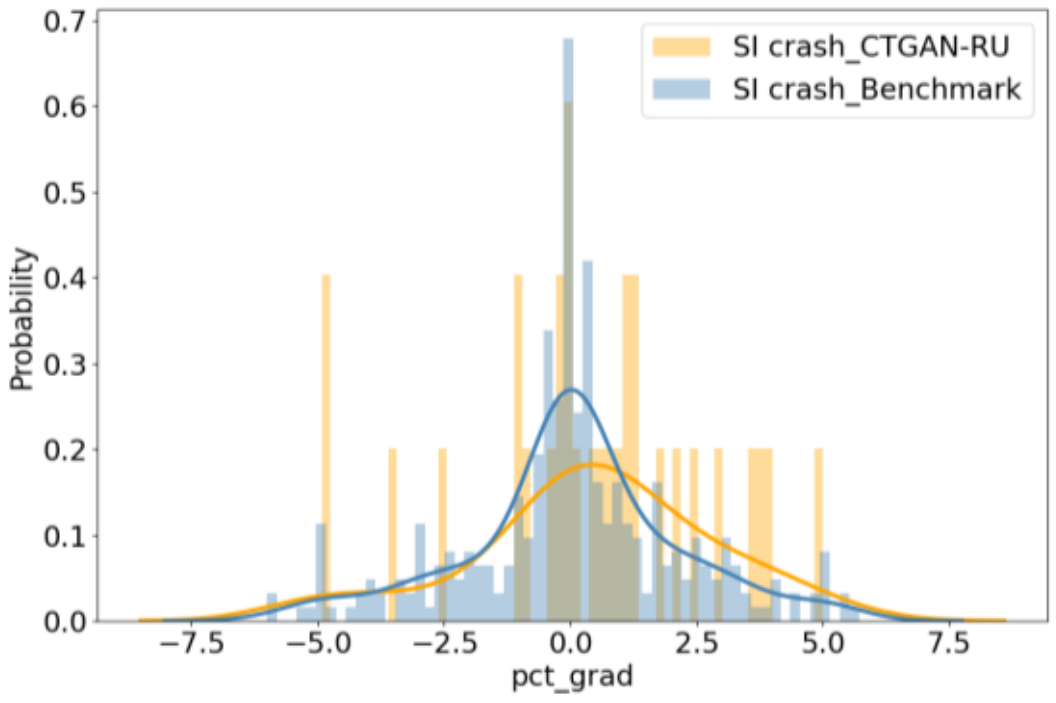

(2) Ratio of SI: nSI = 8:1

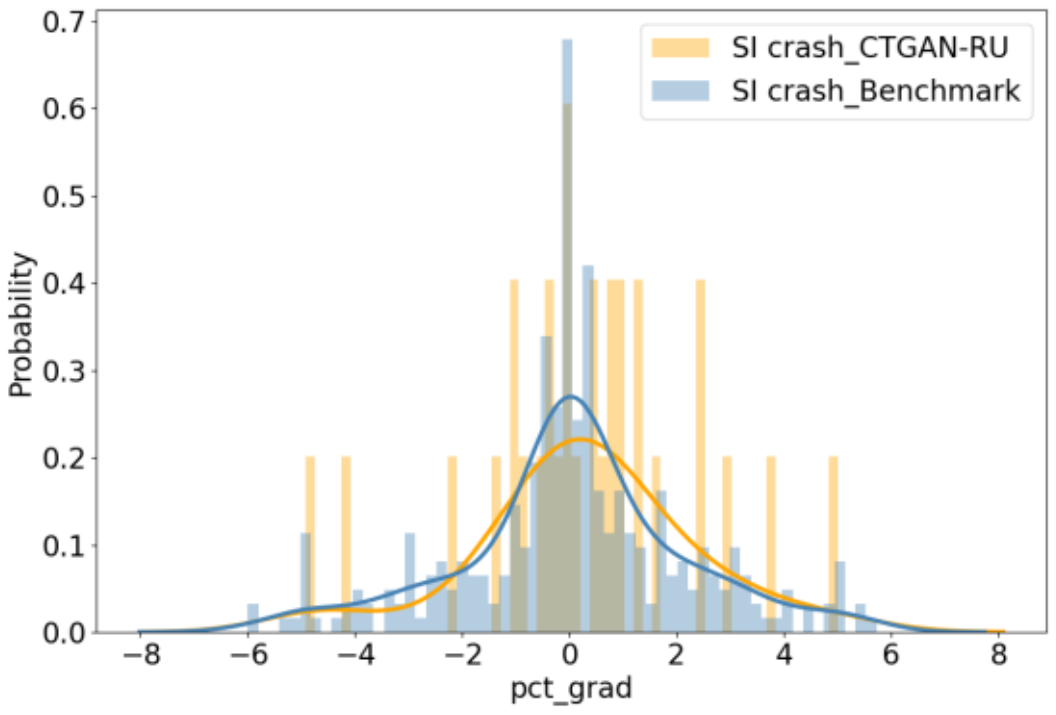

(3) Ratio of SI: nSI = 5:1

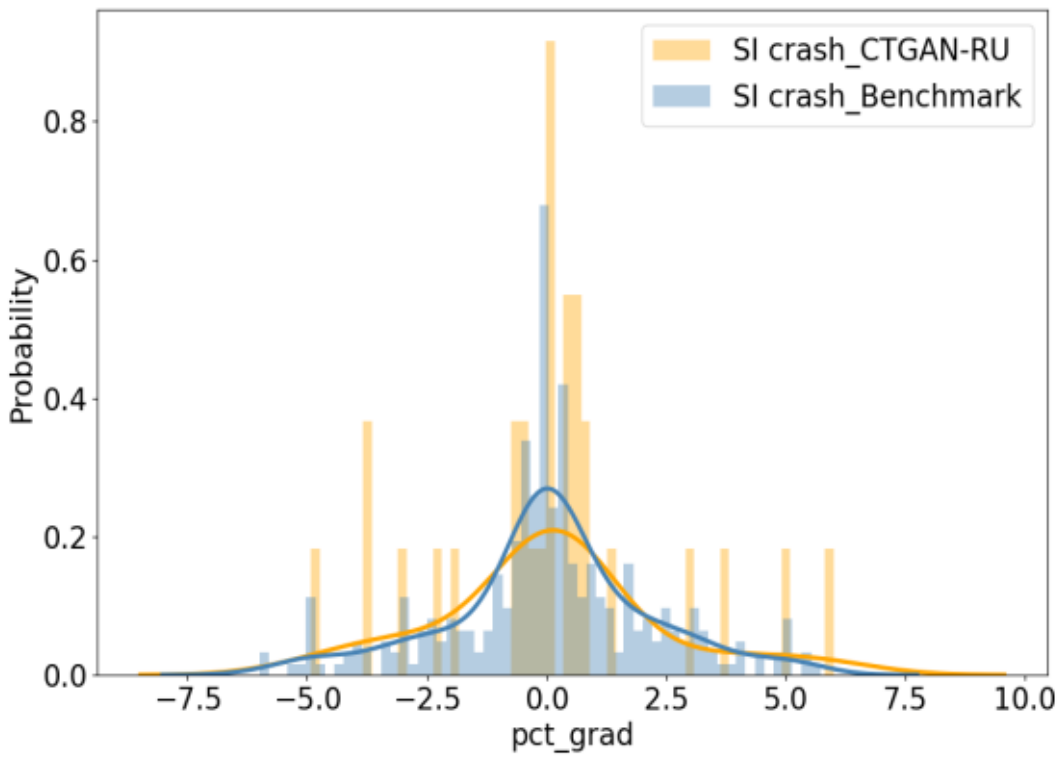

(4) Ratio of SI: nSI = 4:1

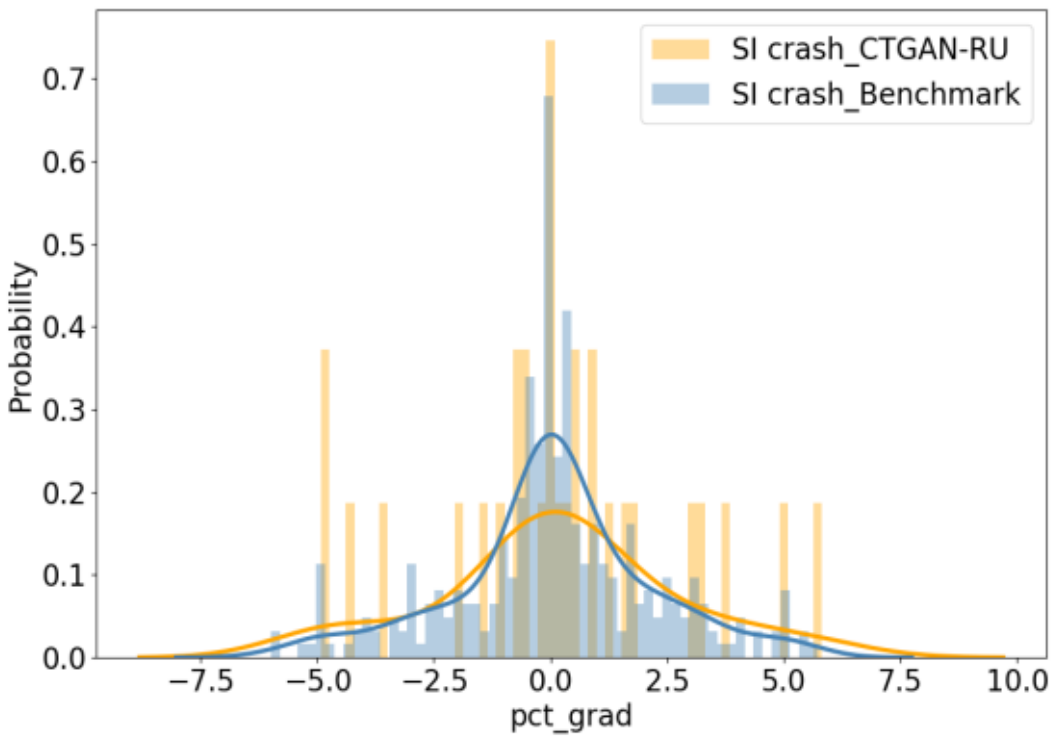

(5) Ratio of SI: nSI = 2:1

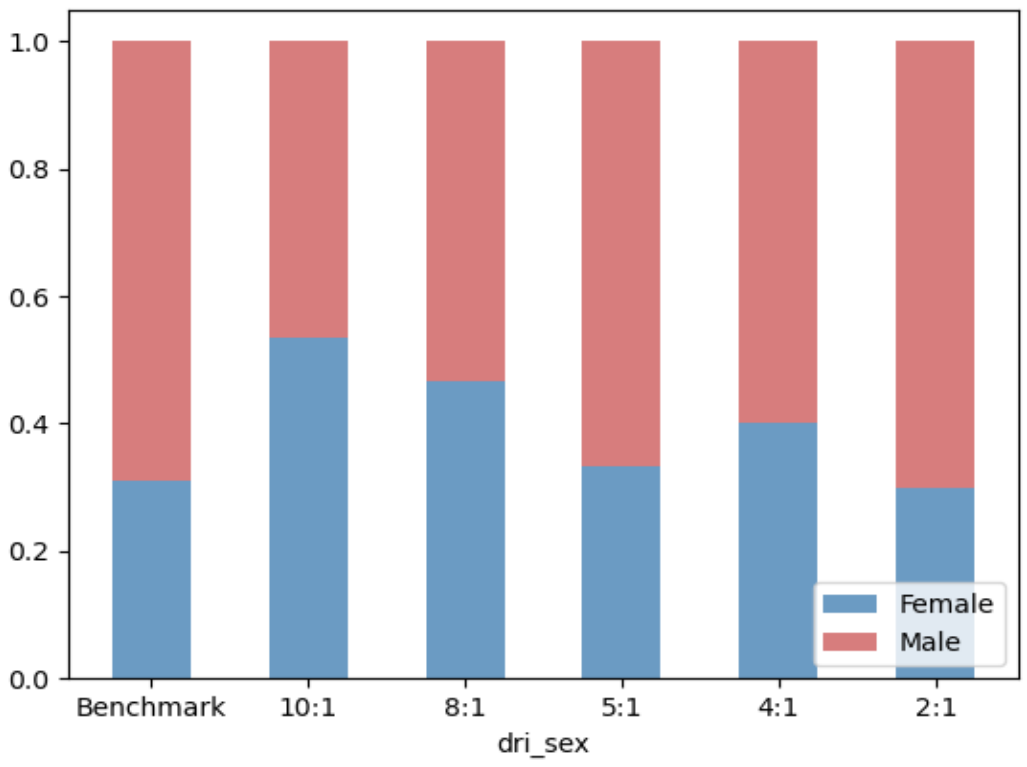

(6) Driver's sex (Various resampling ratios)

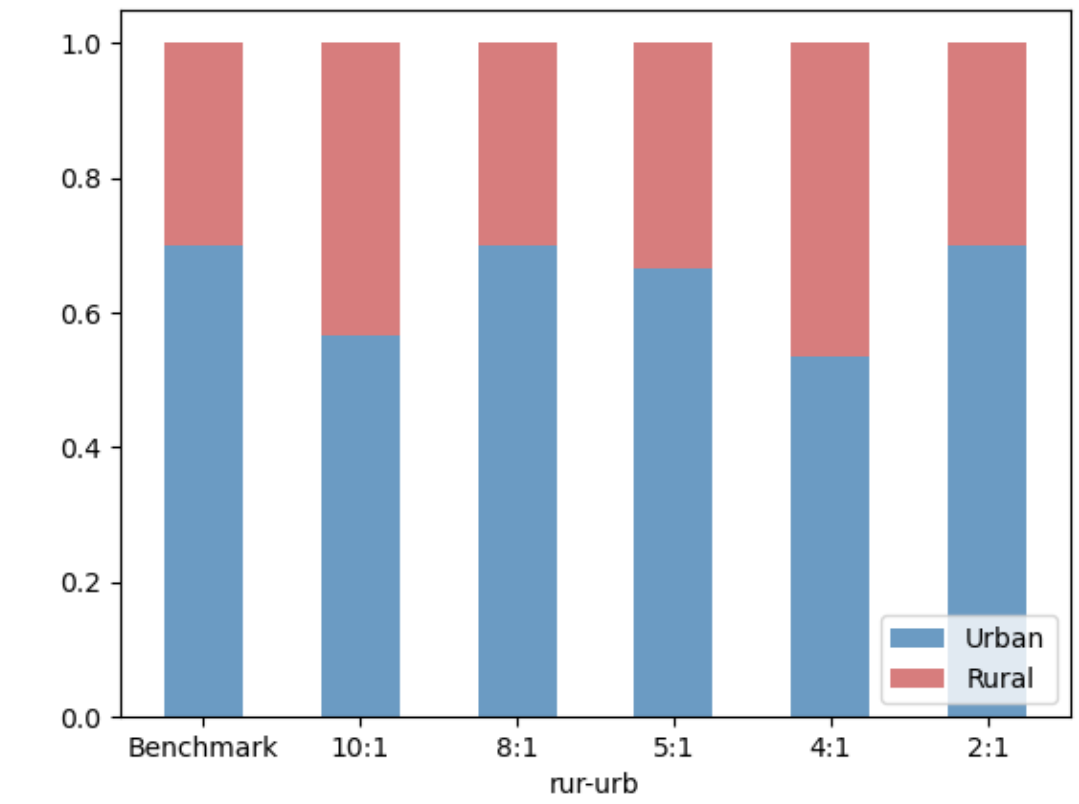

(7) Location (Various resampling ratios)

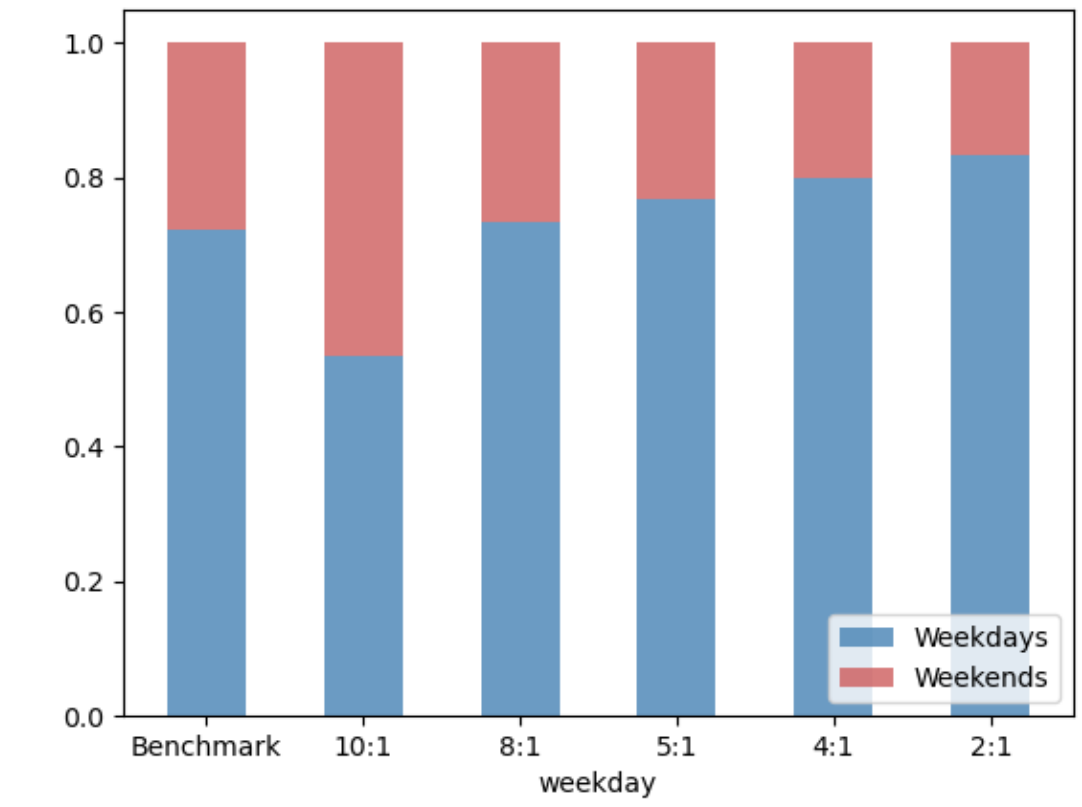

(8) Weekday (Various resampling ratios)

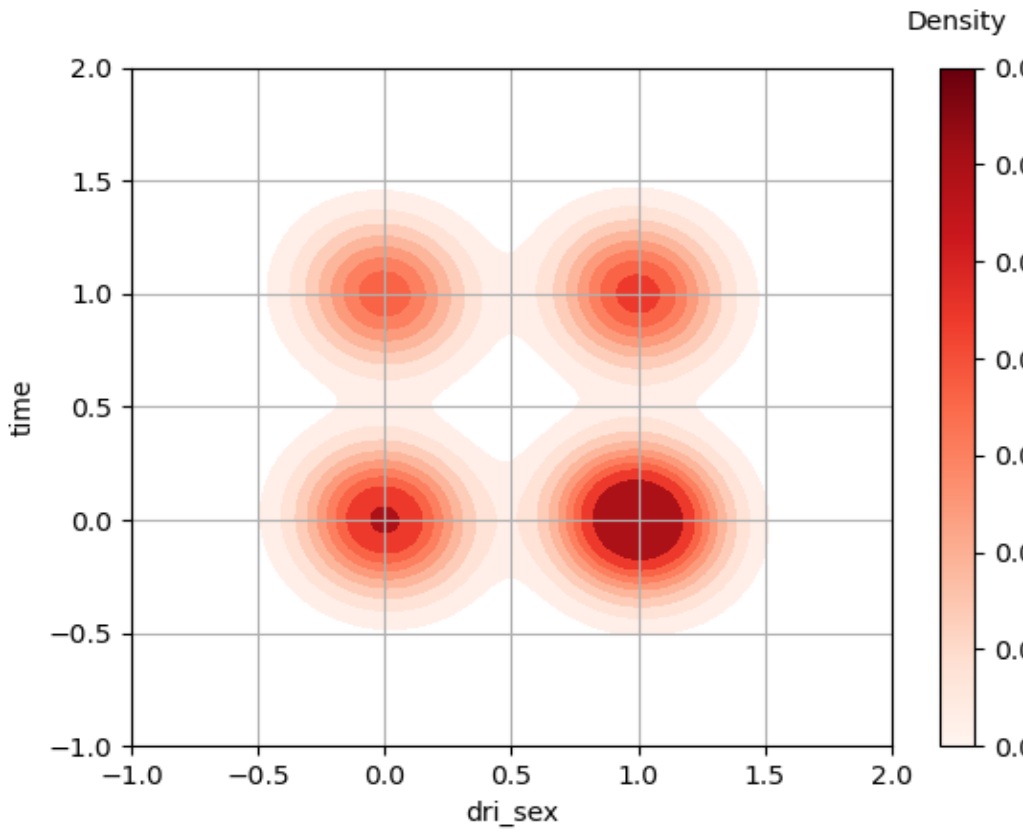

(9) Peak_hour and Driver's sex (Benchmark)

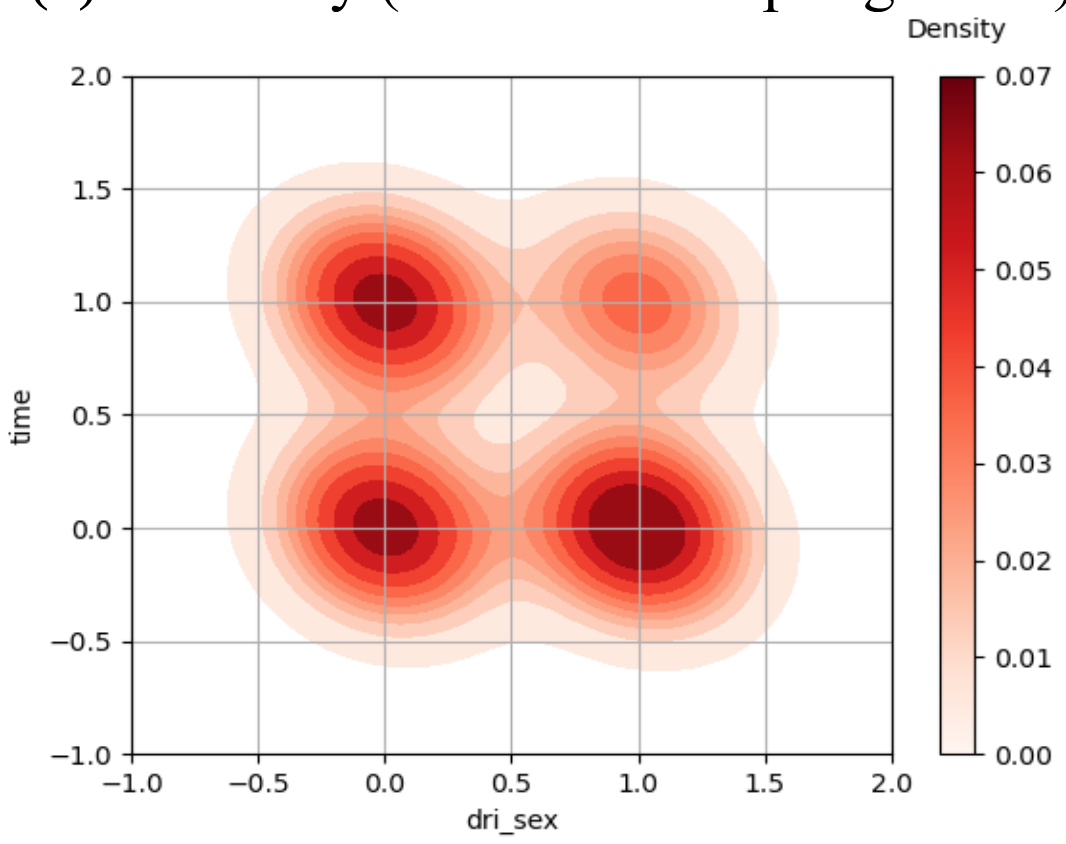

(10) Peak_hour and Driver's sex (CTGAN-RU, Ratio of SI: nSI = 10:1)

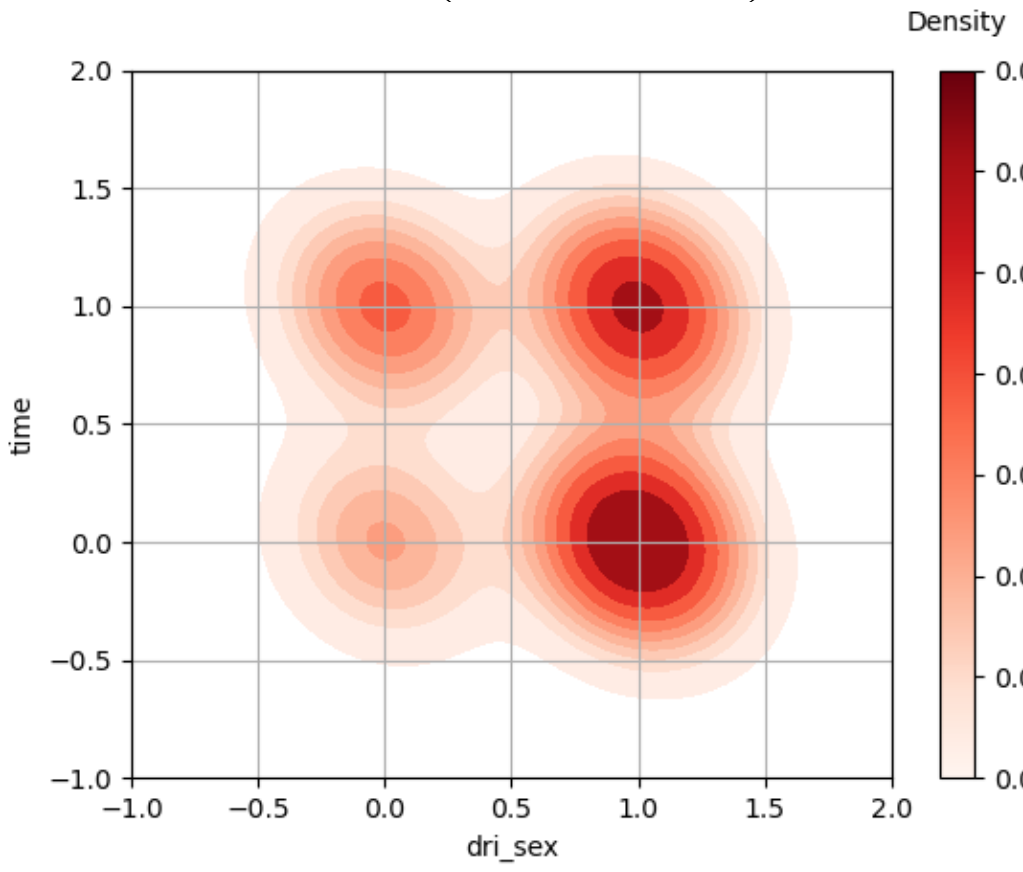

(11) Peak_hour and Driver's sex (CTGAN-RU, Ratio of SI: nSI = 8:1)

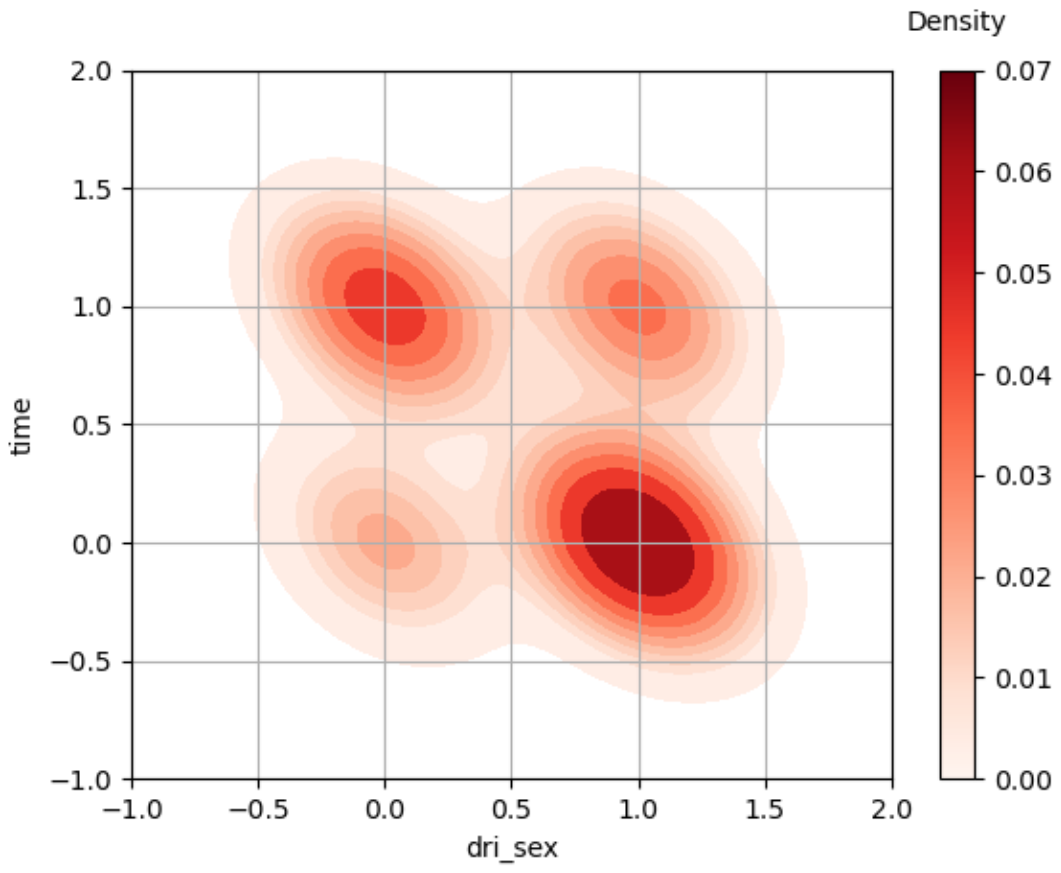

(12) Peak_hour and Driver's sex (CTGAN-RU, Ratio of SI: nSI = 5:1)

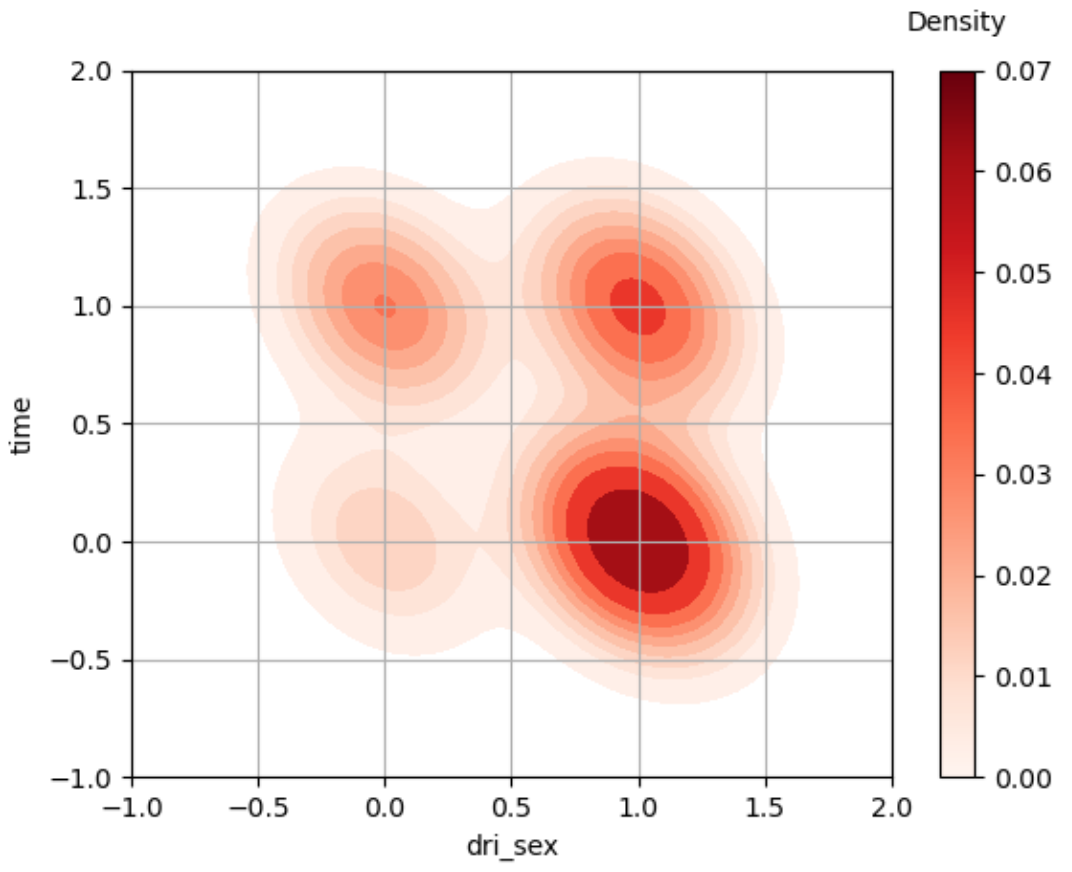

(13) Peak_hour and Driver's sex (CTGAN-RU, Ratio of SI: nSI = 4:1)

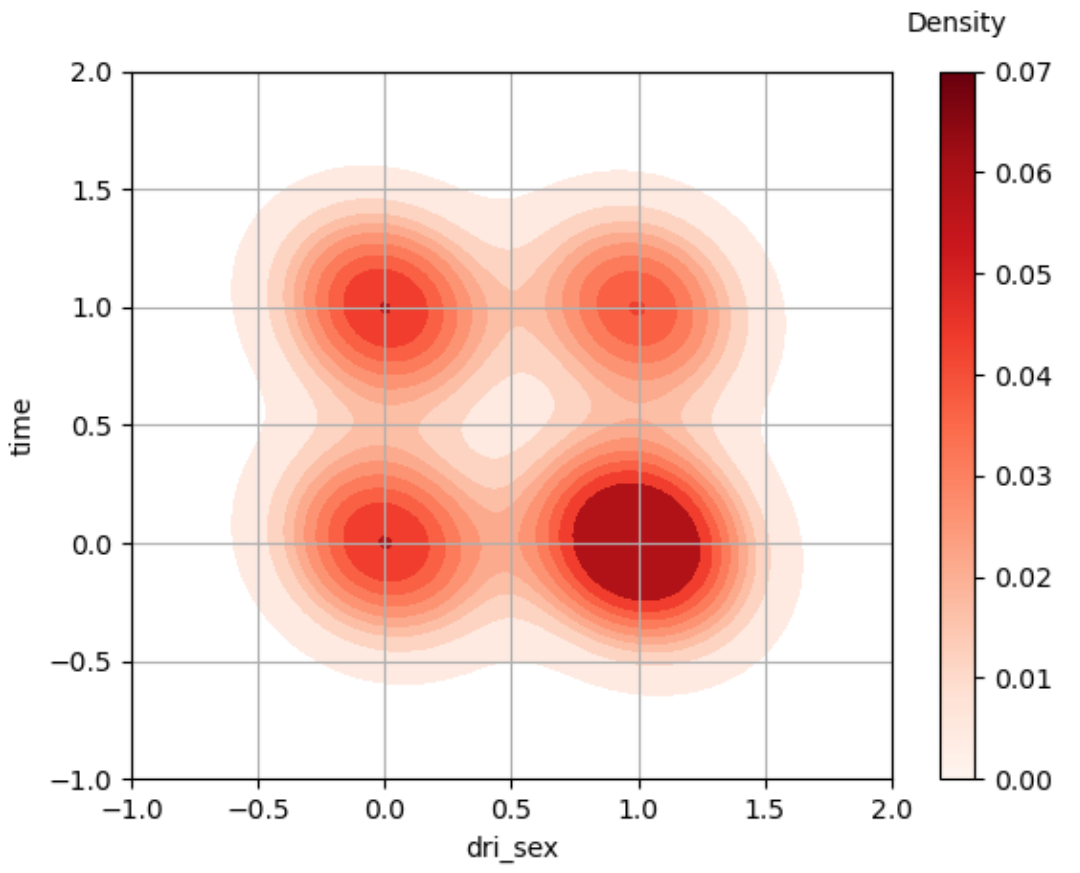

(14) Peak_hour and Driver's sex (CTGAN-RU, Ratio of SI: nSI = 2:1)

Fig. 7 Conditional and joint probability distribution of synthetic data for CTGAN-RU with various resampling ratios

### *5.3.3 Three-class imbalance*

In alignment with the experiments for two-class imbalance in section 5.3.1, we delineate the specific experiment design for assessing interpretative consistency within a three-class imbalance scenario for ordered crash severity modeling, as outlined in Table 9. The table provides comprehensive information regarding the balanced benchmark dataset and the CTGAN-RU resampled datasets based on the real dataset. Within this framework, Class 0, 1, and 2 correspond to property damage only, non-severe injury crashes (i.e., possible injury crashes and non-severe injury crashes), and severe injury crashes (i.e., severe crashes and fatal crashes), respectively. As depicted in Fig. 8, we align with the binary logit modeling methodology and outline the experiment design for evaluating the interpretative consistency in a three-class imbalance scenario. The experimental design of the simulation study is presented as follows:

1) To achieve a balanced dataset with a ratio of 1:1:1 of all three classes, we have retained the original number of samples for the minority class 2, which is 453. Consequently, the balanced dataset now consists of 453 samples for each class. We randomly selected a test dataset from this balanced dataset, comprising 30% of the samples, while the remaining 70% constitute the benchmark dataset.

2) Considering two types of three-class imbalance scenarios, we design Experiments 1 and 2 by randomly selecting class 2 or 3 from the benchmark dataset to explore the performance of the proposed data generation method. Experiment 1: two minorities; and Experiment 2: one minority to two majorities.

3) Based on the imbalanced training dataset, class 2 and 3 are both generated and expanded to 126 samples, while class 1 is dropped to 126 samples by using CTGAN-RU. In Experiment 2, class 3 is generated and expanded to 126 samples while class 1 and 2 are dropped to 126 samples by using CTGAN-RU.

Table 9 Number of observations for three-class imbalance

| Data type | Balanced total dataset | Benchmark dataset | CTGAN-RU Experiment 1 | | | CTGAN-RU Experiment 2 | | |
|---|---|---|---|---|---|---|---|---|
| | | | Imbalanced | | Balanced | Imbalanced | | Balanced |
| | | | testing | training | | testing | training | |
| Class 0 | 453 | 317 | 136 | 317 | 126 | 136 | 317 | 126 |
| Class 1 | 453 | 317 | 27 | 63 | 126 (generated) | 136 | 317 | 126 |
| Class 2 | 453 | 317 | 27 | 63 | 126 (generated) | 27 | 63 | 126 (generated) |
| Total | 1359 | 951 | 190 | 443 | 378 | 299 | 697 | 378 |

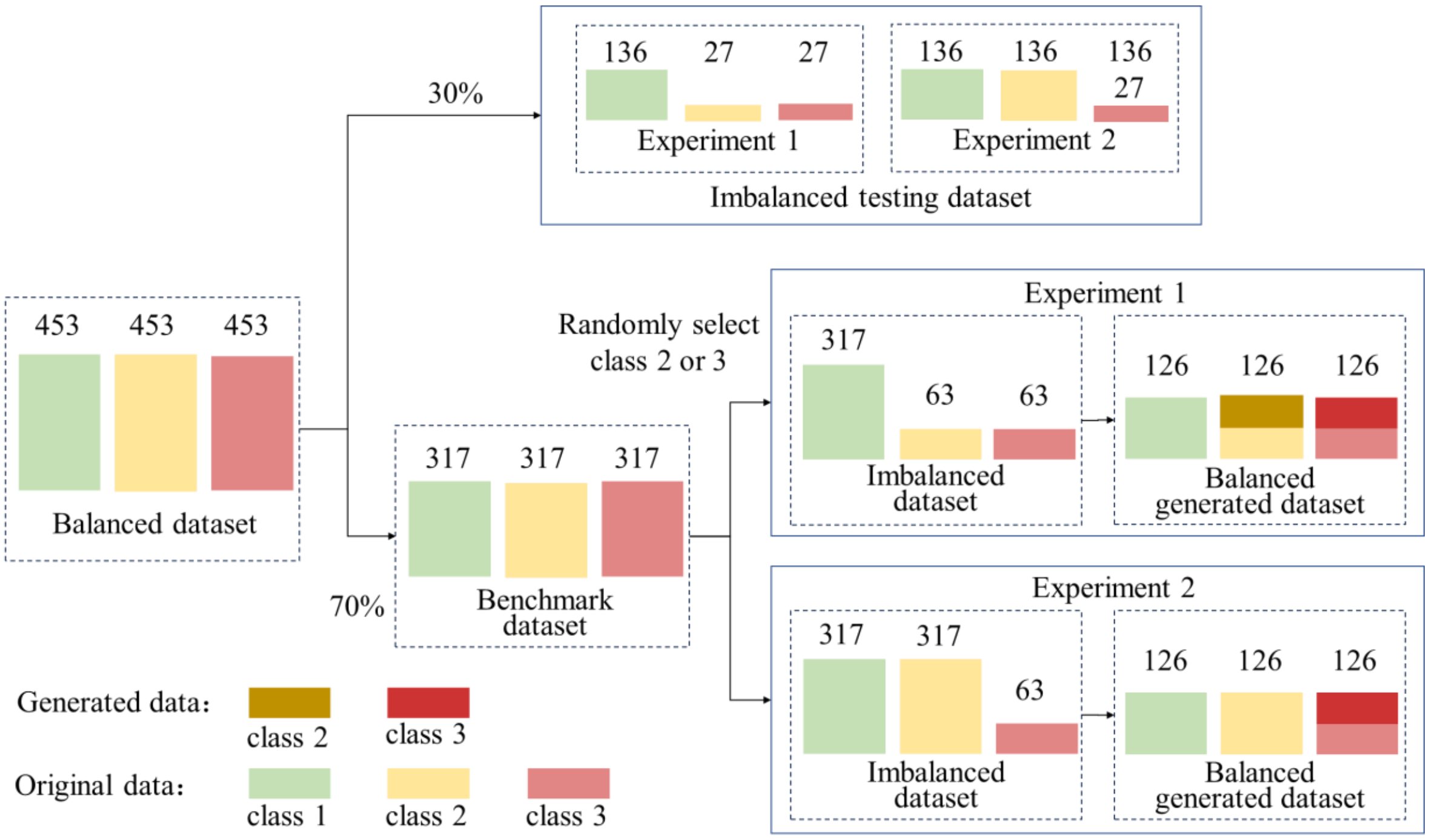

Fig. 8 Experimental design for three-class imbalance

In contrast to binary crash severity modeling, there is a need to adapt regression modeling techniques and accuracy metrics for ordered crash severity modeling. In this paper, we utilize the ordered logit model, which accommodates the ordered nature of the dependent variable, namely, crash severity. Sensitivity, specificity, and G-mean for the multiple classes are calculated in a manner analogous to binary classification (Mohammadpour *et al.* 2023). The classification accuracy of the ordered logit model is summarized in Table 10. The performance of the models using the synthesized dataset generated by CTGAN-RU is better than benchmark data. Comparing the results to the one minority to two majorities scenario (CTGAN-RU Experiment 1), the two minorities to one majority scenario (CTGAN-RU Experiment 2) aligns more closely with the benchmark dataset. These findings underscore the effectiveness of the proposed generation approach.

Table 10 Classification accuracy for ordered crash severity modeling

| Measurements | Sensitivity | | | Specificity | | | G-mean |
|---|---|---|---|---|---|---|---|
| | Class 0 | Class 1 | Class 2 | Class 0 | Class 1 | Class 2 | |
| Benchmark | 0.728 | 0.185 | 0.741 | 0.611 | 0.798 | 0.926 | 0.734 |
| CTGAN-RU Experiment 1 | 0.765 | 0.185 | 0.667 | 0.574 | 0.822 | 0.933 | 0.747 |
| CTGAN-RU Experiment 2 | 0.765 | 0.185 | 0.741 | 0.593 | 0.847 | 0.914 | 0.755 |

Fig. 9 and 10 present the distributions of the balanced benchmark data and synthetic data generated by CTGAN-RU for Experiment 1 and Experiment 2, respectively. In Fig. 8(1)-(10), corresponding to Experiment 1, deviations in the conditional and joint distribution between synthetic data by CTGAN-RU and benchmark dataset are minimal for both class 1 and 2. When compared to Experiment 1, the deviations in the distribution of synthetic data for class 2 are slightly larger in Experiment 2, yet they remain comparable to the benchmark dataset. These findings underscore that CTGAN-RU is proficient in capturing data characteristics involving sparse variables, providing a well-fitted distribution of risk factors for majority crashes, in scenarios of both one minority to two majorities and two minorities to one majority.

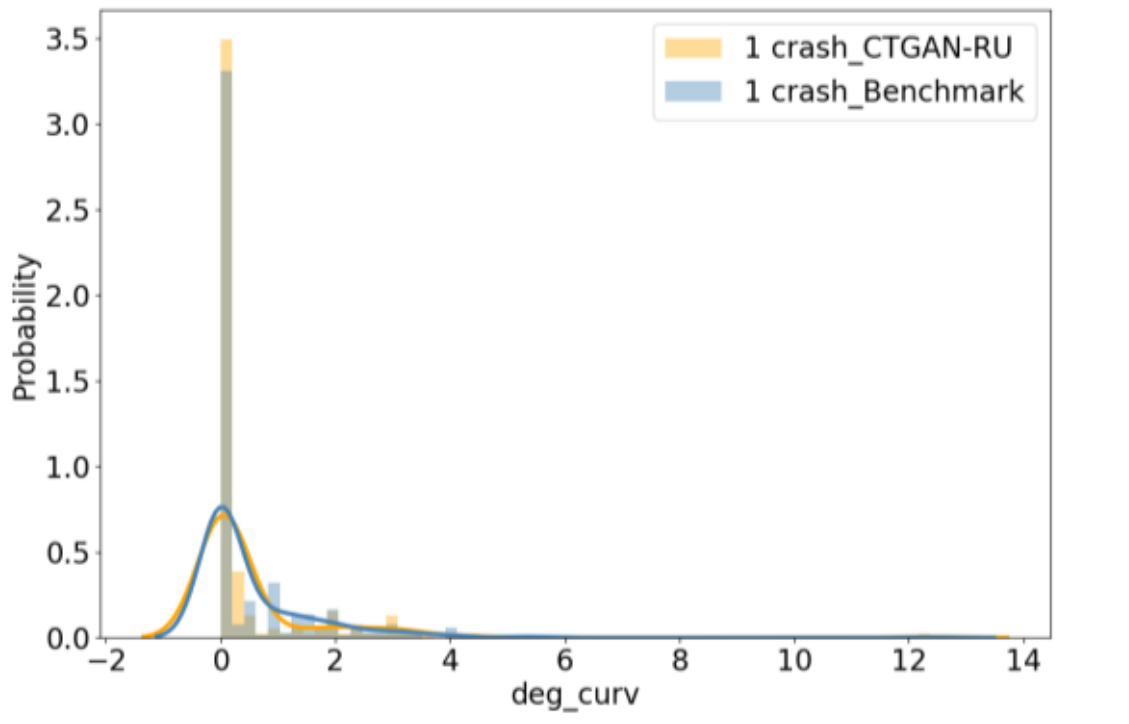

(1) Curvature degree (Class 1)

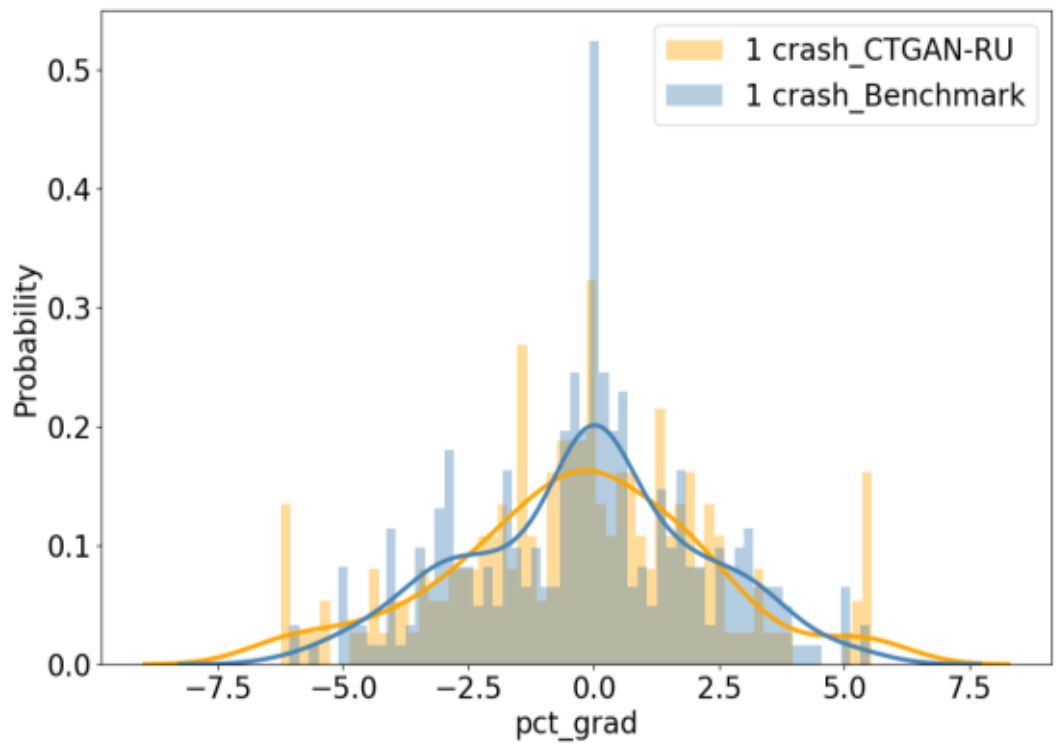

(2) Grade percentage (Class 1)

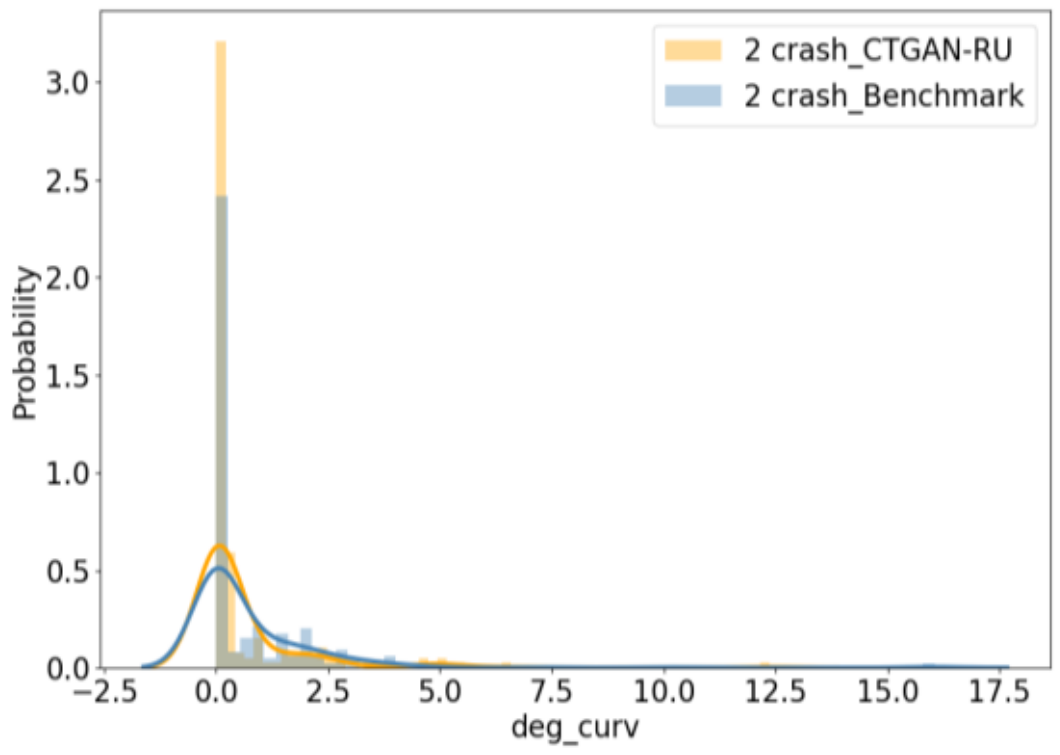

(3) Curvature degree (Class 2)

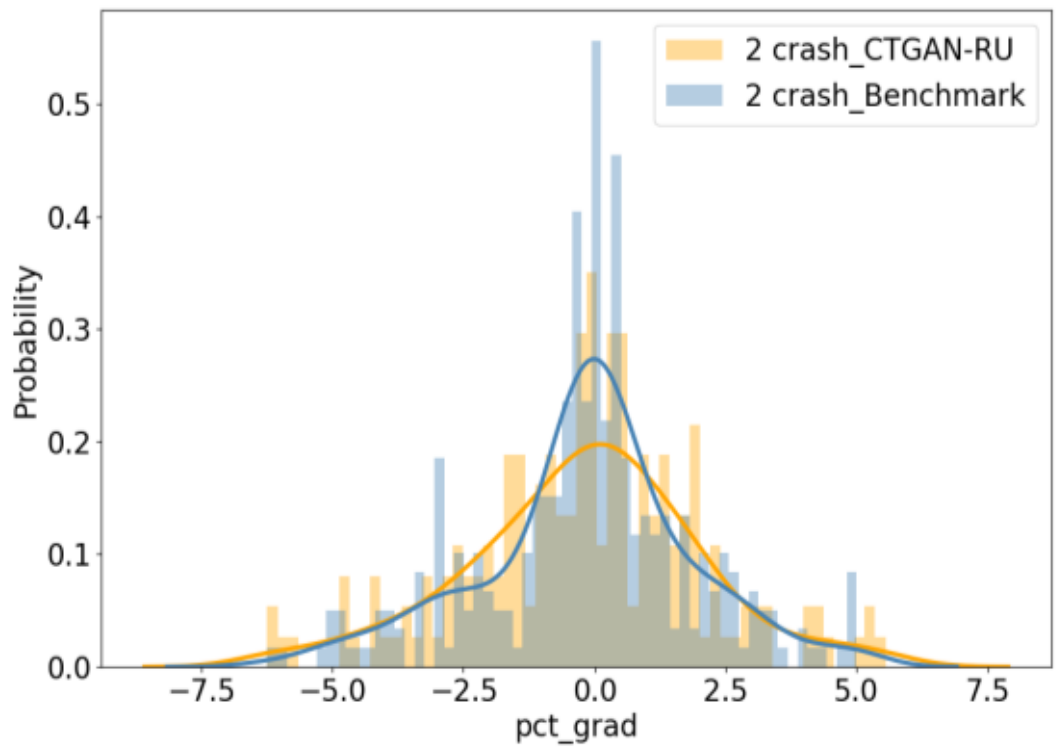

(4) Grade percentage (Class 2)

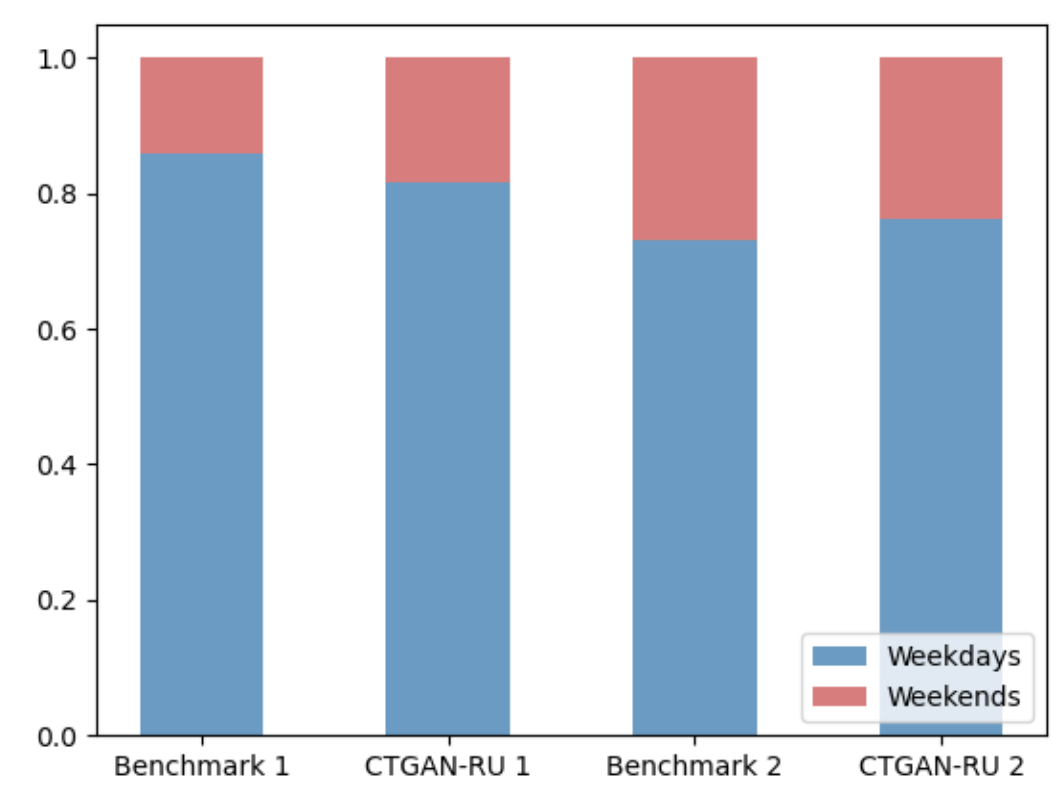

(5) Weekdays (Class 1 and class 2)

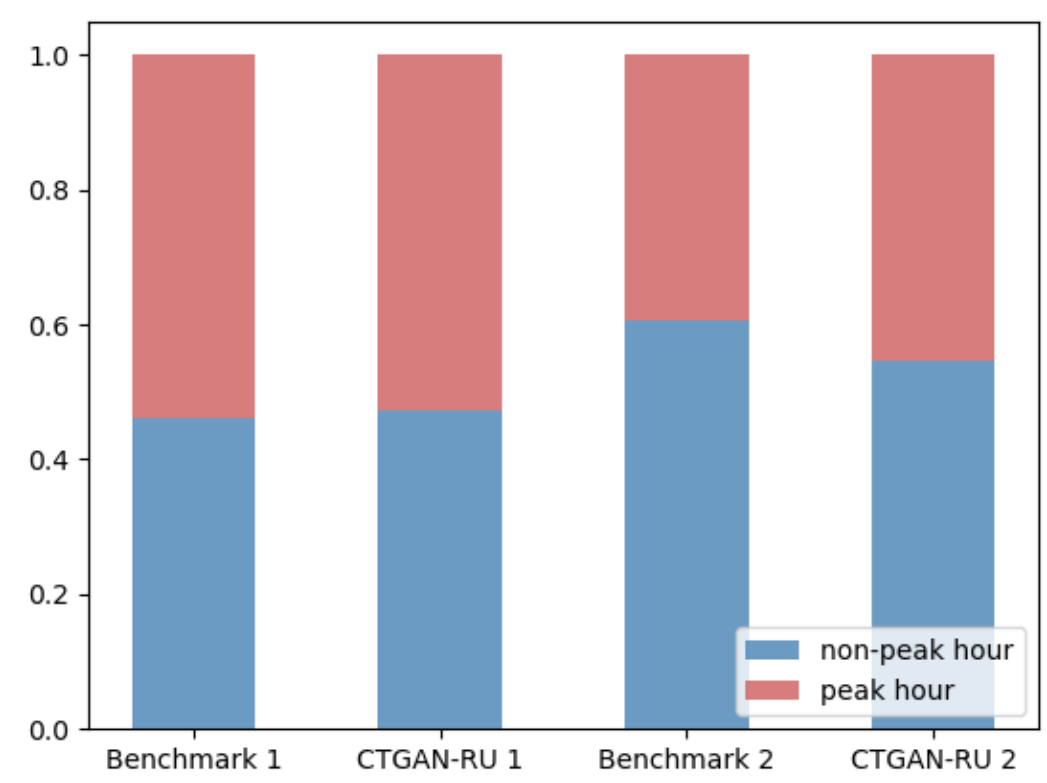

(6) Peak hour (Class 1 and class 2)

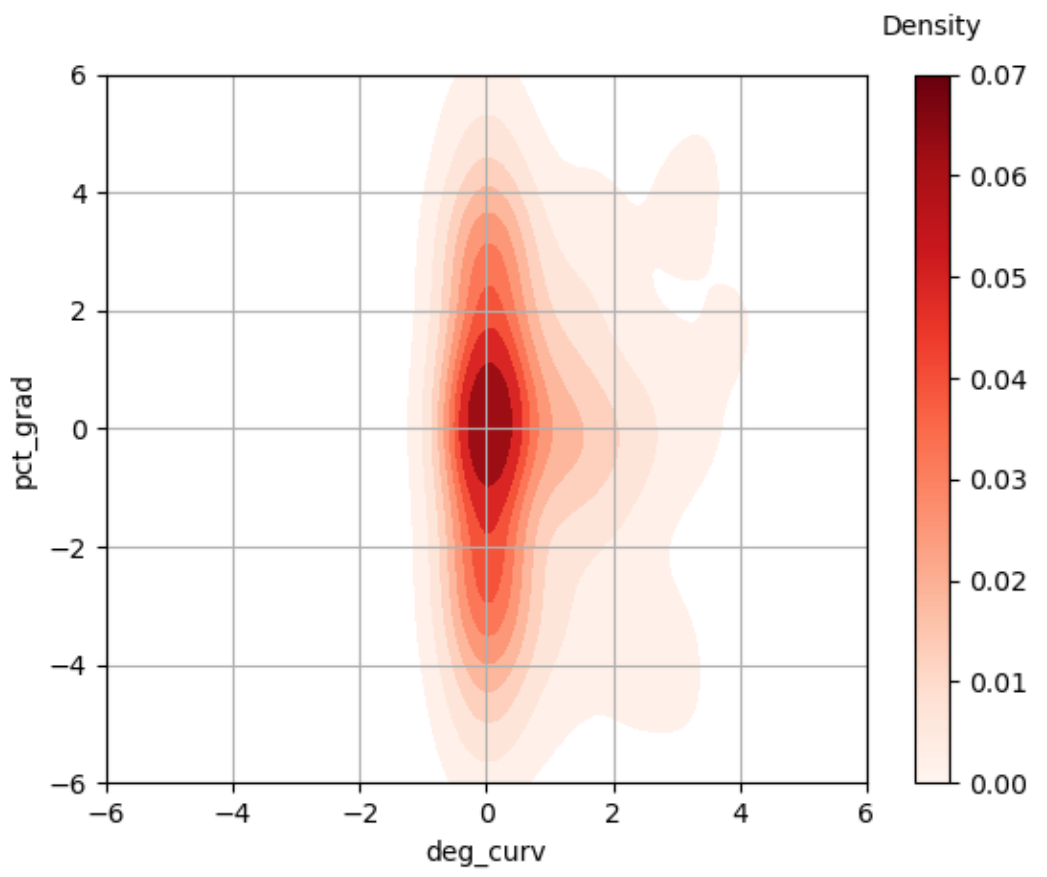

(7) Grade percentage and Curvature degree (Benchmark class 1)

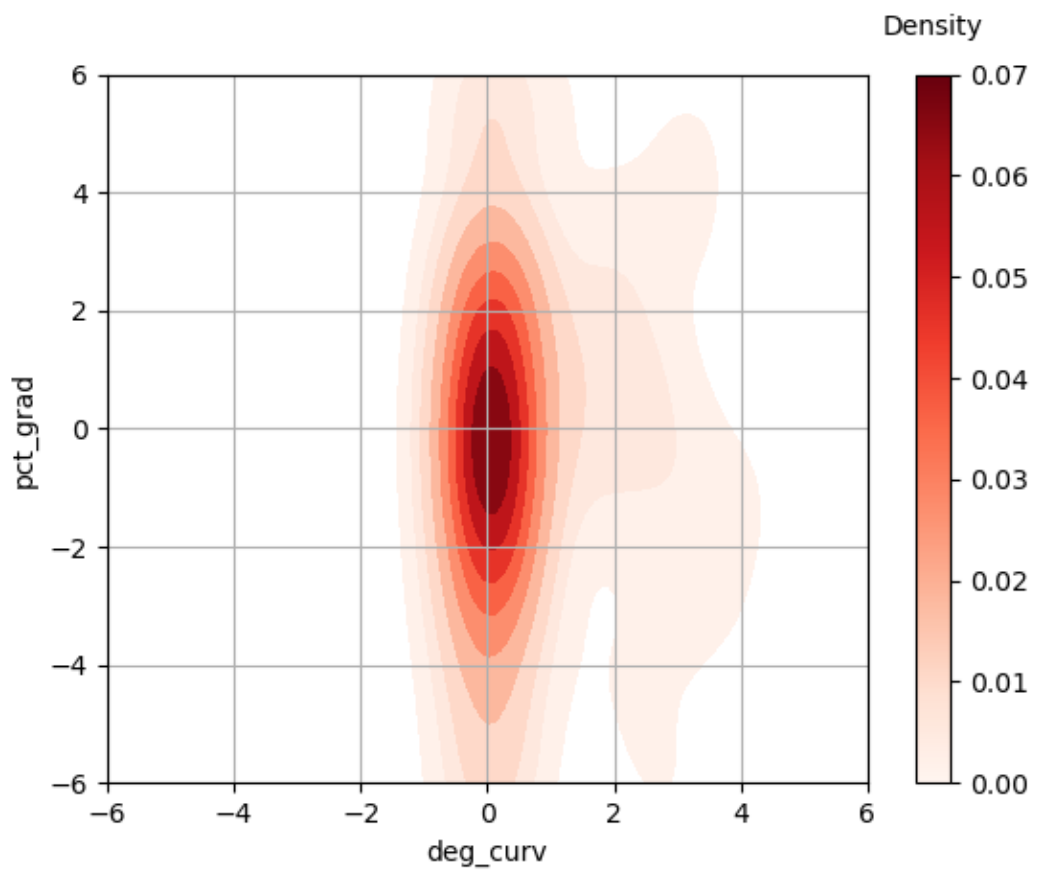

(8) Grade percentage and Curvature degree (CTGAN-RU class 1)

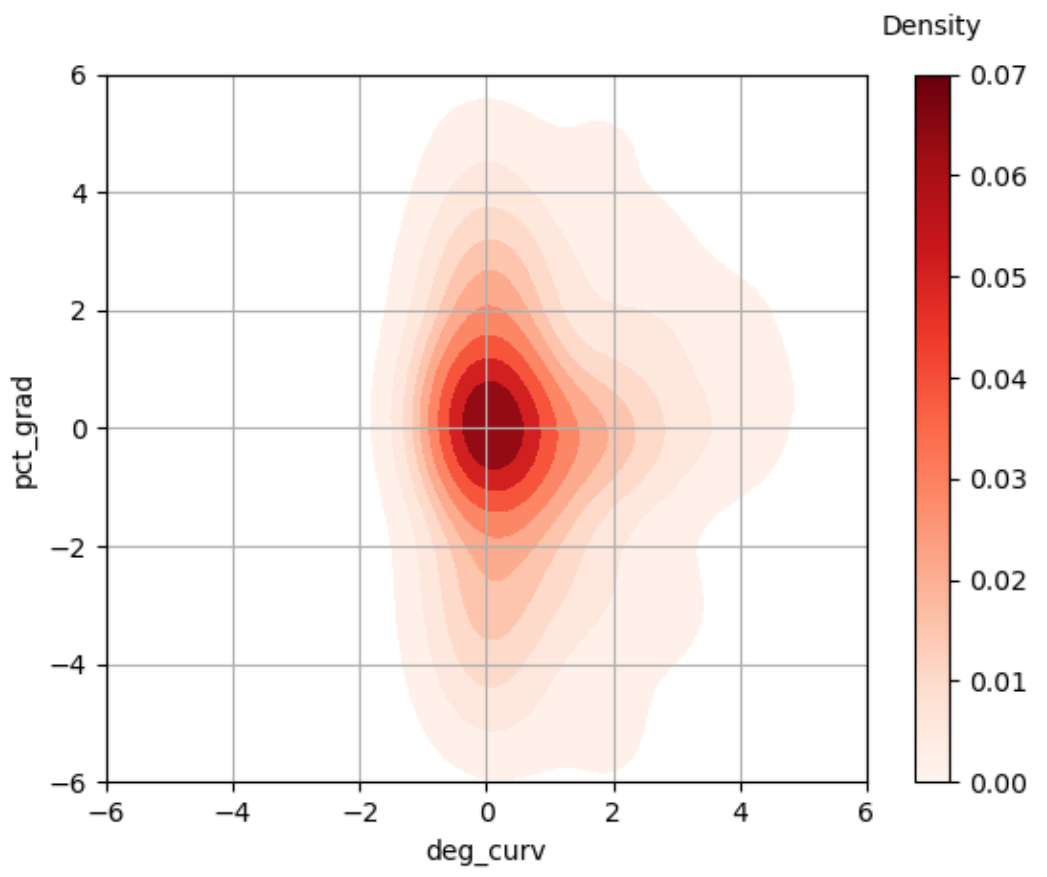

(9) Grade percentage and Curvature degree (Benchmark class 2)

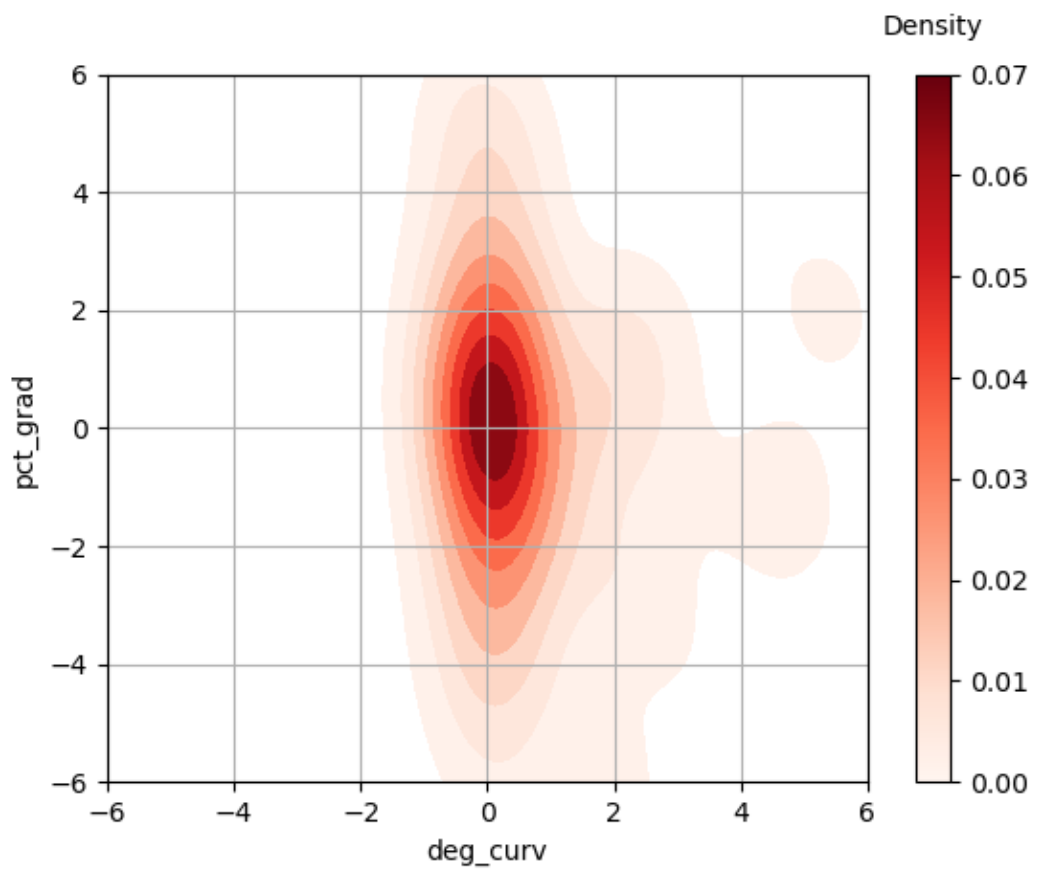

(10) Grade percentage and Curvature degree (CTGAN-RU class 2)

Fig. 9 Conditional and joint probability distribution of synthetic data by CTGAN-RU in Experiment 1

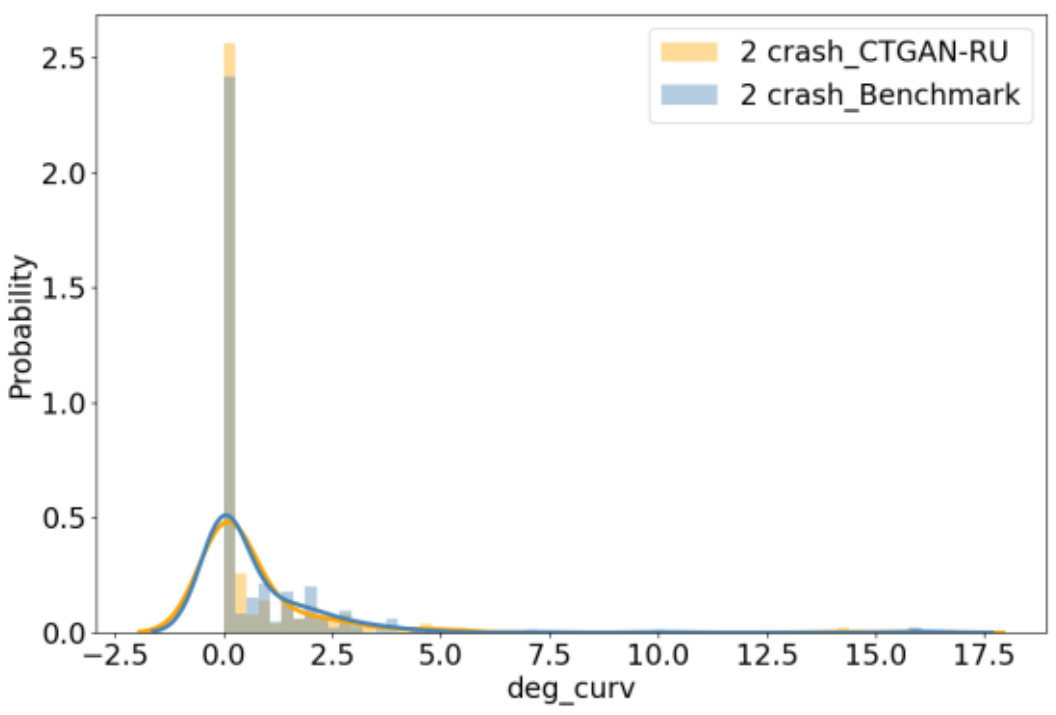

(1) Curvature degree (Class 2)

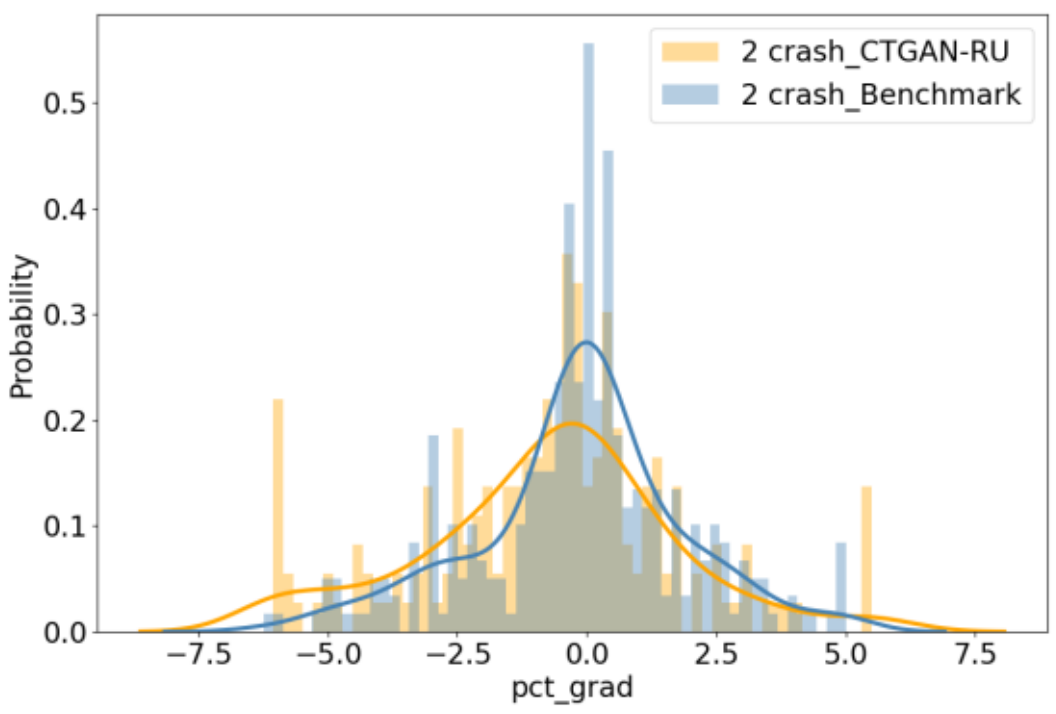

(2) Grade percentage (Class 2)

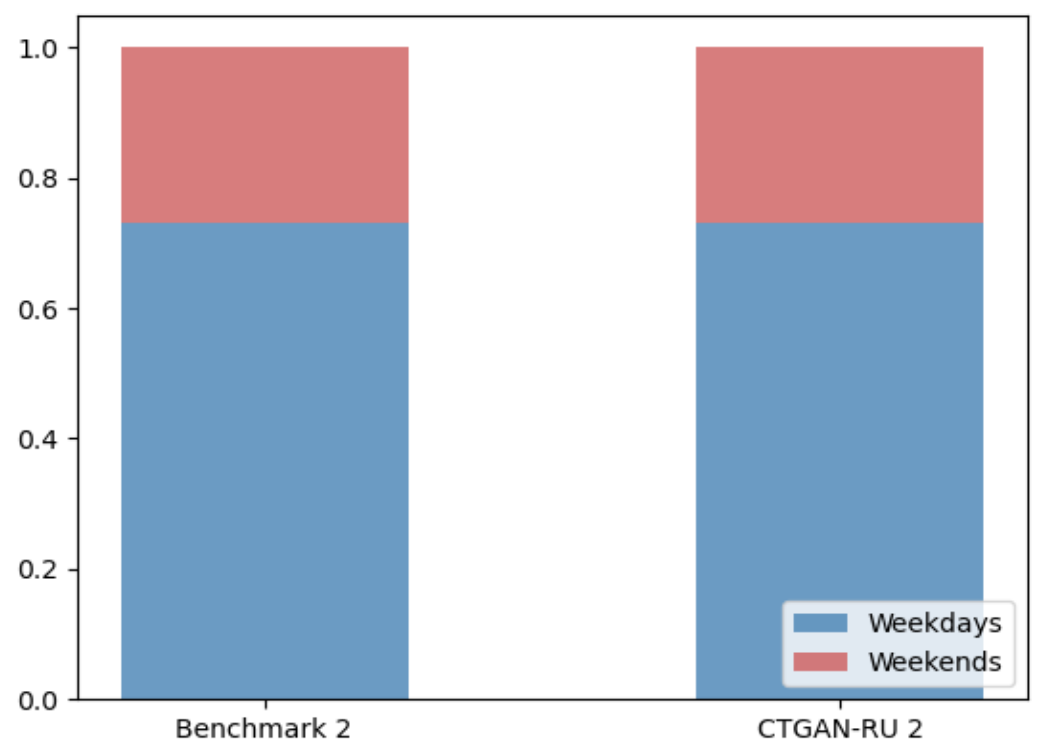

(3) Weekdays (Class 2)

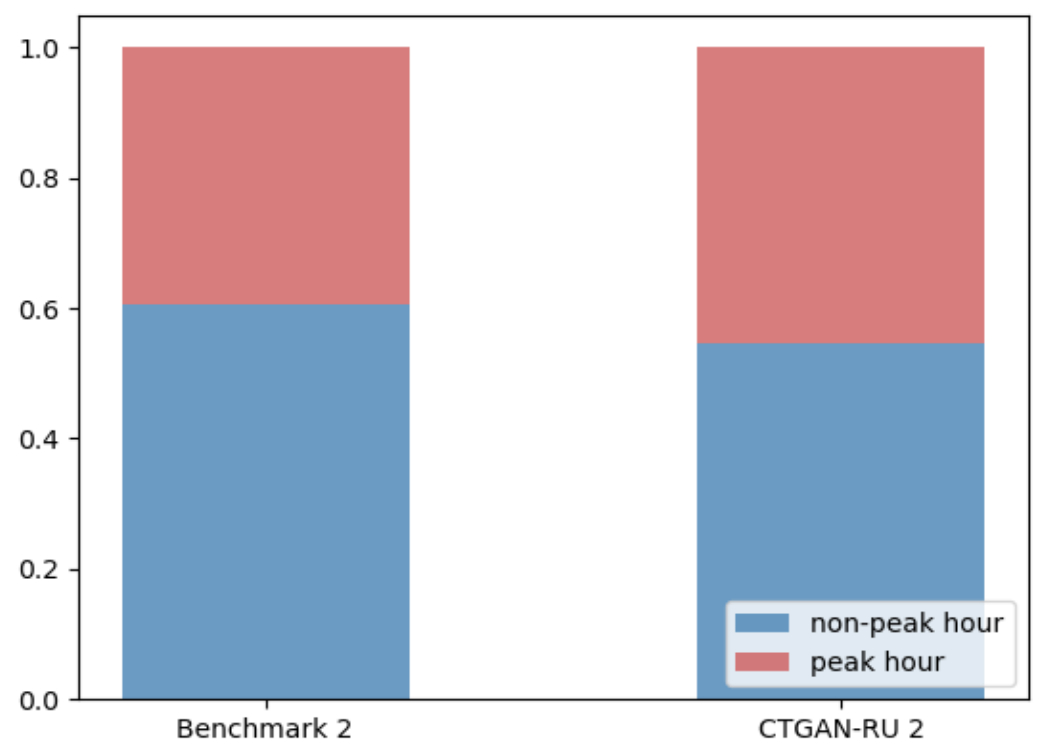

(4) Peak hour (Class 2)

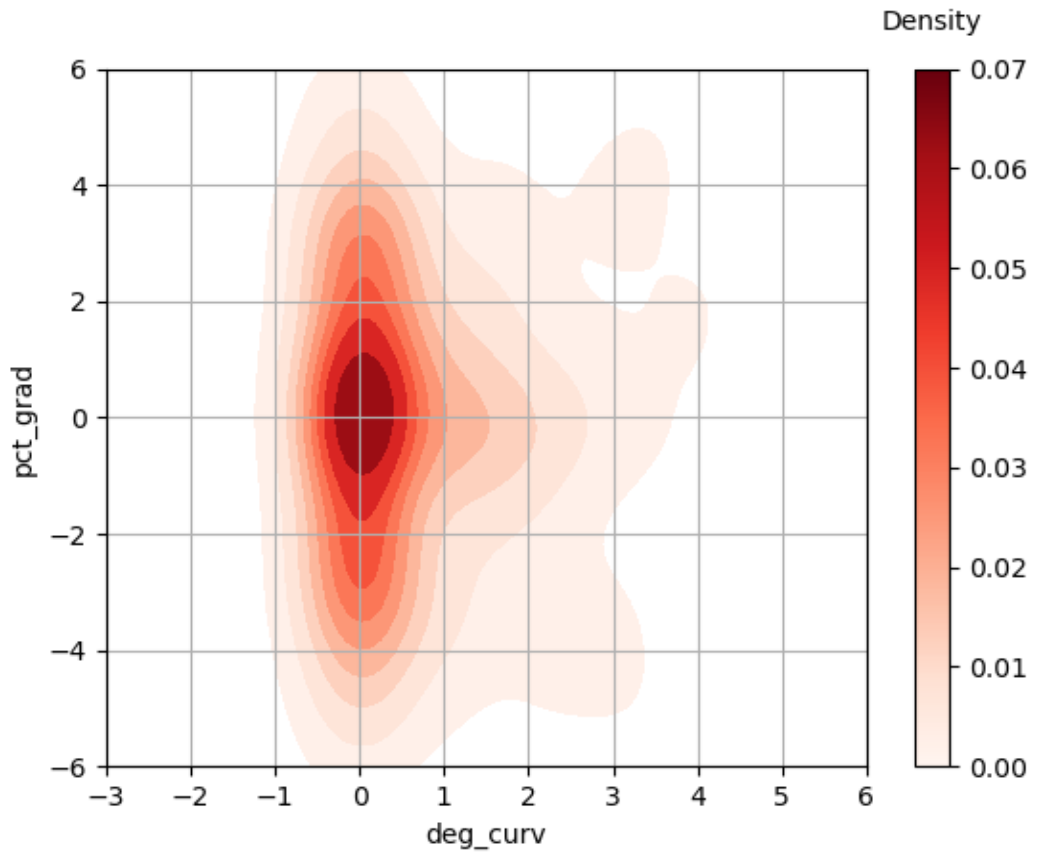

(5) Grade percentage and Curvature degree (Benchmark class 2)

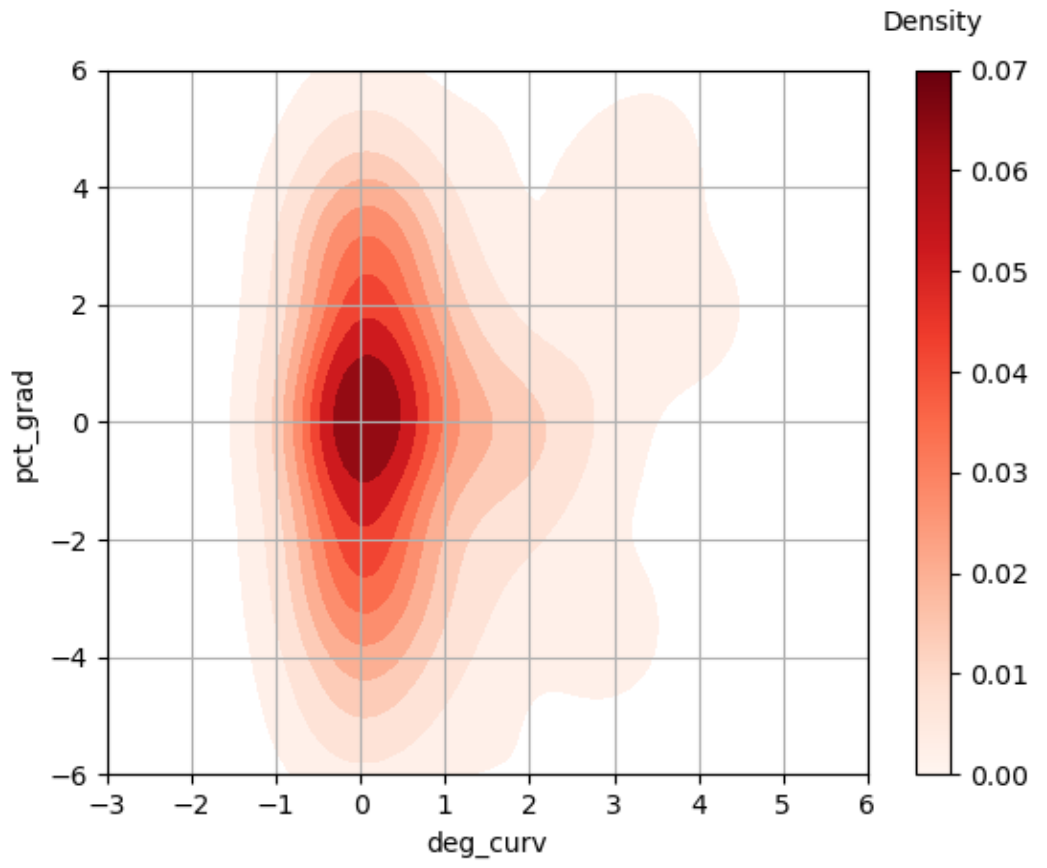

(6) Grade percentage and Curvature degree (CTGAN-RU class 2)

Fig. 10 Conditional and joint probability distribution of synthetic data by CTGAN-RU in Experiment 2

*5.4 Monte Carlo Simulation for Parameter Recovery*

Considering the true values of model parameters are unknown in real datasets, we have conducted a series of Monte Carlo simulations with predefined model parameters serving as the true values. Monte Carlo simulations are structured based on the preceding experiments presented in Fig. 4 and Fig. 8, respectively. Firstly, simulated benchmark datasets are generated with pre-defined parameters. Subsequently, we apply the proposed data generation method CTGAN-RU to create balanced datasets based on the imbalanced dataset, and then compare the estimated parameters with the true values. Finally, we investigate how the resampling ratio affects the accuracy of parameter estimation and probability prediction.

To explore the parameter recovery properties of the proposed data generation method (CTGAN-RU), two normally distributed independent variables $X_1, X_2$ and one imbalanced binary independent variable $X_3$ are generated based on the following function (see Eq. (8)):

$$X_1 \sim \text{Nomal}(0,1),\ X_2 \sim \text{Nomal}(0,1),\ X_3 \sim \ \text{Bernoulli}(0.2) \tag{8}$$

For binary logit modeling, the observed $Y$ follows a Bernoulli distribution (see Eq. (4)). For ordered logit modeling, the observed $Y$ is determined by the latent variable $Y^*$ and thresholds $\gamma_m,\ m \in \{1,2,3\}$ (see Eq. (9)). The error term $\varepsilon$ follows a standard logistic distribution. We here define $\beta =(2,2,2)$.

$$Y^* = X^{'}\beta + \varepsilon, Y = m,\ \text{if}\ \gamma_{m-1} < Y^* \le \gamma_m \tag{9}$$

We first compare the estimated parameters. Specifically, the true parameters are set to be known in the simulated dataset. Fig. 11 presents the outcomes obtained from 1,000 Monte Carlo replications across three distinct scenarios: binary logit modeling, varying sampling ratios (i.e., the ratio of the majority to minority classes), and ordered logit modeling. For binary logit modeling and ordered logit modeling, the sample sizes are set to be 4,000 and 6,000 respectively. From Fig. 11 one can see that the box plots display the estimated parameters from the CTGAN-RU synthetic dataset. The blue dashed lines represent the true values while the red line denotes the mean value. In the case of binary logistic modeling, one can observe that the estimated parameters ($\beta_1, \beta_2, \beta_3$) are similar to the true values (Fig. 11 (1)), demonstrating the effectiveness of the proposed deep generative approach in parameter recovery. Regarding the various resampling ratios, it is found that the estimated parameters ($\beta_1, \beta_2$) of the normally distributed independent variables are not sensitive to the sampling ratios (Fig. 11 (2) and (3)). Nevertheless, for imbalanced resampling ratios exceeding 1, we identify a bias in the estimated parameter ($\beta_3$) of the imbalanced independent variable (i.e., the red line in the box plots). When it comes to ordered logit modeling, the estimated parameters based on our proposed data generation method exhibit are also similar to the true values. It is worth noting that, Experiment 1 (comprising one majority class and two minority classes; Fig. 11(5)), exhibits a larger bias in comparison to Experiment 2 (comprising two majority classes and one minority class; Fig. 11(6)). This observation suggests that generating multi-class datasets may lead to an increase in bias for the estimated parameters. To sum up, these findings indicate that the estimated parameters based on the dataset synthesized by the proposed generation method consistently align with those based on the simulated benchmark dataset.

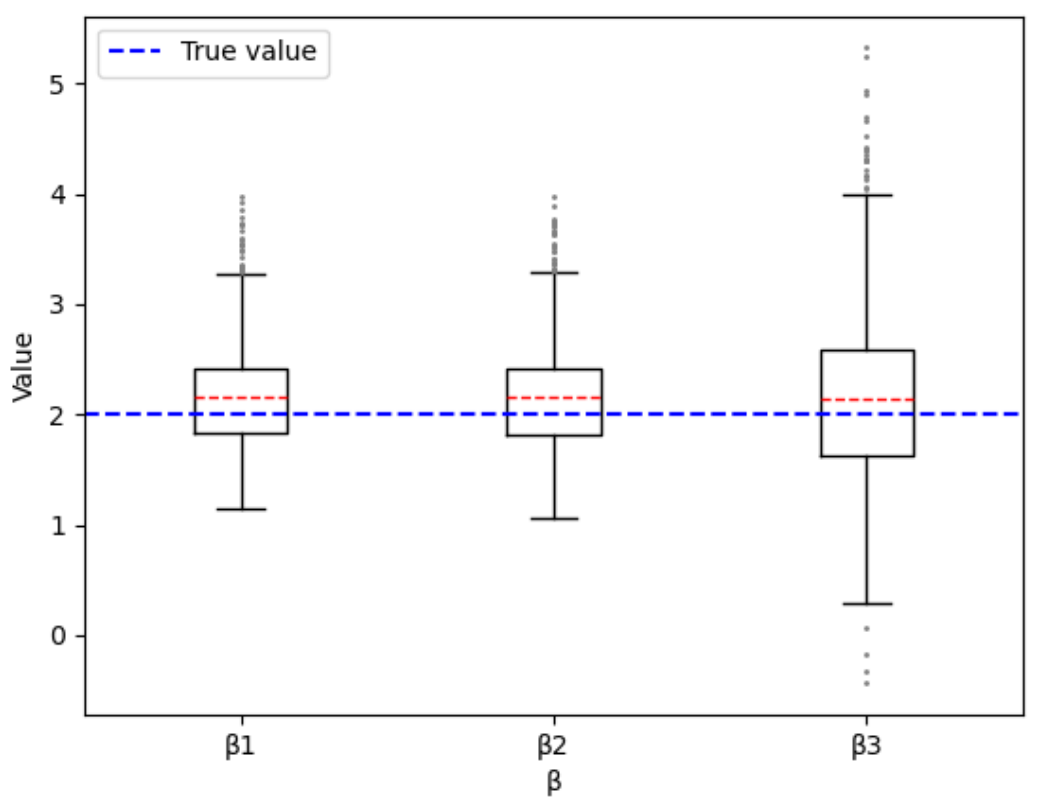

(1) Binary logit modeling ($\beta_1, \beta_2, \beta_3$)

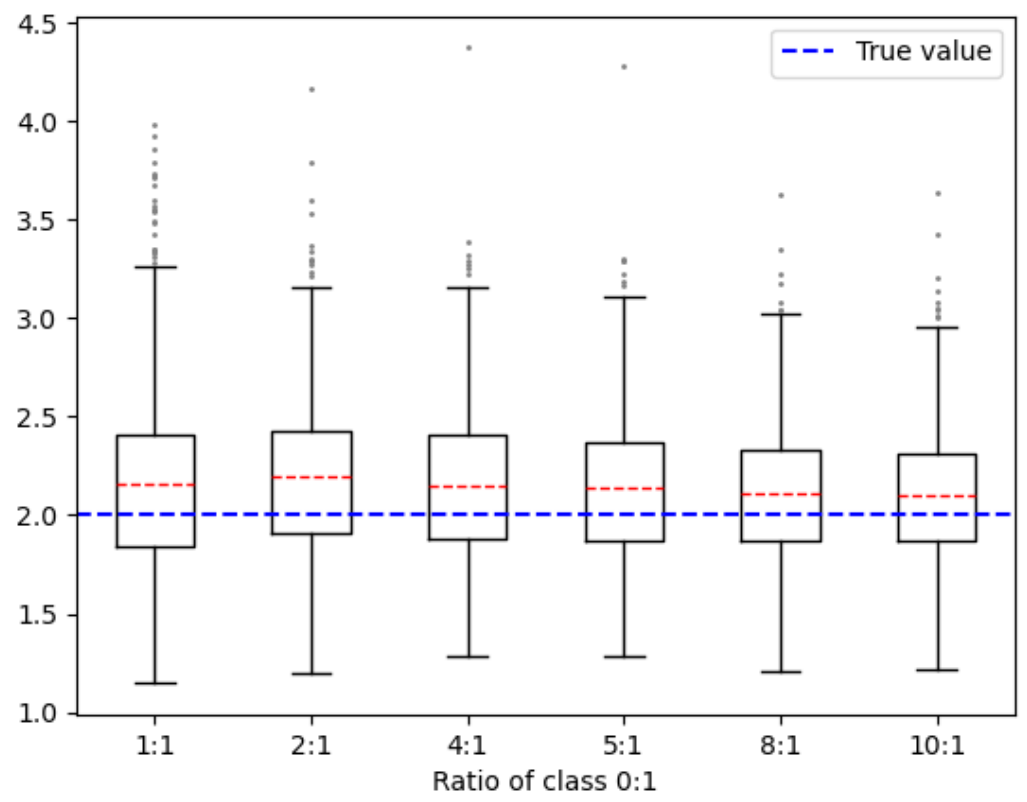

(2) Various resampling ratios ($\beta_1$)

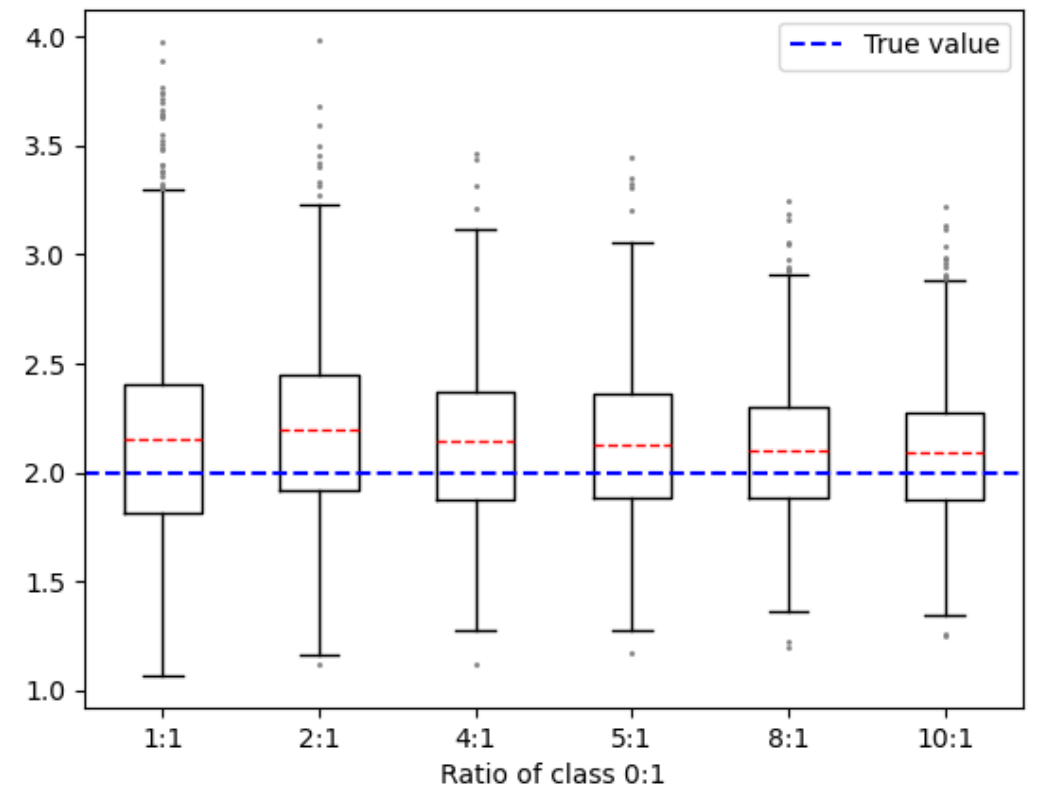

(3) Various resampling ratios ( $\beta_2$ )

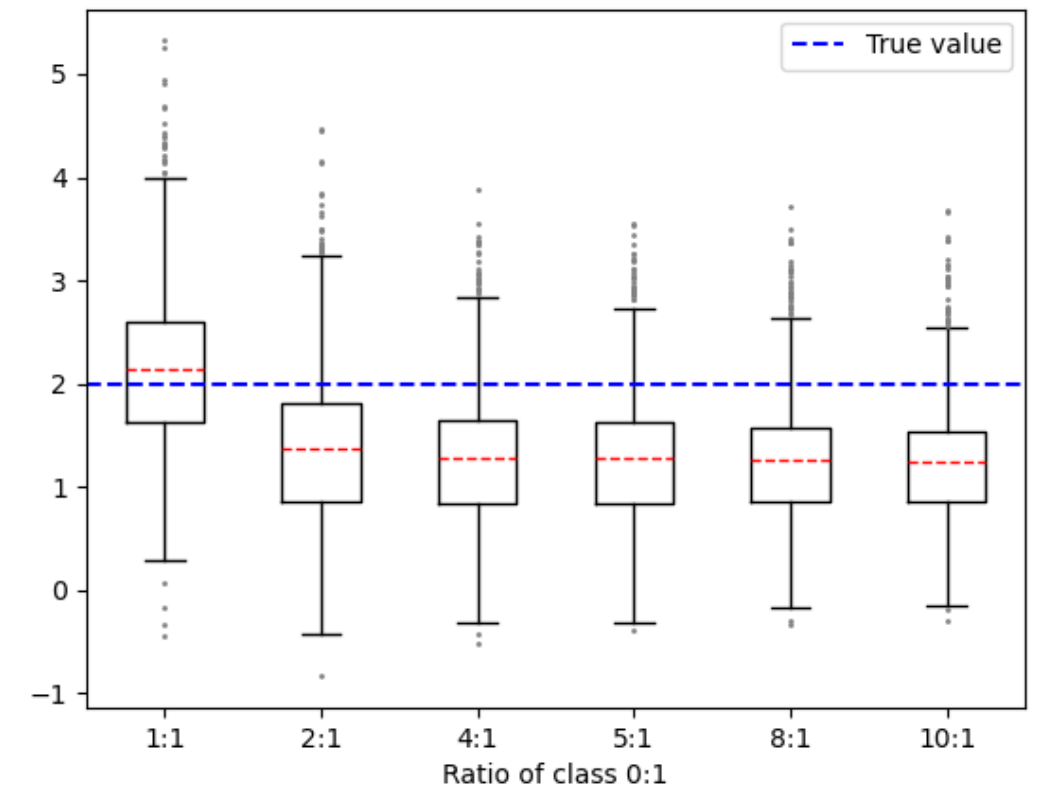

(4) Various resampling ratios ( $\beta_3$ )

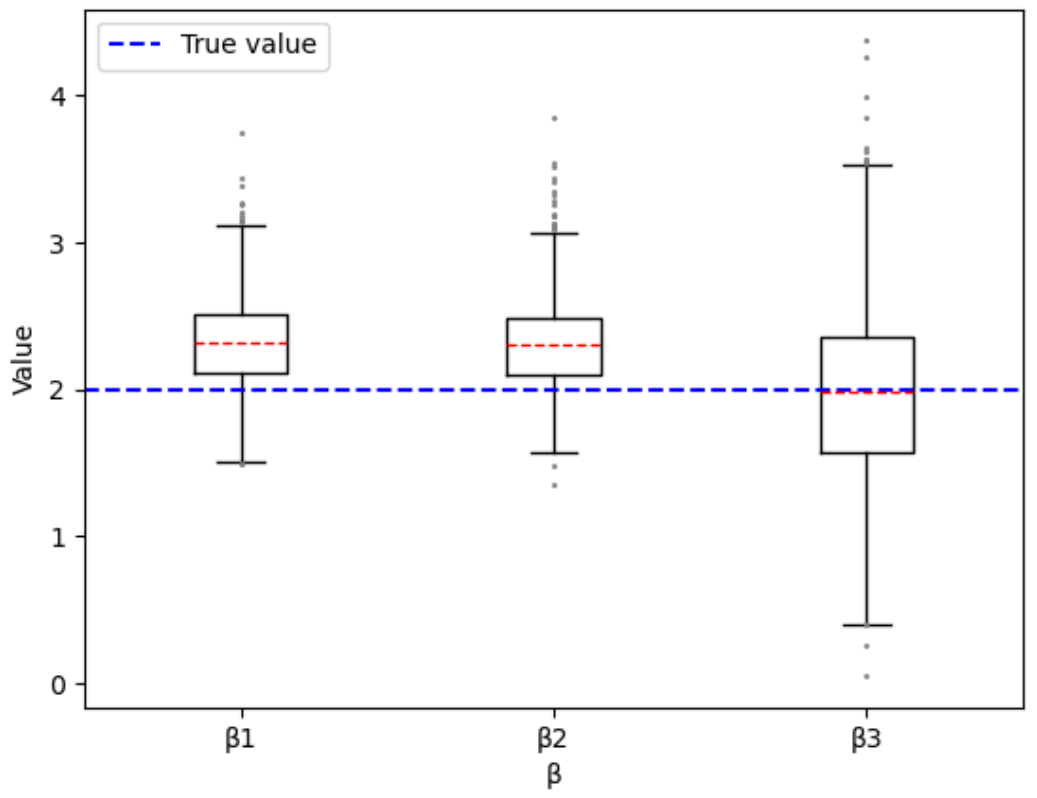

(5) Ordered logit modeling for Experiment 1 ( $\beta_1, \beta_2, \beta_3$ )

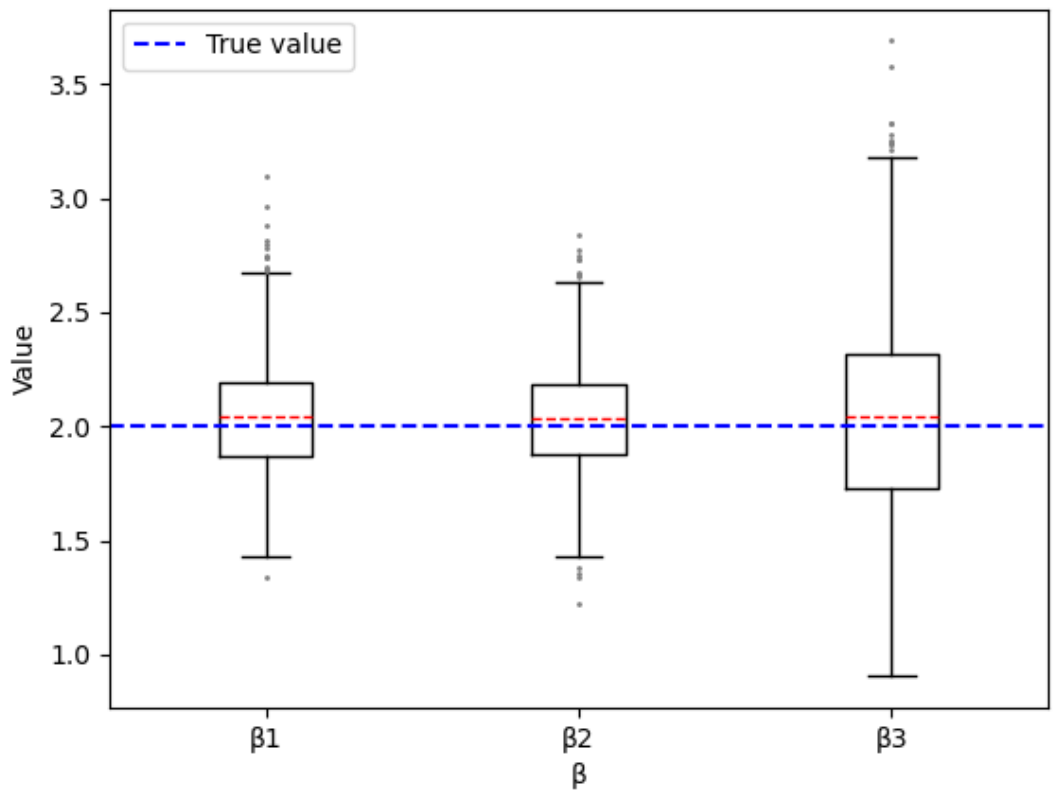

(6) Ordered logit modeling for Experiment 2 ( $\beta_1, \beta_2, \beta_3$ )

Fig. 11 Box plots for the estimated parameters using synthetic dataset by CTGAN-RU in three scenarios: binary logit modeling, various resampling ratios, ordered logit modeling

In addition to the exact values of $\beta_1, \beta_2, \beta_3$, the accuracy of estimated probability ( $P$ ) is significant in interpretative analysis. To estimate the probability ( $P$ ) accurately, one needs to have good estimates of both intercept $\beta_0$ and parameters $\beta_1, \beta_2, \beta_3$ (King and Zeng 2001). Therefore, we access the accuracy of estimated intercept (i.e., binary: $\beta_0$ ; ordered: $\beta_{01}$ and $\beta_{02}$ ) and probability ( $P$ ) using a synthetic balanced dataset by CTGAN-RU. The box plots in Fig. 12 and 13 illustrate the results for three scenarios: binary logistic modeling, various resampling ratios, and ordered logistic modeling. In each scenario, the balanced synthetic dataset and the imbalanced dataset are labeled as 'Balance' and 'Imbalance', respectively. Note that blue line represents the true values obtained from the simulation while the red line denotes the mean value. In the case of binary logistic modeling, we observe that the estimated intercepts ($\beta_0$) from the synthetic balanced dataset closely align with the true values (Fig. 12 (1)). However, there is a notable bias in the estimated intercepts ($\beta_0$) for the imbalanced dataset. MSE (Fig. 12 (2)) and Average Mean Square Error (AMSE) (see Eq. 10) of the estimated probability ($P$) (as shown in Table 11) are considerably smaller for the balanced synthetic dataset compared to those for the imbalanced dataset. There is a similar phenomenon in ordered logit modeling, as can be seen in Fig. 13 and Table 11. For various resampling ratios, the bias of the estimated intercept ($\beta_0$)and MSE of probability ($P$) become larger with the increase of the ratio of the majority class to the minority class. When the ratio is 1:1, the estimated intercept and probability are the most comparable to the true values.

The true probability is known in the simulated dataset. therefore, AMSE for $R$ Monte Carlo replications is given in Equation (10) to assess the conditional class probability prediction accuracy.

$$AMSE = \frac{1}{R}\sum_{r=1}^{R}\frac{1}{N}\sum_{i=1}^{N}\frac{1}{M}\sum_{m=1}^{M}(\mathrm{P}[Y_{i,r} = m \mid X_{i,r} = x] - \hat{\mathrm{P}}[Y_{i,r} = m \mid X_{i,r} = x])^2 \qquad (10)$$

where $N$ represents the sample size, and $M$ represents the number of $Y$ classes. We run 1000 simulation replications for three defined scenarios.

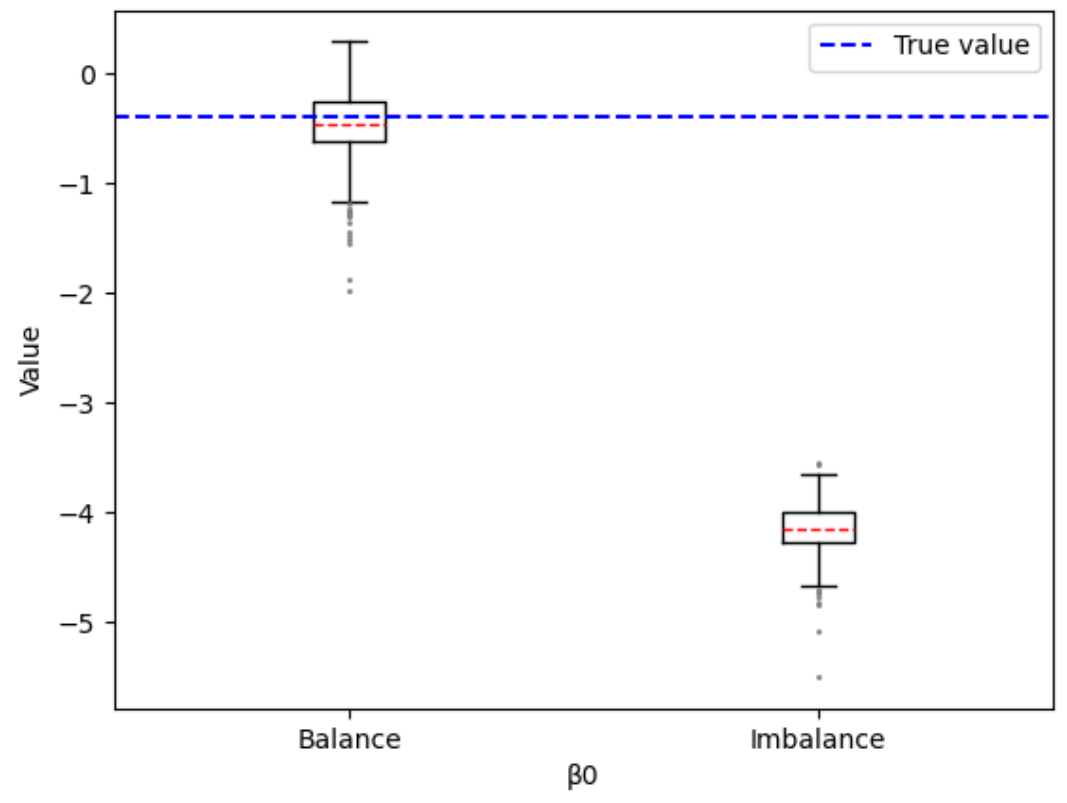

(1) Binary logit modeling ($\beta_0$)

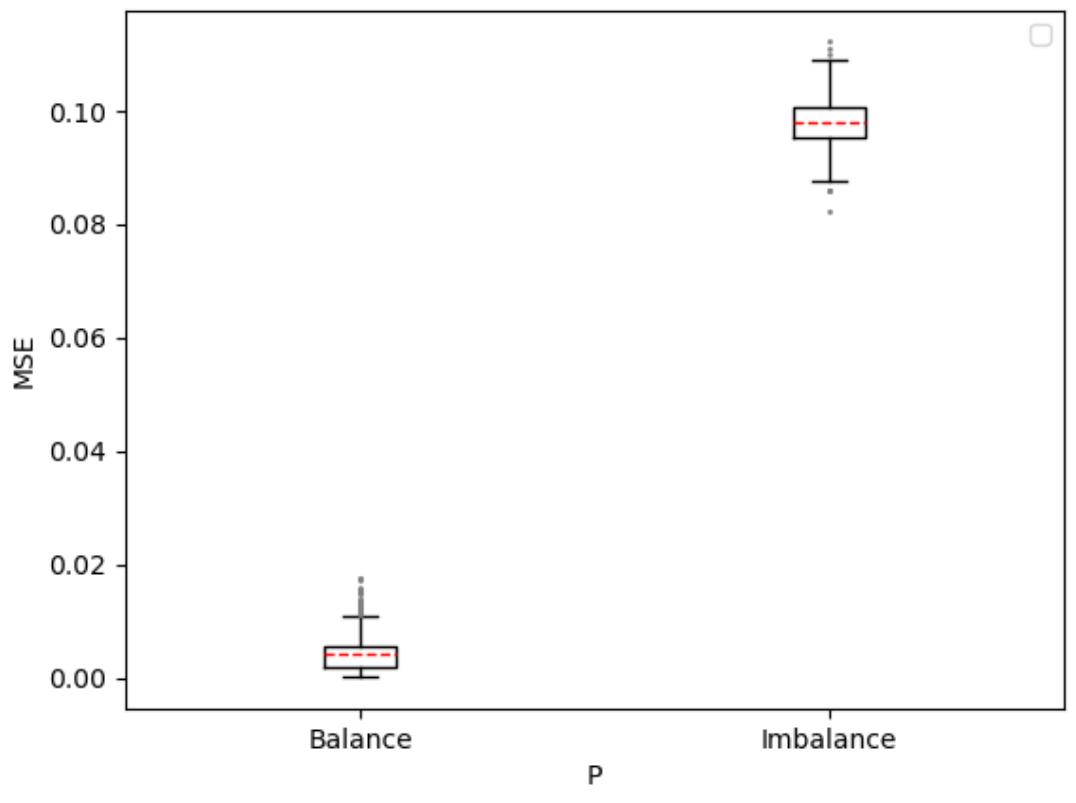

(2) Binary logit modeling ($P$)

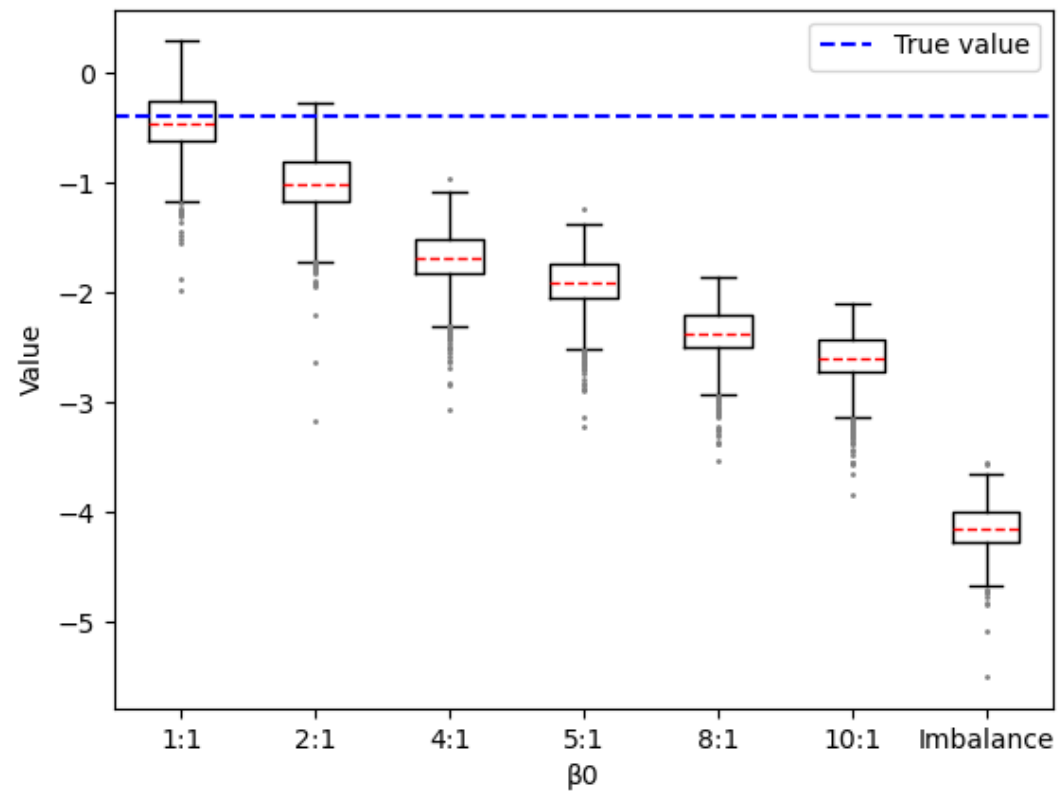

(3) Various resampling ratios ( $\beta_0$ )

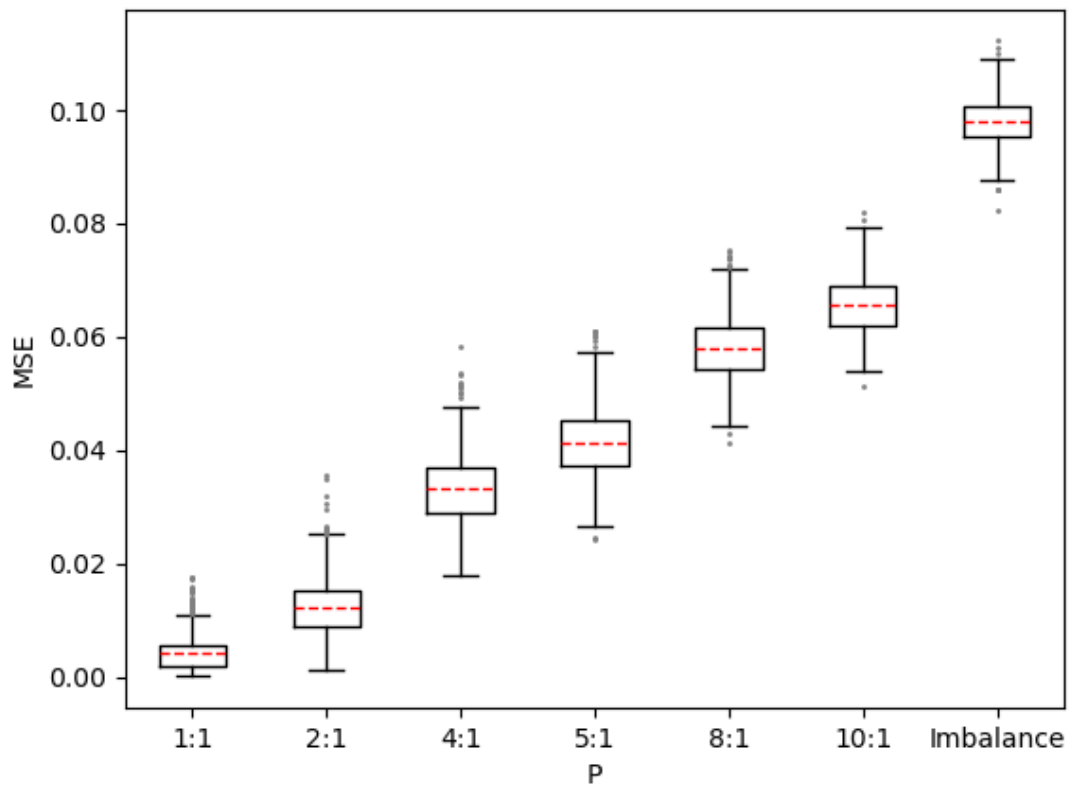

(4) Various resampling ratios ( $P$ )

Fig. 12 Box plots for the estimated intercept and probability using a synthetic dataset by CTGAN-RU in two scenarios: binary logit modeling and various resampling ratios

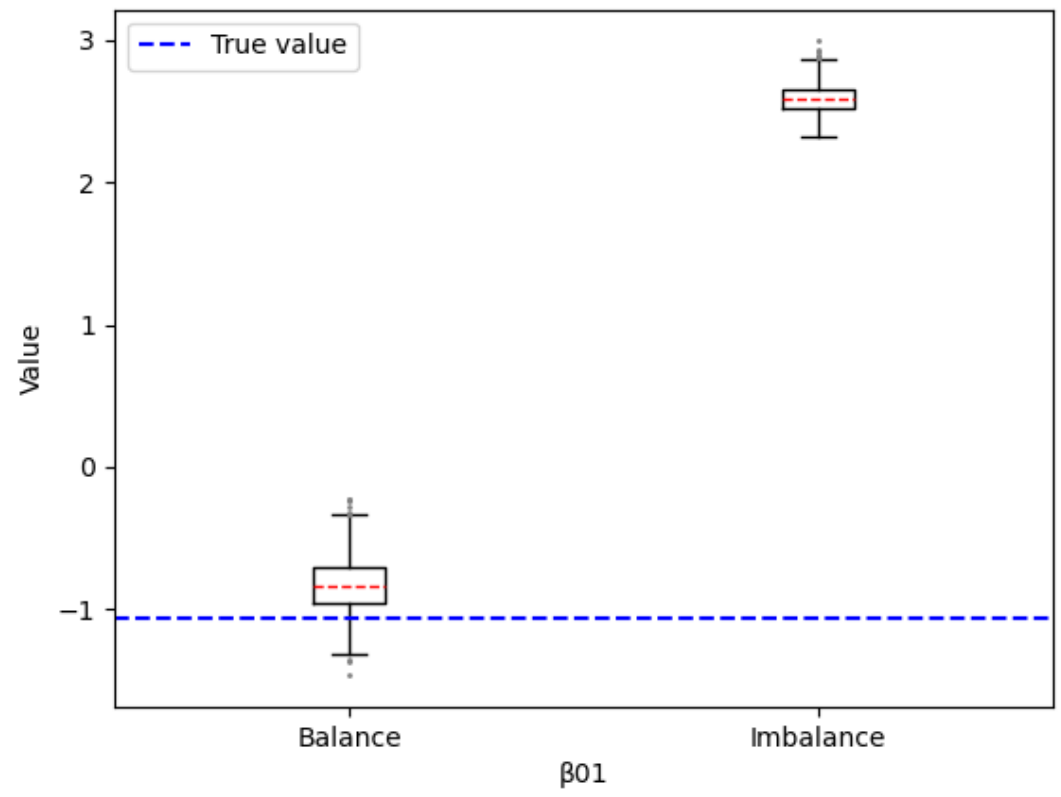

(1) Ordered logit modeling for Experiment 1 ( $\beta_{01}$ )

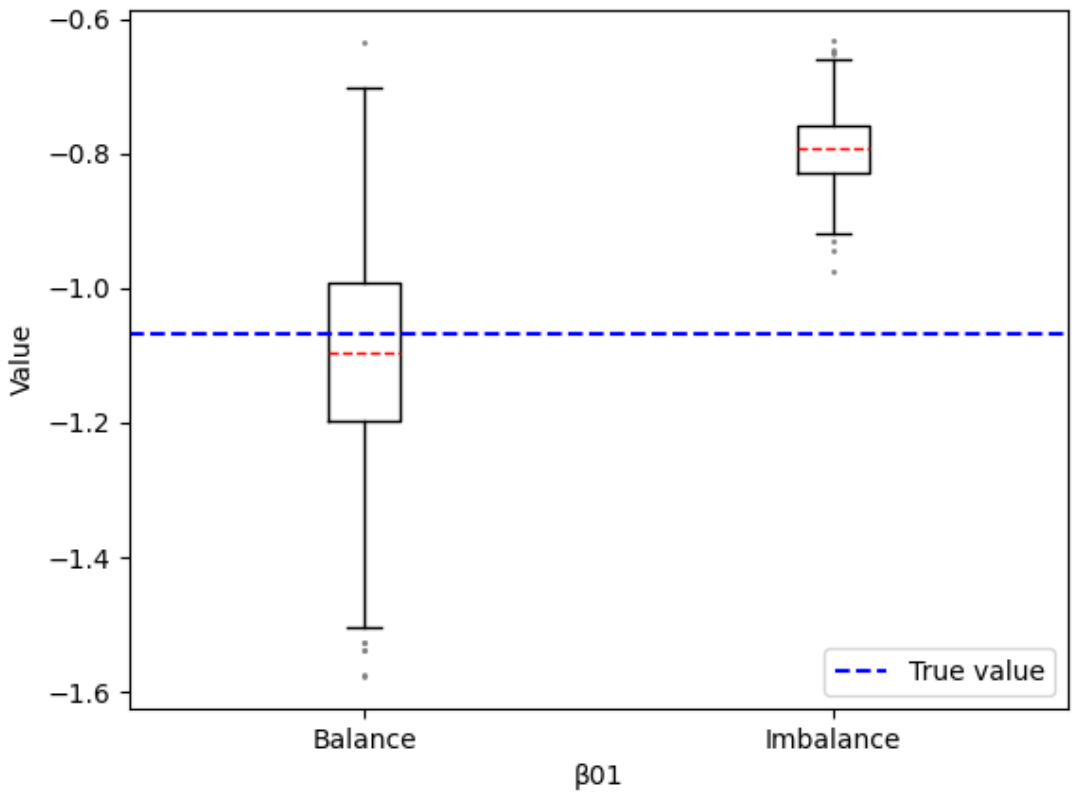

(2) Ordered logit modeling for Experiment 2 ( $\beta_{01}$ )

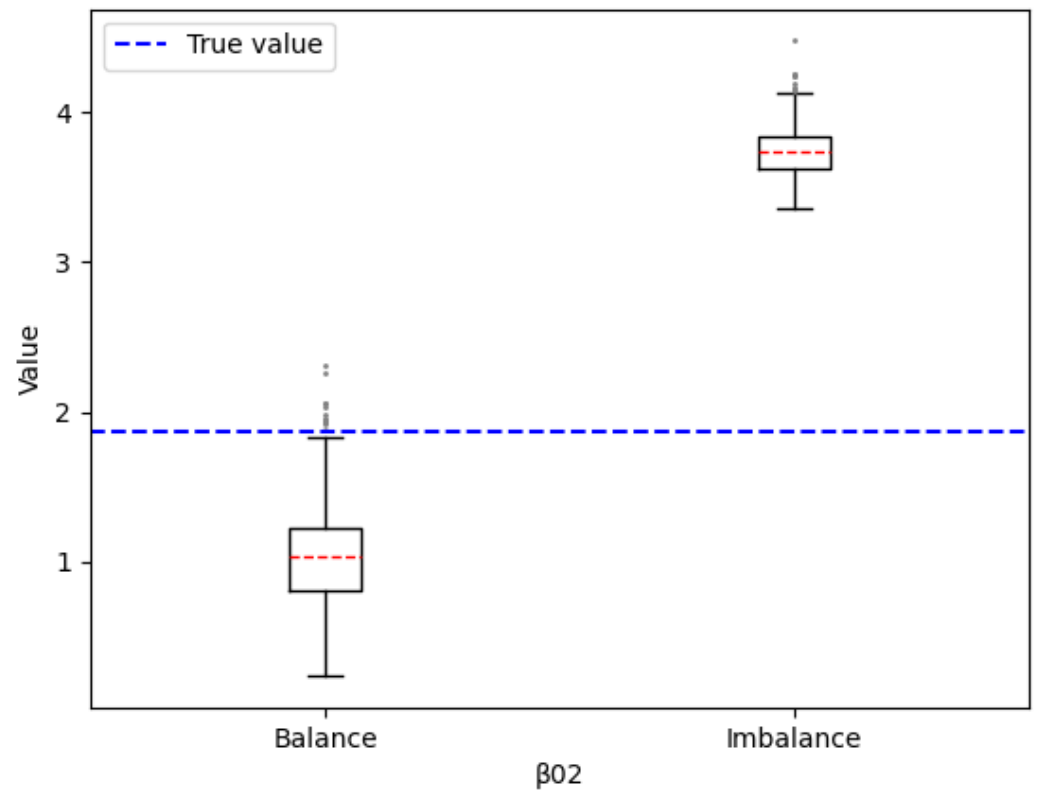

(3) Ordered logit modeling for Experiment 1 ( $\beta_{02}$ )

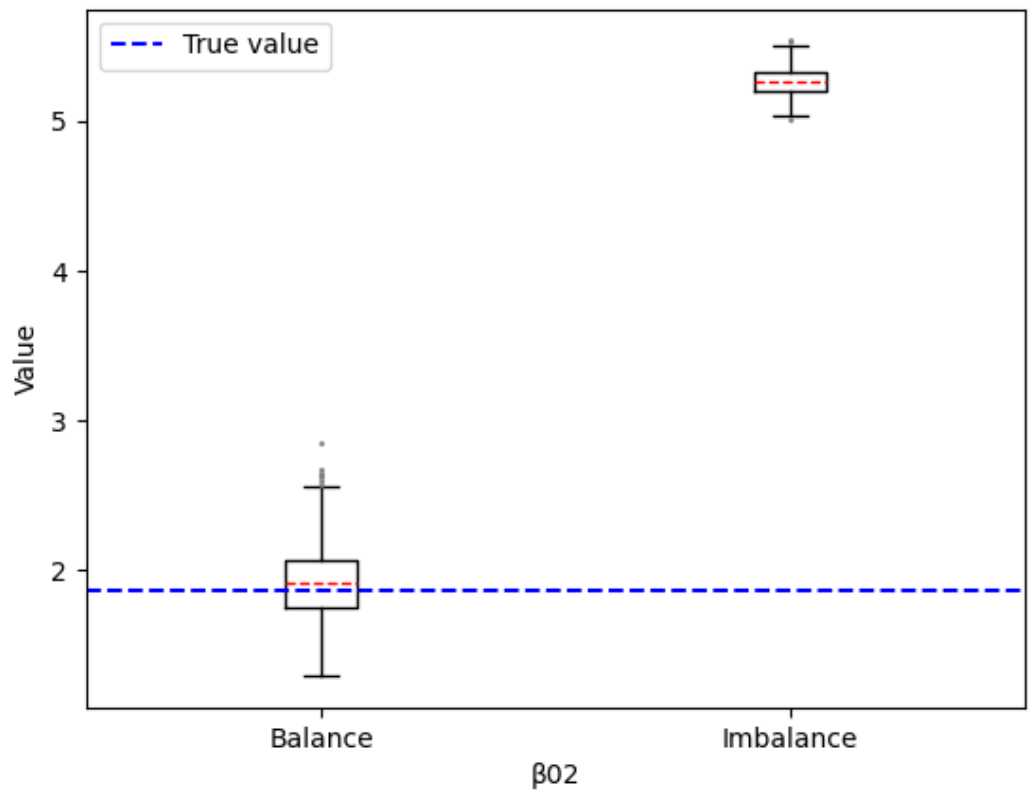

(4) Ordered logit modeling for Experiment 2 ( $\beta_{02}$ )

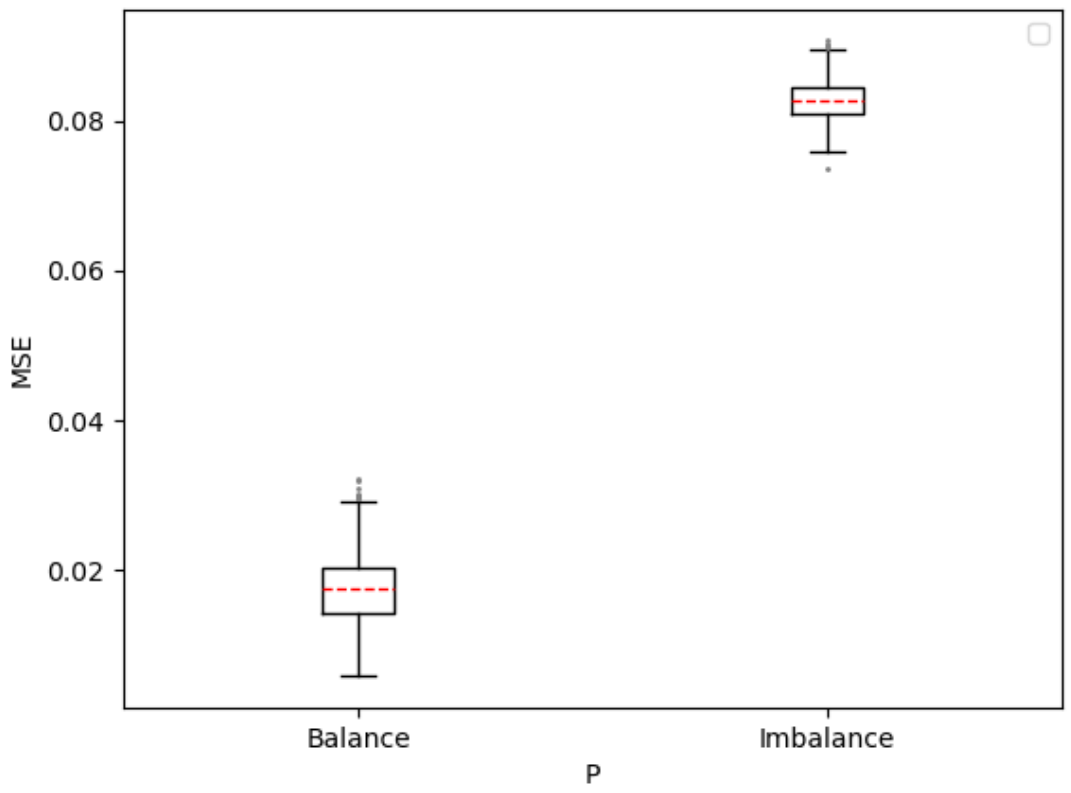

(5) Ordered logit modeling for Experiment 1 ( $P$ )

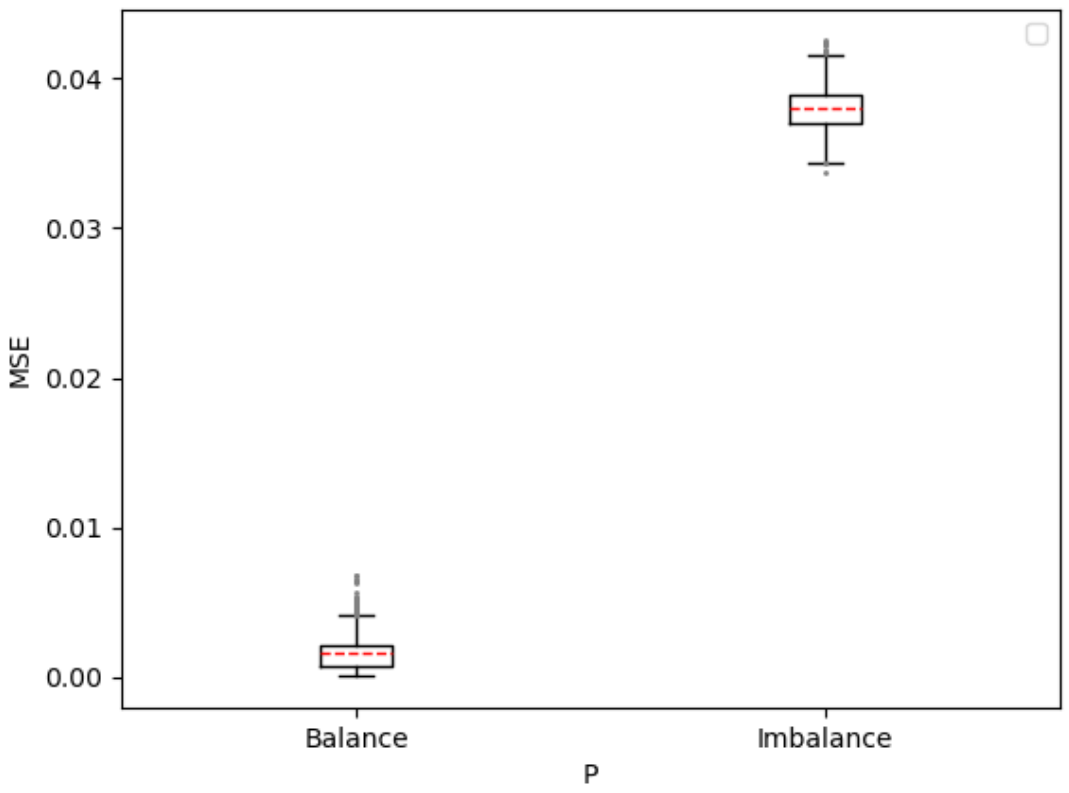

(6) Ordered logit modeling for Experiment 2 ( $P$ )

Fig. 13 Box plot for the estimated intercept and probability using synthetic dataset by CTGAN-RU for ordered logit modeling

Table 11 AMSE for estimated probability ( $P$ ) in 1000 replications simulation

| | | Binary logit Modeling (1:1) | Various resampling ratios | | | | | Ordered logit modeling | |
|---|---|---|---|---|---|---|---|---|---|
| | | | 2:1 | 4:1 | 5:1 | 8:1 | 10:1 | Ex. 1 | Ex. 2 |
| AMSE | Balanced dataset | 0.004 | 0.012 | 0.033 | 0.041 | 0.058 | 0.066 | 0.018 | 0.002 |
| | Imbalanced dataset | | | 0.098 | | | | 0.083 | 0.038 |

Notes: Ex. represents Experiment.

In short, the proposed deep learning-based data generation method has good performance in parameter and probability estimation based on the Monte Carlo simulation results

### *5.5 Crash Severity Interpretation Based on CTGAN-RU*

Table 9 shows results of parameter estimation of crash severity classification for fatal and non-fatal crashes using generation data by proposed CTGAN-RU method. Note that for the rest of the paper, the model fitted to the original data will be called the "original model", while the model fitted to the synthetic data produced by CTGAN-RU will be referred to as the "CTGAN-RU model". As illustrated in Table 12, one can see that the CTGAN-RU model performs better than the original model in terms of R-squared. This again confirms that the proposed data generation method, CTGAN-RU, can better capture the significant risk factors and the effects of risk factors.

For driver characteristics-related factors, it is identified that *Driver's_sex*, *Drunk_driving*, and *Drug_driving* are statistically significant in the CTGAN-RU model at the 90% confidence level or higher. Interestingly, the correlation between *Driver's_*sex and crash severity is found to be negative in the original model but positive in the CTGAN-RU model. The positive correlation aligns with previous studies showing that male drivers tend to take more risks than female drivers (Abdel-Aty *et al.* 1998, Li *et al.* 2021a). *Drunk_driving* and *Drug_driving* are not statistically significant in the original model. However, there is a significantly positive relationship between them and crash severity in the CTGAN-RU model, which is consistent with previous studies indicating that driving under the influence of drugs or alcohol increases the fatality risk (Donaldson et al. 2006, Buddhavarapu et al. 2013). It's important to note that there is an imbalance in the severity of these risk factors, which poses a challenge for the crash severity model in capturing the characteristics of this minority class. As shown in Table 2, for example, *Drug_Driving* contains 604 cases (0.4%) with drug involvement and 14,496 cases (99.6%) without drug involvement. Among the cases with drug involvement, only 10 (1.7%) had the FI crash severity outcome, while 594 (98.3%) did not. The original model struggled to identify these significant risk factors, but the CTGAN-RU model effectively addresses this issue, even in the presence of sparse data for these risk factors.

While the original model did not consider angle collision and head-on collision as significant factors, the CTGAN-RU model identifies their significant impact on crash severity. Moreover, the estimated parameter for head-on collision is higher than that of the other three collision types. Past research has primarily concentrated on the impact of rear-end collisions and has not given enough attention to the importance of angle collision and head-on collision on crash severity (Khattak and Fontaine 2020, Hyun et al. 2021, Li et al. 2021a, Zhou et al.

2021). While rear-end collisions are the most common type of crashes, accounting for 73.8% of total crashes (as shown in Table 2), their fatality rate is relatively low at 0.27%. On the contrary, head-on collisions are less frequent, accounting for only 0.3% of total crashes, but they have a much higher fatality rate of 4.8%. The reason for the higher fatality rate in head-on collisions is that they frequently occur on roads with high-speed limits, resulting in a higher relative speed between the vehicles. Additionally, sideways collisions are more likely to cause fatal injuries when they occur off the road, as there is a greater risk of the vehicle colliding with barriers or fixed objects. Airbag is significantly positively associated with crash severity at the 99% confidence level. This is reasonable since the deployment of airbag could most likely occur in fatal crashes that involve significant energy release (Mannering *et al.* 2016).

Another interesting finding is that although both the original and CTGAN-RU model indicate that *Location* is a significant factor at the 99% confidence level, the signs of the estimated parameters are opposed. This again reveals that CTGAN-RU can contribute to accurately capturing the effects of the risk factors. The positive correlation in the CTGAN-RU model agrees with previous studies reporting that the risk of death is higher on rural roads (Cabrera-Arnau et al. 2020). The reasons for this are manifold: lower law enforcement, higher speed limits, increased alcohol use, more distracted driving, less use of seat belts, and more (Borgialli et al. 2000, Donaldson et al. 2006, Buddhavarapu et al. 2013, Beck et al. 2017, Kardar and Davoodi 2020). Besides, grade percentage is significantly positively associated with crash severity at the 99% confidence level. This indicates that roads with a higher percentage of upgrades or downgrades can lead to more fatal crashes. Indeed, previous studies have proven that a steeper grade on the roads can result in a reduced line of sight and provide less reaction time for drivers to respond appropriately to potential collisions (Yu and Abdel-Aty 2014, Zeng et al. 2019).

Table 12 Results of parameter estimation for fatal and non-fatal crashes

| Variables | Description | Original training dataset | CTGAN-RU |
|---|---|---|---|
| | | Coeff. | Coeff. |
| Constant | | -0.780 | -2.925 |
| ***Driver characteristics*** | | | |
| Driver's_age | 26-65 | -0.957*** | 0.538 |
| | ≥66 | 0.102 | 1.098 |
| Driver's_sex | Male | -0.956*** | 0.810* |
| Drunk_driving | Drunk driving | 0.854 | 2.382* |
| Drug_driving | Drug driving | 0.537 | 4.491*** |
| ***Crash characteristics*** | | | |
| Type_of_collision | Angle collision | 0.548 | 3.349*** |
| | Sideswipe collision | -0.397 | -0.578 |
| | Head-on collision | 1.043 | 6.043** |
| Airbag_status | Airbag active | 1.482*** | 2.487*** |
| Number_of_vehicles_involved | | -0.834** | 0.431* |
| ***Temporal characteristics*** | | | |
| Peak_Hour | Peak hours | -0.973*** | -0.531 |
| Weekday | Weekends | -0.166 | -1.369** |
| ***Road characteristics*** | | | |
| Location | Rural | -1.912*** | 1.718*** |
| Curvature degree | | -0.668*** | -0.022 |
| Grade_percentage | | 0.013 | 0.323*** |
| Posted_speed_limit | | -0.020 | -0.001 |
| ***Goodness-of-fit measures*** | | | |
| Pseudo R-squ. | | 0.088 | 0.416 |

Note: Peak hours are from 6:00 to 9:00, and 16:00 to 19:00;
"+" represents an upgrade, a "-" represents a downgrade;
***p-value < 0.01; ** p-value < 0.05; * p-value < 0.1.

More importantly, the analysis results in Section 5.5 suggest some policy recommendations to improve road safety. For example, transport operators could enhance driver education, publicity and management to increase the safety awareness of the drivers, especially male drivers and nighttime drivers who may drink. It is imperative to enhance enforcement and increase the frequency of sobriety checkpoints for drivers under the influence of alcohol or drugs. Enhancing road design and maintenance to ensure visibility, and signage, thereby reducing the risk of angle collisions and head-on collisions, which lead to severe crashes. Consider placing warning signs in areas with steep gradients to alert drivers and prevent severe crashes. To reduce the occurrence of severe accidents on rural roads, it is necessary to increase law enforcement, improve infrastructure designs, and enhance emergency response measures.

## 6 CONCLUSIONS

In this study, we address the imbalanced crash severity data issue by developing a mixed-resampling method, which combines conditional tabular generative adversarial networks and random under-sampling (CTGAN-RU). This method takes into account various characteristics (both continuous and discrete) of the imbalanced data during the training process. Through a comprehensive comparison containing three categories of resampling methods (e.g., SMOTE, TVAE, random under-sampling), the binary logit model using synthetic data by CTGAN-RU is verified to have the best performance in terms of classification accuracy. CTGAN-RU showed the best performance in sensitivity and G-mean values, especially when the ratio of RU is 2:1 and CTGAN-RU is 1:1. According to the experiments and Monte Carlo simulation study, our proposed data generation method demonstrates enhanced performance in mitigating mode collapse, distribution variations, and parameter recovery when compared to benchmark data. This improvement is observed not only in the context of addressing binary imbalance issues but also extends to the domain of addressing three-class imbalance issues, outperforming other sampling methods. Applying CTGAN-RU to imbalanced fatal and non-fatal crashes with sparse risk factors resulted in better classification accuracy and mode fit. Our proposed data generation method is better equipped to identify the sparse risk factors associated with fatal crashes.

In future work, our focus will be on enhancing the robustness of the proposed data generation method. Specifically, we aim to improve the method's ability to handle imbalanced crash data with missing data, heteroscedasticity, noisy data, and small sample sizes. Furthermore, we intend to investigate the correction for oversampling ratios using discrete choice models (Rainey and McCaskey 2021, Krueger *et al.* 2023), and to contrast the effectiveness of this model-based approach with our data-driven approach.

## ACKNOWLEDGMENTS

The research is supported by the Ministry of Transport of PRC Key Laboratory of Transport Industry of Comprehensive Transportation Theory (Nanjing Modern Multimodal Transportation Laboratory): [Grant No. MTF2023002]. We thank Dr. Yingheng Zhang and Dr. Chi Wei for useful comments that significantly improved the presentation of this article.

## APPENDIX

In order to demonstrate that the proposed CTGAN-RU can indeed improve accuracy, we conduct multiple experiments with different random seeds (i.e., 1, 2, 3, 4) according to the Section 4.2 Data Resampling. As can be seen in Table A, The results suggest that the proposed method, CTGAN-RU, consistently outperforms the other three commonly used resampling methods in terms of G_mean, indicating the robustness of the proposed model.

Table A Classification accuracy of FI and nFI crash severity models (Mean±Std.dev)

| Measurements | Training dataset | CTGAN | CTGAN-RU | SMOTE-NC | TVAE | RU |
|---|---|---|---|---|---|---|
| Sensitivity | 0 | 0.708± 0.072 | **0.892± 0.063** | 0.541± 0.093 | 0.466± 0.074 | 0.713± 0.059 |
| Specificity | 1 | 0.821± 0.031 | **0.837± 0.020** | 0.867± 0.047 | 0.826± 0.036 | 0.686± 0.071 |
| G_mean | 0 | 0.761± 0.036 | **0.863± 0.031** | 0.682± 0.057 | 0.619± 0.051 | 0.699± 0.034 |

"Std.dev" represents the standard deviation